%% file: thesis.1side.tex
\documentclass[12pt,letterpaper]{book}

\input{format.1side}

\begin{document}

\include{chapter0}

\begin{thesis}

\setcounter{page}{1}

\include{chapter1}

\include{chapter2}

\include{chapter3}

\include{ref}

\end{thesis}

\end{document}

%% file: format.1side.tex
\usepackage{fancyhdr}
\usepackage{cite}
\usepackage{epic}
\usepackage{psfig}
\usepackage{latexsym}
\usepackage{amsmath}
\usepackage[usenames]{color}
\usepackage[dvips]{epsfig}
\usepackage{graphicx}
\usepackage{verbatim}
\usepackage{setspace} 

\renewcommand{\cleardoublepage}{\newpage}

\fancypagestyle{plain}{ 
\fancyhf{} 
\fancyfoot[C]{\bfseries\thepage} 
 
}

\newenvironment{preliminary}%
  {\pagestyle{plain}\pagenumbering{roman}}%
  {\cleardoublepage}

\newenvironment{thesis}
{\pagestyle{fancy}\pagenumbering{arabic}\doublespacing}
{}

\newcommand{\floor}[1]{\lfloor #1 \rfloor}
\newcommand{\ceil}[1]{\lceil #1 \rceil}

\newcommand{\Cor}{\mbox{Cor}} 

\newcommand{\COR}{\mbox{COR}}

\newcommand{\mychoose}[2]{\Big(\begin{tabular}{c} $#1$ \\[-9pt] $#2$
\end{tabular}\Big)}

\newcommand{\given}{\ | \ }
\newcommand{\zi}{z^{(i)}}
\newcommand{\zz}{z^{(1)},\cdots,z^{(n)}}

\newcommand{\itb}{\begin{itemize}}
\newcommand{\itn}{\end{itemize}}
\newcommand{\nmn}{\end{enumerate}}
\newcommand{\nmb}{\begin{enumerate}}
\newcommand{\ctb}{\begin{center}}
\newcommand{\ctn}{\end{center}}
\newcommand{\tb}{\begin{center}\Large \bf}
\newcommand{\tn}{\end{center}}
\newcommand{\arb}{\begin{array}}
\newcommand{\qrn}{\end{array}}
\newcommand{\tblb}{\begin{tabular}}
\newcommand{\tbln}{\end{tabular}}
\newcommand{\fb}{\begin{figure}}
\newcommand{\fn}{\end{figure}}

\def\bern{\mbox{Bernoulli}\,}
\def\betad{\mbox{Beta}\,}

\def\ybar{\overline{y}}

\def\beq{\noindent \begin{eqnarray}}
\def\eeq{\end{eqnarray}}

\def\beqs {\noindent \begin{eqnarray*}}
\def\eeqs {\end{eqnarray*}}

\def\mb#1{\mbox{\boldmath $#1$}}

\def\A{\mathcal{A}}

\def\x{\mb x}
\def\z{\mb z}
\def\thetab{\mb \theta}
\def\phib{\mb \phi}

\def\betab{\mb \beta}

\def\IID{
 \,\begin{array}{cc} \\[-20pt] \mbox{\tiny IID} \\[-15pt] \sim \end{array}\,}

\def\trn{^{\mbox{\tiny train}}}
\def\nulu{^{\mbox{~}}}

\def \cc {\mbox{Cauchy}}

\def \lldots {\ \ldots \ }

\def \trainingdata {\mathcal{D}}

\def \P {\mathcal{P}}

\def \sigmaa {\Sigma_1}
\def \sigmab {\Sigma_2}

\def \s {s_g}
\def \st {s^t_g}

%% file: chapter0.tex

\begin{preliminary}

\setcounter{page}{1}

\renewcommand{\chaptername}{~}

\renewcommand{\thechapter}{~}

\thispagestyle{empty}

\begin{center}
\vspace*{0.5in}  
               {\Large \bf BAYESIAN CLASSIFICATION AND REGRESSION \\[0.25in]
                           WITH HIGH DIMENSIONAL FEATURES}

  			     \vspace*{1.3in}

                       {\Large \bf Longhai Li}\\
  
  			     \vspace*{1.5in}

                   \includegraphics[scale=1]{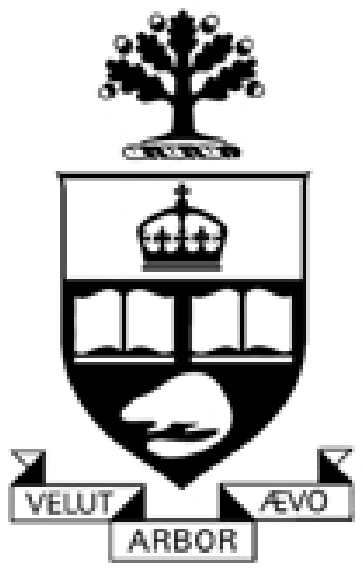}
 
         {\large 
         A thesis submitted in conformity with the requirements \\
                    for the degree of Doctor of Philosophy\\
                    Graduate Department of Statistics\\
                    University of Toronto\\
         } 
     				\vfill
 
		\copyright Copyright 2007 Longhai Li

\end{center}

\newpage


\thispagestyle{plain}

\begin{center}
  
  \textbf{ \Large Bayesian Classification  and  Regression \\ with High
    Dimensional Features}

		       \vspace*{1cm}

  		\textbf{\large Longhai Li}

  		       \vspace*{0.5cm}
  
      {\large Submitted for the Degree of Doctor of Philosophy}\\[0.2in] 

                           August 2007   

  			\vspace*{1cm}
  		 
		 \textbf{\large Abstract}
\end{center}

\doublespacing

\noindent This thesis responds to the challenges of using a large number, such as thousands, of features in regression and classification problems. 

There are two situations where such high dimensional features arise. One is when high dimensional measurements are available, for example, gene expression data produced by microarray techniques. For computational or other reasons, people may select only a small subset of features when modelling such data, by looking at how relevant the features are to predicting the response, based on some measure such as correlation with the response in the training data. Although it is used very commonly, this procedure will make the response appear more predictable than it actually is. In Chapter $2$, we propose a Bayesian method to avoid this selection bias, with application to naive Bayes models and mixture models. 

High dimensional features also arise when we consider high-order interactions.  The number of parameters will increase exponentially with the order considered. In Chapter $3$, we propose a method for compressing a group of parameters into a single one, by exploiting the fact that many predictor variables derived from high-order interactions have the same values for all the training cases. The number of compressed parameters may have converged before considering the highest possible order. We apply this compression method to logistic sequence prediction models and logistic classification models. 

We use both simulated data and real data to test our methods in both chapters.

\newpage


\thispagestyle{plain}

\chapter*{\vspace*{-1.6in}Acknowledgements}

\vspace*{-0.3in}

\ \ \ \ \ \ I can never overstate my gratitude to my supervisor Professor Radford Neal who guided me throughout the whole PhD training  period. Without his inspiration, confidence, and insightful criticism, I would have lost in the course of pursuing this degree. It has been my so far most valuable academic experience  to learn from him how to ponder problems, how to investigate them, how to work on them, and how to present the results.  

I would like to thank my PhD advisory and exam committee members --- Professors  Lawrence Brunner, Radu Craiu, Mike Evans, Keith Knight,  Jeffrey Rosenthal and Fang Yao. Their comments enhance this final presentation. Most of the aforementioned professors,  and in addition Professors Andrey Feuerverger, Nancy Reid, Muni Srivastava, and Lei Sun,  have taught me in various graduate courses. Much knowledge from them has become part of this thesis silently. 

I wish to specially thank my external appraiser, Professor Andrew Gelman. Many of his comments have greatly improved the previous draft of this thesis.

I am grateful to the support provided by the statistics department staff --- Laura Kerr, Andrea Carter, Dermot Wheland and Ram Mohabir. They have made the student life in this department so smooth  and enjoyable. 

I am indebted to my many student colleagues, who accompanied and shared knowledge with me. Special thanks go to Shelley Cao, Meng Du, Ana-Maria Staicu, Shuying Sun, Tao Wang, Jianguo Zhang, Sophia Lee, Babak Shahbab, Jennifer Listgarten, and many others.

I would like to thank my wife, Yehua Zhang. It would have been impossible to finish this thesis without her support and love.

I wish to thank my sister, Meiwen Li. She provided me with much support at the most difficult time to me.  

The last and most important thanks go to my parents, Baoqun Jie and Yansheng Li. They bore me, raised me and supported me. To them I dedicate this thesis.


\tableofcontents

\listoffigures

\listoftables

\end{preliminary}


%% file: chapter1.tex
\thispagestyle{fancy}

\setcounter{equation}{0}

\renewcommand{\chaptername}{Chapter}
\renewcommand{\thechapter}{1}

\chapter{Introduction}

\renewcommand{\chaptername}{Introduction} 

\section{Classification and Regression} \label{sec-intro-learning}

Methods for predicting a response variable $y$ given a set of features $\x=(x_1,\ldots,x_p)$ are needed in numerous scientific and industrial fields. A doctor wants to diagnose whether a patient has a certain kind of disease from some laboratory measurements on this patient; a post office wants to use a machine to recognize the digits and characters on envelopes; a librarian wants to classify documents using a pre-specified list of topics; a businessman wants to know how likely a person is to be interested in a new product according to this person's expenditure history; people want to know the temperature tomorrow given the meteorologic data in the past; etc. Many such problems can be summarized as finding a \textit{predictive function} $C$ linking the features $\x$ to a prediction for $y$:

\beq
\hat y = C(\x)
\eeq

\noindent The choice of function $C$ depends also on the choice of loss function one wishes to use in making a decision. In scientific discussion, we focus on finding a probabilistic predictive distribution:

\beq
P(y\given \x)
\eeq

\noindent Here, $P(y\given\x)$ could be either a probability density function for continuous $y$ (a regression model), or a probability mass function for discrete or categorical $y$ (a classification model). Given a loss function, one can derive the predictive function $C$ from the predictive distribution $P(y\given\x)$ by minimizing the average loss in the future. For example, when $y$ is continuous, if we use a squared loss function $L(\hat y,y) = (\hat y -y)^2$, the best guess of $y$ is the mean of $P(y\given \x)$; if we use an absolute loss function $L(\hat y,y) = |\hat y -y|$, the best guess is the median of $P(y\given \x)$; and when $y$ is discrete, if we use $0-1$ loss function $L(\hat y,y)= I(\hat y \not= y)$, the best guess is the mode of $P(y\given \x)$.

One approach to finding $P(y\given\x)$ is to learn from empirical data --- data on a number of subjects that have known values of the response and  values of features, denoted by $\{(y\sp{(1)},\x\sp{(1)}),\ldots,(y\sp{(n)},\x\sp{(n)})\}$, or collectively by $(y\trn,\x\trn)$. This is often called ``training'' data, and the subjects are called ``training'' cases, as we are going to use these data to ``train'' an initially ``unskilled'' predictive model, as discussed later. In contrast, a subject whose response and features are denoted by $(y^*,\x^*)$, for which we need to predict the response, is called a ``test'' case,  because we can use the prediction result to test how good a predictive model is if we are later given the true $y^*$.

There are many methods to learn from the training data (Hastie, Tibshirani and  Friedman 2001 and Bishop 2006). One may estimate $P(y^*\given\x^*)$ using  the empirical distribution of the responses in the neighbourhood  of $\x^*$ in some metric, as in the $k$-nearest-neighbourhood method. Such methods are called nonparametric methods. In this thesis, we consider parametric methods, in which we use a closed-form function with unknown parameters to model the data. Once the parameters are inferred from the training data we can discard the training data because we only need the parameters of the ``trained'' model for making predictions on test cases.

One class of parametric methods, called conditional modelling methods, start by defining $P(y\given\x)$ as a function involving some unknown parameters, denoted by $\thetab$. These parameters will be inferred from training data. For continuous $y$, the simplest and most commonly used form for $P(y\given\x)$ is a Gaussian model:

\beq
P(y\given \x,\betab,\sigma) = \frac{1}{\sqrt{2\pi}}\,
                \exp\left(-\frac{(y-f(\x,\betab))^2}{2\,\sigma^2}\right)
\eeq
 
\noindent  For a discrete $y$ that takes $K$ possible values $0,\ldots,K-1$, one
may use a logistic form for $P(y\given\x)$:

\beq
P(y=k\given \x,\thetab) = 
\frac{\exp( f_k(\x,\betab_k)) }
     {\sum_{j=0}^{K-1} \exp(f_j(\x,\betab_j)) }
\eeq

\noindent The function $f(\x,\betab)$ or functions $f_j(\x,\betab_j)$ link $\x$ to $y$. They are often linear functions of $\x$, but may be also nonlinear functions of $\x$ defined, for example, by multilayer perceptron networks. Our work in Chapter $3$ uses linear logistic models. 

Another class of methods model the joint distribution of $y$ and $\x$ by some formula with unknown parameters $\thetab$, written as $P(y,\x\given\thetab)$. The conditional probability $P(y\given \x,\thetab)$ can be found by:

\beq 
P(y\given \x,\thetab) = {P(y, \x \given \thetab)\over P(\x \given\thetab)}
\eeq

\noindent  Examples of such $P(y,\x,\thetab)$ include naive Bayes models, mixture models, Bayesian networks, and Markov random fields, etc., all of which use conditional independency in specifying $P(y,\x \given\thetab)$. For example naive Bayes models assume all features $\x$ are independent given $y$. Our work in Chapter $2$ uses naive Bayes models and mixture models. 

There are two generally applicable approaches for inferring $\thetab$ from the training data. One is to estimate $\thetab$ using a single value, $\hat \thetab$, that maximizes the likelihood function or a penalized likelihood function, i.e., the value that best fits the training data subject to some constraint. This single estimate will be plugged in to $P(y\given \x,\thetab)$ or $P(y,\x\given \thetab)$ to obtain the predictive distribution $P(y^*\given \x^*,\hat \thetab)$ for a test case. 

Alternatively, we can use a Bayesian approach, in which we first define a prior distribution, $P(\thetab)$, for $\thetab$, which reflects our ``rough'' knowledge about $\thetab$ before seeing the data, and then update our knowledge about $\thetab$ after we see the data, still expressed with a probability distribution, using Bayes formula:

\beq
P(\thetab \given y\trn,\x\trn) = 
\frac{P(y\trn,\x\trn \given \thetab)\,P(\thetab)}
     {P(y\trn,\x\trn)}
\eeq

\noindent $P(\thetab \given y\trn,\x\trn)$ is called the posterior distribution of $\thetab$. The joint distribution of a test case $(y^*,\x^*)$ given the training data $(y\trn,\x\trn)$ is found by integrating over $\thetab$ with respect to the posterior distribution:

\beq  P(y^*,\x^* \given y\trn,\x\trn)  = \int P(y^*,\x^* \given
y\trn,\x\trn,\thetab)P(\thetab \given y\trn,\x\trn)\,d\thetab \eeq 

The predictive distribution can then be found as $P(y^*,\x^* \given y\trn,\x\trn)/P(\x^*\given y\trn,\x\trn)$, which will be used to make predictions on test cases in conjunction with our loss function.

\section{Challenges of Using High Dimensional Features} \label{sec-ch-hd}

In many regression and classification problems, a large number of features are available for possible use. DNA microarray techniques can simultaneously measure  the expression levels of thousands of genes (Alon et.al. 1999, Khan et.al. 2001); the HIRIS instrument for the Earth Observing System generates image data in 192  spectral bands simultaneously (Lee and Landgrebe et.al. 1993);  one may consider numerous high-order interactions of discrete features; etc. 

There are several non-statistical difficulties in using high-dimensional features, depending on the purpose the data is used for. The primary one is computation time. Models for high dimensional data will require high dimensional parameters. Consequently, the time for training the model and making predictions on test cases may be intolerable. For example, a speech recognition program or data compression program  must be able to give out the prediction very quickly to be practically useful. Also, in some cases, measuring high dimensional features takes substantially more time or money. 

Serious statistical problems also arise with high dimensional features. When the number of features is larger than the number of training cases, the usual estimate of the covariance matrix of features is singular, and therefore can not be  used to compute the density  function. Regularization methods that shrink the estimation to a diagonal matrix have been proposed in the literature (Friedman 1998, Tadjudin and Landgrebe 1998, 1999). Such methods usually need to adjust some parameters that control the degree of shrinkage to a diagonal matrix, which may be difficult to determine. Another aspect of this problem is that even a simple model, such as a linear model, will overfit data with high dimensional features. Linear logistic models with the coefficients estimated by the maximum likelihood method will have some coefficients equal to $\infty$; the solution is also not unique. This is because the training cases can be divided by some hyperplanes in the space of features into groups such that all the cases with the same response are in a group; indeed, there are infinitely many such hyperplanes. The resulting classification rule works perfectly on the training data but may perform poorly on the test data. Overfitting problems usually arises because one uses more complex models than the data can support. For example, when one uses a polynomial function of degree $n$ to fit the relationship between $x$ and $y$ in $n$ data points $(y^{(i)},x^{(i)})$, there are infinitely many such polynomial functions that go exactly through each of these $n$ points. 

A sophisticated Bayesian method can overcome the overfitting problem by using a prior that favours simpler models. But unless one can analytically integrate with respect to the posterior distribution, implementating such a Bayesian method by Markov chain sampling is difficult. With more parameters, a Markov chain sampler will take longer for each iteration and require more memory. It may need more iterations to converge, or get trapped more easily in local modes. Also, with high dimensional features, it is harder to come up with a prior that reflects all of our knowledge of the problem. 

\section{Two Problems Addressed in this Thesis}

For the above reasons, people often use some methods to reduce the dimension of features before applying regression or classification methods. However, a simple implementation of such a ``preprocessing'' procedure may be invalid. For example, we may first select a small set of features that are most correlated with the response in the training data, then use these features to construct a predictive distribution. This procedure will make the response variable appear more predictable than it actually is. This overconfidence may be more pronounced when there are more features available, as more actually useless features will by chance pass the selection process, especially when very few useful features exist. In Chapter $2$, we propose a method to avoid this problem with feature selection in a Bayesian framework. In constructing the posterior distribution of parameters, we condition not only on the retained features, but also on the information that a number of features are discarded because of their weak correlations with the response. The key point in our solution is that we need only calculate the probability that one feature is discarded, then raise it to the power of the number of discarded features. We therefore can save much computation time by selecting only a very small number of features for use, and at the same time make well-calibrated predictions for test cases. We apply this method to naive Bayes models and mixture models for binary data.

A huge number of parameters will arise when we consider very high order interactions of discrete features. But many interaction patterns are expressed by the same training cases. In Chapter $3$, we use this fact to reduce the number of parameters by effectively compressing a group of parameters into a single one. After compressing the parameters, there are many fewer parameters involved in the Markov chain sampling. The original parameters can later be recovered efficiently by sampling from a splitting distribution. We can therefore consider very high order interactions in a reasonable amount of time. We apply this compression method to logistic sequence prediction models and logistic classification models.

\section{Comments on the Bayesian Approach} \label{sec-intro-bayesian}

The Bayesian approach is sometimes criticized for its use of prior distributions. Many people view the choice of prior as arbitrary because it is subjective.  The prior is the distribution of $\thetab$ that generates, through a defined sampling distribution, the class of data sets that will enter our analysis. Thus, there is only one prior that accurately defines the characteristics of the class of data sets, which may be described in another way, such as in words.  Different individuals may define different classes of data sets. The choice of prior is therefore subjective, but not arbitrary, since we may indeed decide that a prior distribution is wrong if the data sets it generates contradict our beliefs. Typically we choose a diffuse prior to include a wide class of data sets, but de-emphasize some data sets we believe less likely to appear in our analysis, for example a data set generated by a linear logistic model with coefficient equal to $10000$ for a binary feature. This distribution is therefore also phrased as expressing our prior belief, or our ``rough'' knowledge about which $\thetab$ may have generated our data set. 

There is usually useful prior information available for a problem before seeing any data set, such as relationships between the parameters (or data). For example, a set of body features of a human should be closer to those of a monkey than to other animals.  A sophisticated prior distribution can be used to capture such relationships.  For example, we can  assign the two groups of parameters, which are used to define the distribution of body features of a human and a monkey, a joint prior distribution in which they are positively correlated (Gelman, Bois and Jiang 1996). We usually construct such joint distributions by introducing some extra parameters that are shared by a group of parameters, which may also have meaningful interpretations. One way is to define the priors of the parameters of likelihood function in terms of some unknown hyperparameters, which is again given a higher level distribution. For example, in Automatic Relevance Determination (ARD) priors for neural network regression (Neal 1996), all the coefficients related to a feature are controlled by a common standard deviation. Such priors enable the models to decide whether a feature is useful automatically, through adjusting the posterior distribution of the common standard deviation. Similarly, in the priors for the models in Chapter $2$ we use a parameter $\alpha$ to control the overall degree of relationship between the features and response. Our method for avoiding the bias from feature selection has the effect of adjusting the posterior distribution of $\alpha$ to be closer to the right one (as would be obtained using the complete data), by conditioning on {\em all information known to us}, both the retained features and information about the feature selection process. Another way of introducing dependency is to express a group of parameters as the functions of a group of ``brick'' parameters. For example, in Chapter $3$, the regression coefficients for the highest order interaction patterns are expressed as sums of parameters representing the effects of lower order interaction patterns. Such priors enable the models to choose the orders automatically.

Once we have assigned an appropriate prior distribution for a problem, all forms of inference for unknown quantities, including the unknown parameters, can be carried out very straightforwardly in theory using only the rules of probability, since the result of inference is also expressed by a probability distribution. These predictions are found by averaging over all sets of values of $\thetab$ that are plausible in light of the training data. Compared with non-Bayesian methods, which use only a single set of parameters, the Bayesian approach has the following advantages from a practical viewpoint. 

First, the prediction is automatically accompanied by information on its uncertainty in making predictions, since the prediction is expressed by a probability distribution. 

Second, Bayesian prediction may be better than prediction based on only a single set of parameters. If the set of parameters that best explains the training data, such the MLE, is not the true set of parameters that generates the training data, we still have the chance to make good predictions, since the true set of parameters should be plausible given the training data and therefore will be considered as well in Bayesian prediction. 

Third, sophisticated Bayesian models, as described earlier, will self-adjust the complexity of a model in light of the data. We can define a model through a diffuse prior that can cover a wide class of data sets, from those with a low level of complexity to those with a high level of complexity. If the training data does not favour the high complexity, the posterior distribution will choose to use the simple model. In theory we do not need to change the complexity of a model according to the properties of the data, such as the number of observations. The overfitting problem in applying a complex model to a data set of small size is therefore overcome in Bayesian framework. Although more complex models may make the computation harder, Bayesian methods are, at least, much less sensitive to the choice of model complexity level than non-Bayesian methods.

Bayesian inference, however, is difficult to carry out, primarily for computational reasons. The posterior distribution is often on a high dimensional space, often  takes a very complicated form, and may have a lot of isolated modes. Markov chain Monte Carlo (MCMC) methods (Neal 1993, Liu 2001 and the references therein) are so far the only feasible methods to draw samples from a posterior distribution (Tierney 1994).  In the next section, we will briefly introduce these methods. However, for naive Bayes models in Chapter $2$ we do not use MCMC, due to the simplicity of naive Bayes models.

\section{Markov Chain Monte Carlo Methods} \label{sec-intro-mcmc}

We can simulate a Markov chain governed by a transition distribution $T(\thetab'\given \thetab)$ to draw samples from a distribution $\pi(\thetab)$, where $\thetab\in S$, if $T$ leaves $\pi$ invariant:

\beq \int_S \pi(\thetab)T(\thetab'\given \thetab)\,d\thetab = \pi(\thetab') 
\label{condition-invariant}
\eeq

\noindent and satisfies the following conditions: the Markov chain should be aperiodic, i.e., it does not explore the space in a cyclic way, and the Markov chain should be irreducible, i.e., the Markov chain can explore the whole space starting from any point. Given these conditions, it can be shown there is only one distribution $\pi$ satisfying the invariance condition~(\ref{condition-invariant}) for a Markov chain transition $T$ if there is one. (The condition of aperiodicity is not actually required for Monte Carlo estimation, but it is convenient in practice if a Markov chain is aperiodic, since we have more freedom in choosing the iterations for making Monte Carlo estimation. And it is obviously required to ensure that the result in~(\ref{eqn-convergence-mcmc}) is true.)

Let us denote a Markov chain by $\thetab^{(0)},\thetab^{(1)},\ldots$ . (Roberts and Rosenthal 2004) shows that if a Markov chain transition $T$ satisfies all the above conditions with respect to $\pi$, then starting from any point $\thetab_0$ for $\thetab^{(0)}$, the distribution of $\thetab^{(n)}$ will converge to $\pi$:

\beq \lim_{n->\infty} P(\thetab^{(n)} = \thetab\given \thetab^{(0)} =
\thetab_0)  = \pi(\thetab), \ \ \ \ \ \ \mbox{for any}\ \ \thetab,\thetab_0\in
S  \label{eqn-convergence-mcmc} \eeq

In words, after we run a Markov chain sufficiently long, the distribution of $\mb \theta^{(n)}$ (regardless the starting point) will be close to the target distribution $\pi(\mb \theta)$ in some metric (Rosenthal 1995 and the references therein). We can therefore use the states afterward as samples from $\pi(\mb \theta)$ (though correlated) for making Monte Carlo estimations. It is tremendously difficult to determine in advance how long we should run for an arbitrary Markov chain, though we can do this for some types of Markov chains (Rosenthal 1995). In practice we check the convergence by running multiple chains starting from different points and see whether they have mixed at a certain time (see for example Cowles and Carlin 1996, and the references therein).

It is usually not difficult to construct a Markov chain transition $T$ that satisfies the invariance condition for a desired distribution $\pi$ and the other two conditions as well, based on the following facts. First, one can show that a Markov chain transition $T$ leaves $\pi$ invariant if it is reversible with respect to $\pi$:

\beq \pi(\thetab)\,T(\thetab'\given \thetab) = \pi(\thetab')\,T(\thetab\given
\thetab'), \ \ \ \ \ \ \mbox{for any}\ \ \thetab',\thetab \in S
\label{condition-rev} \eeq

\noindent It therefore suffices to devise a Markov chain that is reversible with respect to $\pi$. Second, applying a series of Markov chain transition $T_i$ that have been shown to leave $\pi$ invariant will also leave $\pi$ invariant. Also, applying a series of appropriate Markov chain transition $T_i$ that explores only a subset of $S$ can explore the whole space, $S$.

Gibbs sampling method (Geman and Geman 1984, and Gelfand and  Smith 1990) and the Metropolis-Hastings method (Metropolis et. al. 1953, and Hastings 1970) are two basic methods to devise a Markov chain transition that leaves $\pi$ invariant. We usually use a combination of them to devise a Markov chain transition satisfying the above conditions for a complicated target distribution $\pi$.  

Let us write $\thetab=(\theta_1,\ldots,\theta_p)$.  Gibbs sampling defines the transition from $\thetab^{(t-1)}$ to $\thetab^{(t)}$ as follows:

\vspace{0.1in}

\hspace*{1.5in}\begin{tabular}{l}

Draw $\theta_1^{(t)}$ from $\pi(\theta_1 \given
\theta_2^{(t-1)},\ldots,\theta_p^{(t-1)})$ \\

Draw $\theta_2^{(t)}$ from $\pi(\theta_2 \given
\theta_1^{(t)},\theta_3^{(t-1)},\ldots,\theta_p^{(t-1)})$\\

\vdots \\

Draw $\theta_i^{(t)}$ from $\pi(\theta_i \given
\theta_1^{(t)},\ldots,\theta_i^{(t)},\theta_{i+1}^{(t-1)},\ldots,\theta_p^{(t-1)})$\\

\vdots\\

Draw $\theta_p^{(t)}$ from $\pi(\theta_p \given
\theta_1^{(t)},\ldots,\theta_{p-1}^{(t)})$

\end{tabular}

\vspace{0.1in}

\noindent The order of updating  $\theta_i$ can be any permutation  of $1,\ldots,p$. One can show each updating of $\theta_i$ is reversible with respect to $\pi(\thetab)$, and a complete updating of all $\theta_i$ therefore leaves $\pi(\thetab)$ invariant.  Sampling from the conditional distribution for $\theta_i$  can also be replaced with any transition that leaves the conditional distribution  invariant, for example, a Metropolis-Hastings transition as described next. 

The Metropolis-Hastings method first samples from a proposal distribution $\hat T(\thetab^*\given  \thetab^{(t-1)})$ to propose a candidate $\thetab^*$, then draws a random number $U$ from the uniform distribution over $(0,1)$.  If 

\beq U < \min\left(1,\ \frac{\pi(\thetab^*)\,\hat T(\thetab^{(t-1)}\given
\thetab^*)} {\pi(\thetab^{(t-1)})\,\hat T(\thetab^*\given \thetab^{(t-1)})}
\right), \eeq

\noindent we let $\thetab^{(t)}=\thetab^*$, otherwise we let $\thetab^{(t)}=\thetab^{(t-1)}$. One can show that such a transition is reversible with respect to $\pi$, and hence leave $\pi$ invariant.

\section{\vspace*{-0.1in}Outline of the Remainder of the Thesis}

We will discuss in detail our method for avoiding bias from feature selection in Chapter $2$, with application to naive Bayes models and  mixture models. In Chapter $3$ we discuss how to compress the parameters in Bayesian regression and classification models with high-order interactions, with application to logistic sequence prediction models and to logistic classification models. We conclude separately at the end of each chapter.


%% file: chapter2.tex
\renewcommand{\chaptername}{Chapter}
\renewcommand{\thechapter}{2}

\chapter{Avoiding Bias from Feature Selection}\label{chapter-bias}

\renewcommand{\chaptername}{Avoiding Bias from Feature Selection}

\doublespacing

{\bf Abstract.} For many classification and regression problems, a large number of features are available for possible use --- this is typical of DNA microarray data on gene expression, for example.  Often, for computational or other reasons, only a small subset of these features are selected for use in a model, based on some simple measure such as correlation with the response variable. This procedure may introduce an optimistic bias, however, in which the response variable appears to be more predictable than it actually is, because the high correlation of the selected features with the response may be partly or wholly due to chance.  We show how this bias can be avoided when using a Bayesian model for the joint distribution of features and response.  The crucial insight is that even if we forget the exact values of the unselected features, we should retain, and condition on, the knowledge that their correlation with the response was too small for them to be selected. In this paper we describe how this idea can be implemented for ``naive Bayes'' and mixture models of binary data. Experiments with simulated data confirm that this method avoids bias due to feature selection.  We also apply the naive Bayes model to subsets of data relating gene expression to colon cancer, and find that correcting for bias from feature selection does improve predictive performance. 

\footnotetext[1]{Part of this Chapter appeared as a technical report coauthored
with Jianguo Zhang and Radford Neal.}

\newpage

\section{Introduction}\label{sec-intro}

Regression and classification problems that have a large number of available ``features'' (also known as ``inputs'', ``covariates'', or ``predictor variables'') are becoming increasingly common.  Such problems arise in many application areas.  Data on the expression levels of tens of thousands of genes can now be obtained using DNA microarrays, and used for tasks such as classifying tumors.  Document analysis may be based on counts of how often each word in a large dictionary occurs in each document.  Commercial databases may contain hundreds of features describing each customer.

Using all the features available is often infeasible. Using too many features can result in ``overfitting'' when simple statistical methods such as maximum likelihood are used, with the consequence that poor predictions are made for the response variable (e.g., the class) in new items.  More sophisticated Bayesian methods can avoid such statistical problems, but using a large number of features may still be undesirable.  We will focus primarily on situations where the computational cost of looking at all features is too burdensome. Another issue in some applications is that using a model that looks at all features will require measuring all these features when making predictions for future items, which may sometimes be costly.  In some situations, models using few features may be preferred because they are easier to interpret.

For the above reasons, modellers often use only a subset of features, chosen by some simple indicator of how useful they might be in predicting the response variable --- see, for example, the papers in (Guyon, \textit{et al.} 2006).  For both regression problems with a real-valued response variable and classification problems with a binary (0/1) class variable, one suitable measure of how useful a feature may be is the sample correlation of the feature with the response.  If the absolute value of this sample correlation is small, we might decide to omit the feature from our model.  This criterion is not perfect, of course --- it may result in a relevant feature being ignored if its relationship with the response is non-linear, and it may result in many redundant features being retained even when they all contain essentially the same information.  Sample correlation is easily computed, however, and hence is an attractive criterion for screening a large number of features.

Unfortunately, a model that uses only a subset of features, selected based on their high correlation with the response, will be optimistically biased --- i.e., predictions made using the model will (on average) be more confident than is actually warranted.  For example, we might find that the model predicts that certain items belong to class~1 with probability 90\%, when in fact only 70\% of these items are in class~1.  In a situation where the class is actually completely unpredictable from the features, a model using a subset of features that purely by chance had high sample correlation with the class may produce highly confident predictions that have less actual chance of being correct than just guessing the most common class. The feature selection bias has also been noticed in the literature by a few researchers, see for example, the papers (Ambroise and McLachlan 2002), (Lecocke and Hess 2004), (Singhi and Liu 2006), and (Raudys, Baumgartner and Somorjai 2005). They pointed out that if the feature selection is performed externally to the cross-validation assessment (ie, cross-validation is applied to a subset of features selected in advance based on all observations), the classification error rate will be highly underestimated (could be 0\%). It is therefore suggested that feature selection should be performed internally to the cross-validation procedure, ie, re-selecting features whenever the training set and test set are changed. This modified cross-validation procedure avoids underestimating the error rate and assesses properly the predictive method plus the feature selection method. However, it does not provide a scheme for constructing a better predictive method that can give out well-calibrated predictive probabilities for test cases. We propose a Bayesian solution to this problem.

This optimistic bias comes from ignoring a basic principle of Bayesian inference --- that we should base our conclusions on probabilities that are conditional on \textit{all} the available information.  If we have an appropriate model, this principle would lead us to use all the features.  This would produce the best possible predictive performance.  However, we assume here that computational or other pragmatic issues make using all features unattractive.  When we therefore choose to ``forget'' some features, we can nevertheless still retain the information about how we selected the subset of features that we use in the model.  Properly conditioning on this information when forming the posterior distribution eliminates the bias from feature selection, producing predictions that are as good as possible given the information in the selected features, without the overconfidence that comes from ignoring the feature selection process.

We can use the information from feature selection procedure only when we model the features and the response jointly. We show in this Chapter this information can be easily incorporated into our inference in a Bayesian framework. We particularly apply this method to naive Bayes models and mixture models.

\section{Our Method for Avoiding Selection Bias}
\label{sec-idea}

Suppose we wish to predict a response variable, $y$, based on the information in the numerical features $x_1,\ldots,x_p$, which we sometimes write as a vector, $\x$.  Our method is applicable both when $y$ is a binary ($0/1$) class indicator, as is the case for the naive Bayes models discussed later, and when $y$ is real-valued.  We assume that we have complete data on $n$ ``training'' cases, for which the responses are $y^{(1)},\ldots,y^{(n)}$ (collectively written as $y\trn$) and the feature vectors are $\x^{(1)},\ldots,\x^{(n)}$ (collectively written as $\x\trn$).  (Note that when $y$, $\x$, or $x_t$ are used without a superscript, they will refer to some unspecified case.)  We wish to predict the response for one or more ``test'' cases, for which we know only the feature vector.  Our predictions will take the form of a distribution for $y$, rather than just a single-valued guess.

We are interested in problems where the number of features, $p$, is quite big --- perhaps as large as ten or a hundred thousand --- and accordingly (for pragmatic reasons) we intend to select a subset of features based on the absolute value of each feature's sample correlation with the response.  The sample correlation of the response with feature $t$ is defined as follows (or as zero if the denominator below is zero):\vspace*{-13pt} 

\beq
  \COR (y\trn,\,x\trn_t) & = &
  {\displaystyle\sum_{i=1}^n\, \big(y^{(i)}-\bar y\big)\,
                                 \big(x^{(i)}_t-\bar x_t\big)
  \over
   \sqrt{\sum\limits_{i=1}^n \big(y^{(i)}-\bar y\big)^2}\
   \sqrt{\sum\limits_{i=1}^n \big(x^{(i)}_t-\bar x_t\big)^2} }
\label{eq-cor-def}
\eeq

\noindent where $\bar y\, =\, {1 \over n}\sum\limits_{i=1}^n y^{(i)}$ and $\bar x_t\, =\,
{1 \over n}\sum\limits_{i=1}^n x^{(i)}_t$.  The numerator can be simplified to $\sum\limits_{i=1}^n \big(y^{(i)}-\bar y\big) x^{(i)}_t$.

Although our interest is only in predicting the response, we assume
that we have a model for the joint distribution of the response
together with all the features.  From such a joint distribution, with
probability or density function $P(y,x_1,\ldots,x_p)$, we can obtain
the conditional distribution for $y$ given any subset of features, for
instance $P(y\,|\,x_1,\ldots,x_k)$, with $k<p$.  This is the
distribution we need in order to make predictions based on this
subset.  Note that selecting a subset of features makes sense only
when the omitted features can be regarded as random, with some
well-defined distribution given the features that are retained, since
such a distribution is essential for these predictions to be meaningful.
This can be seen from the following expression:
\beq
  P(y\,|\,x_1,\ldots,x_k)
  & = &
        \int \cdots \int P(y\,|\,x_1,\ldots,x_k,x_{k+1},\ldots,x_p) \cdot \nonumber \\
  &  &  \ \ \ \ \ \ \ \ \ \ \  P(x_{k+1},\ldots,x_p\,|\,x_1,\ldots,x_k)\
                   dx_{k+1} \cdots dx_p  
\eeq
If $P(x_{k+1},\ldots,x_p\,|\,x_1,\ldots,x_k)$ does not exist in any
meaningful sense --- as would be the case, for example, if the data
were collected by an experimenter who just decided arbitrarily what to
set $x_{k+1},\ldots,x_p$ to --- then $P(y\,|\,x_1,\ldots,x_k)$ will
also have no meaning.

Consequently, features that cannot usefully be regarded as random
should always be retained.  Our general method can accommodate such
features, provided we use a model for the joint distribution of the
response together with the random features, conditional on given
values for the non-random features.  However, for simplicity, we will
ignore the possible presence of non-random features in this paper.

We will assume that a subset of features is selected by fixing a
threshold, $\gamma$, for the absolute value of the correlation of a
selected feature with the response.  We then omit feature $t$ from the
feature subset if $|\COR (y\trn,x\trn_t)|\,\le\,\gamma$, retaining
those features with a greater degree of correlation.  Another possible
procedure is to fix the number of features, $k$, that we wish to
retain, and then choose the $k$ features whose correlation with the
response is greatest in absolute value, breaking any tie at random.
If $s$ is the retained feature with the weakest correlation with the
response, we can set $\gamma$ to $|\COR (y\trn,x\trn_s)|$, and we will
again know that if $t$ is any omitted feature, $|\COR
(y\trn,x\trn_t)|\,\le\,\gamma$.  If either the response or the
features have continuous distributions, exact equality of sample
correlations will have probability zero, and consequently this
situation can be treated as equivalent to one in which we fixed
$\gamma$ rather than $k$.  If sample correlations for different
features can be exactly equal, we should theoretically make use of the
information that any possible tie was broken the way that it was, but
ignoring this subtlety is unlikely to have any practical effect, since
ties are still likely to be rare.

Regardless of the exact procedure used to select features, we will
denote the number of features retained by $k$, we will renumber the
features so that the subset of retained features is $x_1,\ldots,x_k$,
and we will assume we know that $|\COR (y\trn,x\trn_t)|\,\le\,\gamma$
for $t=k\!+\!1,\ldots,p$.

We can now state the basic principle behind our bias-avoidance
method:\ \ When forming the posterior distribution for parameters of
the model using a subset of features, we should condition not only on
the values in the training set of the response and of the $k$ features
we retained, but also on the fact that the other $p\!-\!k$ features
have sample correlation with the response that is less than $\gamma$ 
in absolute value.  That
is, the posterior distribution should be conditional on the following
information:\vspace*{-4pt}
\beq
   y\trn,\ \ \x\trn_{1:k},\ \ |\COR (y\trn,x\trn_t)|\,\le\,\gamma\
                              \mbox{for $t = k\!+\!1,\ldots,p$}
\label{eq-cond-info}
\eeq
where $\x\trn_{1:k}\, =\, (x\trn_1,\ldots,x\trn_k)$.

We claim that this procedure of conditioning on the fact that selection occurred
will eliminate the bias from feature selection. Here, ``bias'' does not refer to
estimates for model parameters, but rather to our estimate of how well we can
predict responses in test cases.  Bias in this respect is referred to as a lack
of ``calibration'' --- that is, the predictive probabilities do not represent
the actual chances of events (Dawid 1982).  If the model describes the actual
data generation mechanism, and the actual values of the model parameters are
indeed randomly chosen according to our prior, Bayesian inference always
produces well-calibrated results, on average with respect to the data and 
model parameters generated from the Bayesian model. The proof that the Bayesian 
inference is well-calibrated is given in the Appendix $1$ to this Chapter.

In justifying our claim that this procedure avoids selection bias (ie, is
well-calibrated), we will assume that our model for the joint distribution of
the response and all features, and the prior we chose for it, are appropriate
for the problem, and that we would therefore not see bias if we predicted the
response using all the features.  Now, imagine that rather than selecting a
subset of features ourselves, after seeing all the data, we instead set up an
automatic mechanism to do so, providing it with the value of $\gamma$ to use as
a threshold.  This mechanism, which has access to all the data, will compute the
sample correlations of all the features with the response, select the subset of
features by comparing these sample correlations with $\gamma$, and then erase
the values of the omitted features, delivering to us only the identities of the
selected features and their values in the training cases.  If we now condition
on all the information that \textit{we} know, but not on the information that
was available to the selection mechanism but not to us, we will obtain unbiased
inferences.  The information we know is just that of (\ref{eq-cond-info}) above.

\begin{figure}

\centerline{\psfig{file=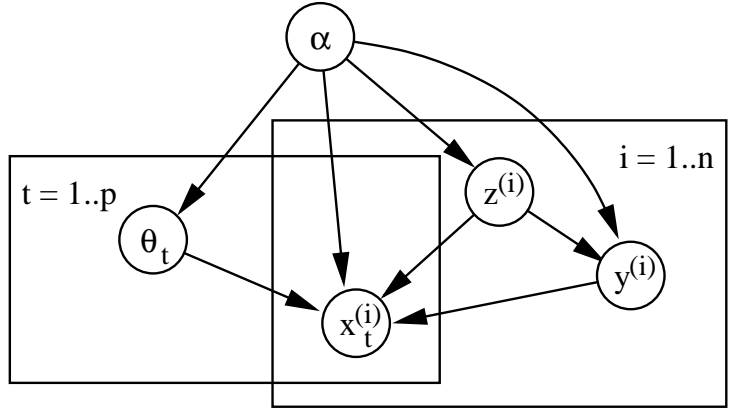}}

\caption[A directed graphical model for the general class of models]{A directed
graphical model for the general class of models we are considering.  Circles
represent variables, parameters, or hyperparameters.  Arrows represent possible
direct dependencies (not all of which are necessarily present in all models in
this class). The rectangles enclose objects that are repeated; an object in both
rectangles is repeated in both dimensions.  The case index, $i$, is shown as
ranging over the $n$ training cases, but test cases (not shown) belong in this
rectangle as well.  This diagram portrays a model where cases are independent
given $\alpha$ and $\thetab$, though this is not essential.}\label{fig-gr-model}

\end{figure}

The class of models we will consider in detail may include a vector of
latent variables, $\z$, for each case.  Model parameters
$\theta_1,\ldots,\theta_p$ (collectively denoted $\thetab$) are associated
with the $p$ features; other parameters or hyperparameters, $\alpha$, not
associated with particular features, may also be present.  Conditional on
$\thetab$ and $\alpha$, the different cases may be independent, though this
is not essential for our method.   Our method does rely on the values of
different features (in all cases) being independent, conditional on
$\thetab$, $\alpha$, $y\trn$, and $\z\trn$.  Also, in the prior
distribution for the parameters, $\theta_1,\ldots,\theta_p$ are assumed to
be conditionally independent given $\alpha$.  These conditional
independence assumptions are depicted graphically in
Figure~\ref{fig-gr-model}.

If we retain all features, our prediction for the response, $y^*$, in
a test case for which we know the features, $\x^* =
(x^*_1,\ldots,x^*_p)$, can be found from the joint predictive
distribution for $y^*$ and $\x^*$ given the data for all training
cases, written as $y\trn$ and $\x\trn$:
\beq
  P(y^*\,|\,\x^*,\,y\trn,\,\x\trn) & = &
  { P(y^*,\,\x^*\,|\,y\trn,\,\x\trn)\ \over P(\x^*\,|\,y\trn,\,\x\trn) } \\[5pt]
  & = &
  {\int \int P(y^*,\,\x^*\,|\,\alpha,\,\thetab)\,
     P(\alpha,\,\thetab\,|\,y\trn,\,\x\trn)\,d\alpha\,d\thetab \over
   \int \int P(\x^*\,|\,\alpha,\,\thetab)\,
     P(\alpha,\,\thetab\,|\,y\trn,\,\x\trn)\,d\alpha\,d\thetab}
\eeq
The posterior, $P(\alpha,\,\thetab\,|\,y\trn,\,\x\trn)$,
is proportional to the product of the prior and the likelihood:
\beq
  P(\alpha,\,\thetab\,|\,y\trn,\,\x\trn) & \propto &
    P(\alpha,\,\thetab)\ P(y\trn,\,\x\trn\,|\,\alpha,\,\thetab) \\[5pt]
  & \propto &
    P(\alpha)\ \prod_{t=1}^p P(\theta_t\,|\,\alpha)\
    \prod_{i=1}^n P(y^{(i)},\,\x^{(i)}\,|\,\alpha,\,\thetab)
\eeq
where the second expression makes use of the conditional independence
properties of the model.

When we use a subset of only $k$ features, the predictive distribution
for a test case will be
\beq
  \lefteqn{P(y^*\,|\,\x^*_{1:k},\,y\trn,\,\x\trn_{1:k},\,{\cal S})} \nonumber \\
  & = &
  {\int \int P(y^*,\,\x^*_{1:k}\,|\,\alpha,\,\thetab\nulu_{1:k})\,
      P(\alpha,\,\thetab\nulu_{1:k}\,|\,y\trn,\,\x\trn_{1:k},\,{\cal S})\,
      d\alpha\,d\thetab\nulu_{1:k}
   \over
   \int \int P(\x^*_{1:k}\,|\,\alpha,\,\thetab\nulu_{1:k})\,
      P(\alpha,\,\thetab\nulu_{1:k}\,|\,y\trn,\,\x\trn_{1:k},\,{\cal S})\,
      d\alpha\,d\thetab\nulu_{1:k}}\ \
\label{eq-pred}\eeq
where ${\cal S}$ represents the information regarding selection from
(\ref{eq-cond-info}), namely $|\COR (y\trn,x\trn_t)|\,\le\,\gamma$
for $t = k\!+\!1,\ldots,p$.  The posterior distribution for
$\alpha$ and $\thetab_{1:k}$ needed for this prediction can be written as
follows, in terms of an integral (or sum) over the values of the latent
variables, $\z\trn$:
\beq
  \lefteqn{P(\alpha,\,\thetab\nulu_{1:k}\,|\,y\trn,\,\x\trn_{1:k},\,{\cal S})}
  \ \ \ \ \ \nonumber\\[5pt]
  & \propto &
    \int P(\alpha,\,\thetab\nulu_{1:k},\,\z\trn\,|\,
           y\trn,\,\x\trn_{1:k},\,{\cal S})
    \,d\z\trn \\[5pt]
  & \propto & \int
    P(\alpha,\,\thetab\nulu_{1:k})\
    P(\z\trn,\,y\trn,\,\x\trn_{1:k}\,|\,\alpha,\,\thetab\nulu_{1:k})\cdot \nonumber \\
  & &
    \ \ \ 
    P({\cal S}\,|\,\alpha,\,\thetab\nulu_{1:k},\,\z\trn,\,y\trn,\,\x\trn_{1:k})
    \,d\z\trn \ \ \ \ \\[5pt]
  & \propto & \int
    P(\alpha,\,\thetab\nulu_{1:k})\
    P(\z\trn,\,y\trn,\,\x\trn_{1:k}\,|\,\alpha,\,\thetab\nulu_{1:k})\
    P({\cal S}\,|\,\alpha,\,\z\trn,\,y\trn)
    \,d\z\trn
\eeq
Here again, the conditional independence properties of the model
justify removing the conditioning on $\thetab\nulu_{1:k}$ and
$\x\trn_{1:k}$ in the last factor.

Computation of $P({\cal S}\,|\,\alpha,\,\z\trn,\,y\trn)$, which
adjusts the likelihood to account for feature selection, is crucial
to applying our method.  Two facts greatly ease this computation.
First, the $x_t$ are conditionally independent given $\alpha$, $\z$,
and $y$, which allows us to write this as a product of factors
pertaining to the various omitted features.  Second, these factors are
\textit{all the same}, since nothing distinguishes one omitted feature
from another.  Accordingly,
\beq
    P({\cal S}\,|\,\alpha,\,\z\trn,\,y\trn) & = &
    \!\!\prod_{t=k+1}^p\!\! P\big(|\COR (y\trn,x\trn_t)|\,\le\,\gamma
                              \,|\,\alpha,\,\z\trn,\,y\trn) \\[5pt]
    & = &
    \Big[ P\big(|\COR (y\trn,x\trn_t)|\,\le\,\gamma
                              \,|\,\alpha,\,\z\trn,\,y\trn) \Big]^{p-k}
\label{eq-cor-lik}\eeq
where in the second expression, $t$ represents \textit{any} of the
omitted features.  Since the time needed to compute the adjustment
factor does not depend on the number of omitted features, we may hope
to save a large amount of computation time by omitting many features.

Computing the single factor we do need is not trivial, however, since
it involves integrals over $\theta_t$ and $x\trn_t$.  We can write
\beq
  \lefteqn{
  P\big(|\COR (y\trn,x\trn_t)|\,\le\,\gamma\,|\,\alpha,\,\z\trn,\,y\trn)}
  \ \ \ \ \ \ \nonumber\\[5pt]
  & = &
  \int P(\theta_t\,|\,\alpha)\, P\big(|\COR (y\trn,x\trn_t)|\,\le\,\gamma
        \,|\,\alpha,\,\theta_t,\,\z\trn,\,y\trn)\,d\theta_t
\label{adjust-factor}
\eeq
Devising ways of efficiently performing this integral over $\theta_t$
and the integral over $x\trn_t$ implicit in the probability statement
occurring in the integrand will be the main topic of our discussion of
specific models below.

Once we have a way of computing this factor, we can use standard Markov chain
Monte Carlo (MCMC) methods to sample from
$P(\alpha,\,\thetab\nulu_{1:k},\,\z\trn\,|\, y\trn,\,\x\trn_{1:k},\,{\cal S})$. 
The resulting sample of values for $\alpha$ and $\thetab\nulu_{1:k}$ can be used
to make predictions using equation~(\ref{eq-pred}), by approximating the
integrals in the numerator and denominator by Monte Carlo estimates.  For the
naive Bayes model we will discuss in Section~\ref{sec-bnaive},  however, Monte
Carlo methods are unnecessary --- a combination of analytical integration and
numerical quadrature is faster.

\section{Application to Bayesian Naive Bayes Models} \label{sec-bnaive} 

In this chapter we show how  to apply the bias correction method to Bayesian naive Bayes models in which both the features and the response are binary.  Binary features are natural for some problems (e.g., test answers that are either correct or incorrect), or may result from thresholding real-valued features.  Such thresholding can sometimes be beneficial --- in a document classification problem, for example, whether or not a word is used at all may be more relevant to the class of the document than how many times it is used. Naive Bayes models assume that features are independent given the response.  This assumption is often incorrect, but such simple naive Bayes models have nevertheless been found to work well for many practical problems (see for example Li and Jain 1998,  Vaithyanathan, Mao, and Dom  2000, Eyheramendy, Lewis, and Madigan 2003).  Here we show how to correct for selection bias in binary naive Bayes models, whose simplicity allows the required adjustment factor to be computed very quickly.  Simulations reported in Section~\ref{sec-sim-bnaive} show that substantial bias can be present with the uncorrected method, and that it is indeed corrected by conditioning on the fact that feature selection occurred.  We then apply the method to real data on gene expression relating to colon cancer, and again find that our bias correction method improves predictions.

\subsection{Definition of the Binary Naive Bayes Models}\label{sub-model-b}

\begin{figure}[t]

\begin{center}

\includegraphics[scale=0.7]{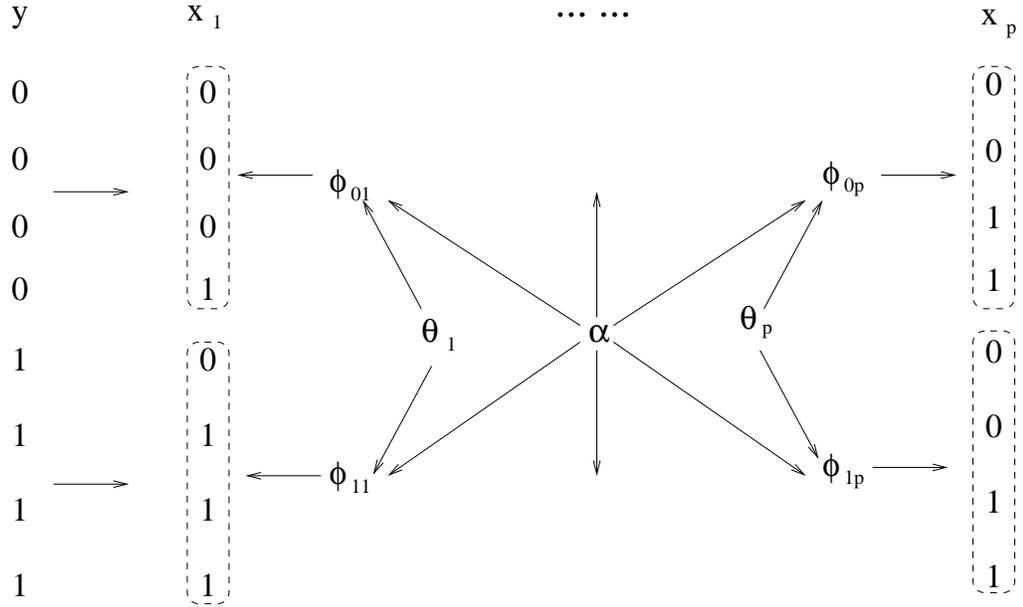}

\caption{A picture of Bayesian naive Bayes models.}

\label{fig-naive}

\end{center}

\end{figure}

Let $\x^{(i)}=(x^{(i)}_1,\cdots,x^{(i)}_p)$ be the vector of $p$ binary features
for case $i$, and let $y^{(i)}$ be the binary response for case $i$, indicating
the class.  For example, $y^{(i)}=1$ might indicates that cancer is present for
patient $i$, and $y^{(i)}=0$ indicate that cancer is not present.  Cases are
assumed to be independent given the values of the model parameters (ie,
exchangeable \textit{a priori}).  The probability that $y=1$ in a case is given
by the parameter $\psi$.  Conditional on the class $y$ in some case (and on the
model parameters), the features $x_1,\ldots,x_p$ are assumed to be independent,
and to have Bernoulli distributions with parameters
$\phi_{y,1},\ldots,\phi_{y,p}$, collectively written as $\phib_y$, with
$\phib=(\phib_0,\phib_1)$ representing all such parameters.
Figure~\ref{fig-naive} displays the models. Formally, the data is modeled as

\beq
   y^{(i)}\ |\ \psi & \sim & \bern(\psi),\ \ \ \mbox{for $i=1,\ldots,n$}
   \label{sample-y-b}
\\[4pt]
   x^{(i)}_j\ |\ y^{(i)},\, \phib & \sim & \bern(\phi_{y^{(i)},j}),
   \ \ \ \mbox{for $i=1,\ldots,n$ and $j=1,\ldots,p$}
   \label{sample-x-b}
\eeq

We use a hierarchical prior that expresses the possibility that some features
may have almost the same distribution in the two classes.  In detail, the prior
has the following form:

\beq
 \psi & \sim & \betad(f_1,f_0) \label{prior-psi-b} \\[4pt]
 \alpha & \sim & \mbox{Inverse-Gamma}(a,b)\label{prior-alpha-b}
 \\[4pt]
 \theta_1,\ldots,\theta_p &\IID & \mbox{Uniform}(0,1)
   \label{prior-theta-b}
 \\[4pt]
 \phi_{0,j},\,\phi_{1,j}\ |\ \alpha,\,\theta_j & \IID &
   \betad(\alpha\theta_j,\,\alpha(1\!-\!\theta_j)),
   \ \ \ \mbox{for $j=1,\ldots,p$}
 \label{prior-phi-b}
\eeq

The hyperparameters $\thetab\, =\, (\theta_1,\ldots,\theta_p)$ are used to
introduce dependence between $\phib_{0,j}$ and $\phib_{1,j}$, with $\alpha$
controlling the degree of dependence. Features for which $\phi_{0,j}$ and
$\phi_{1,j}$ differ greatly are more relevant to predicting the response.  When
$\alpha$ is small, the variance of the Beta distribution in~(\ref{prior-phi-b}),
which is  $\theta_j\,(1\!-\!\theta_j)\,/\,(\alpha\!+\!1)$, is large, and many
features are likely to have predictive power, whereas when $\alpha$ is large, it
is likely that most features will be of little use in predicting  the response,
since $\phi_{0,j}$ and $\phi_{1,j}$ are likely to be  almost equal. We chose an
Inverse-Gamma prior for $\alpha$ (with density function proportional to
$\alpha^{-(1+a)}\exp(-b/\alpha)$) because it has a heavy upward tail, allowing
for the possibility that $\alpha$ is  large.  Our method of correcting selection
bias will have the effect of modifying the likelihood in a way that favors
larger values for $\alpha$ than would result from ignoring the effect of
selection.

\subsection{Integrating Away $\psi$ and $\phib$}\label{sub-int-b}

Although the above model is defined with $\psi$ and $\phib$ parameters for
better conceptual understanding, computations are simplified by integrating
them away analytically.  

Integrating away $\psi$, the joint probability of
$y\trn = (y^{(1)},\ldots,y^{(n)})$ is as follows, 
where $I(\,\cdot\,)$ is the indicator function, equal to 1 if the enclosed 
condition is true and 0 if it is false:
\beq
  P(y\trn)
  & = & \int_0^1 {{\Gamma(f_0+f_1)}\over{\Gamma(f_0)\Gamma(f_1)}}
                \psi^{f_1}(1\,-\,\psi)^{f_0}\
        \psi^{\sum\limits_{i=1}^n I(y^{(i)}=1)}\,
        (1-\psi)^{\sum\limits_{i=1}^n I(y^{(i)}=0)}d\psi\\[3pt]
  & = & U\textstyle
  \Big(f_1,\,f_0,\,\sum\limits_{i=1}^n\, I\big(y^{(i)}=1\big),\,
                   \sum\limits_{i=1}^n\, I\big(y^{(i)}=0\big)\Big)
\label{py-b}
\eeq
The function $U$ is defined as\vspace*{-5pt}
\beq
  U(f_1,f_0,n_1,n_0)
  & = & { \Gamma(f_0+f_1) \over \Gamma(f_0)\Gamma(f_1)} \,
           { \Gamma(f_0+n_0)\Gamma(f_1+n_1) \over \Gamma(f_0+f_1+n_0+n_1)} \\
  & = & { \prod\limits_{\ell=1}^{n_0} (f_0+\ell-1)\,
     \prod\limits_{\ell=1}^{n_1} (f_1+\ell-1)
     \over \prod\limits_{\ell=1}^{n_0+n_1} (f_0+f_1+\ell-1)}
  \ \ \ \ \ \ \
\eeq
The products above have the value one when the upper limits of $n_0$ or $n_1$ 
are zero.  The joint probability of $y\trn$ and the response, $y^*$, for
a test case is similar:
\beq
  \lefteqn{P(y\trn,y^*)} \nonumber\\
& = & U\textstyle
  \Big(f_1,\,f_0,\,\sum\limits_{i=1}^n\, I\big(y^{(i)}=1\big)\,
                  +I(y^*=1),\,
                  \sum\limits_{i=1}^n\, I\big(y^{(i)}=0\big)+I(y^*=0)\Big)
\label{pystar-b}
\eeq
Dividing $P(y\trn,y^*)$ by $P(y\trn)$ gives 
\beq
   P(y^*\ |\ y\trn) & = &
   \bern(y^*; \hat{\psi})
    \label{predp-y}
\eeq

Here, $\bern(y;\psi)\,=\,\psi^y\,(1-\psi)^{1-y}$ and
$\hat{\psi}\,=\,(f_1+N_1)\,/\,(f_0+f_1+n)$, with  $N_y =
\sum\limits_{\ell=1}^{n} I(y^{(\ell)}=y)$. Note that $\hat{\psi}$ is just the
posterior mean of $\psi$ based on $y^{(1)},\ldots,y^{(n)}$.

Similarly, integrating over $\phi_{0,j}$ and $\phi_{1,j}$, we find 
that\vspace*{-3pt}
\beq
  P(x\trn_j\ |\ \theta_j,\,\alpha,\,y\trn) &\! =\! &
  \prod_{y=0}^1\,U(\alpha\theta_j,\,\alpha(1\!-\!\theta_j),\,
  I_{y,j},\,O_{y,j})
\label{pxj-b}\\[-13pt]\nonumber
\eeq
where $O_{y,j}\,=\,\sum\limits_{i=1}^n I(y^{(i)}=y,\,x_j^{(i)}=0)$ and
$I_{y,j}\,=\,\sum\limits_{i=1}^n I(y^{(i)}=y,\,x_j^{(i)}=1)$.

With $\psi$ and $\phib$ integrated out, we need deal only with
the remaining parameters, $\alpha$ and $\thetab$.  Note that
after eliminating $\psi$ and the $\phib$, the cases are no longer
independent (though they are exchangeable).  However, conditional 
on the responses, $y\trn$, and on $\alpha$, the values of
different features are still independent.  This is crucial to the
efficiency of the computations described below.

\subsection{Predictions for Test Cases using Numerical
Quadrature}\label{sub-pred-b}

We first describe how to predict the class for a test case when we are
either using all features, or using a subset of features without any
attempt to correct for selection bias.  We then consider how to make
predictions using our method of correcting for selection bias.

Suppose we wish to predict the response, $y^*$, in a test case for which we
know the retained features $\x_{1:k}^*=(\x^*_1,\cdots,\x^*_k)$ (having
renumbered features as necessary).  For this, we need the following
predictive probability:
\beq
  P(y^*\, |\, \x_{1:k}^*,\,\x_{1:k}\trn,\,y\trn) & = &
  { P(y^*\, |\, y\trn)\,P(\x_{1:k}^*\, |\, y^*,\,\x_{1:k}\trn,\,y\trn
    )_{\rule{0pt}{10pt}} \over
    \sum\limits_{y=0}^1
      P(y^*=y\, |\, y\trn)\,P(\x_{1:k}^*\, |\, y^*=y,\,\x_{1:k}\trn,\,y\trn)}
  \label{eq-unc-pp}
\eeq
Ie, we evaluate the numerator above for $y^*=0$ and
$y^*=1$, then divide by the sum to obtain the predictive probabilities.
The first factor in the numerator, $P(y^*\, |\, y\trn)$, is given
by equation~(\ref{predp-y}).  It is sufficient to obtain the
second factor up to a proportionality constant that doesn't depend on $y^*$,
as follows:
\beq
 P(\x_{1:k}^*\ |\ y^*,\,\x_{1:k}\trn,\,y\trn)
 &=& {P(\x_{1:k}^*,\,x_{1:k}\trn\ |\ y^*,\,y\trn)_{\rule{0pt}{9pt}} \over
     P(x_{1:k}\trn\ |\ y\trn)^{\rule{0pt}{7pt}}} \\
 &\propto& P(\x_{1:k}^*,\,x_{1:k}\trn\ |\ y^*,\,y\trn)
\eeq
This can be computed by integrating over $\alpha$, noting that conditional
on $\alpha$ the features are independent:\vspace*{-4pt}
\beq
  P(\x_{1:k}^*,\,x_{1:k}\trn\ |\ y^*,\,y\trn)
  & = & 
  \int P(\alpha)\,P(\x_{1:k}^*,\,x_{1:k}\trn\ |\ \alpha,\,y^*,\,y\trn)\,d\alpha
  \label{eq-alpha-int}\\[4pt]
  & = & 
  \int P(\alpha)\, \prod_{j=1}^k
   P(\x_j^*,\,x_j\trn\ |\ \alpha,\,y^*,\,y\trn)\,d\alpha
  \label{eq-alpha-int2}
\eeq
Each factor in the product above is found by using equation~(\ref{pxj-b})
and integrating over $\theta_j$:
\beq
 \lefteqn{P(\x_j^*,\,x_j\trn\ |\ \alpha,\,y^*,\,y\trn)} \nonumber\\
& = &
  \int_{0}^{1}\!
  P(\x_j^*\ |\ \theta_j,\,\alpha,\,\x_j\trn,\,y\trn,\, y^*)\,
  P(\x_j\trn\ |\ \theta_j,\,\alpha,\,y\trn)\,d\theta_j \ \ \ \ \ \ \\[2pt]
 &=&
 \int_{0}^{1}\!
  \bern(\x^*_j;\hat\phi_{y^*,j})\,
  \prod_{y=0}^1\,U(\alpha\theta_j,\,\alpha(1\!-\!\theta_j),
  \,I_{y,j},\,O_{y,j})\, d\theta_j\ \ \ \ \ \
  \label{eq-jfact}
\eeq
where $\hat\phi_{y^*,j}= (\alpha\theta_j+I_{y^*,j})\ /\
(\alpha+N_{y^*})$, the posterior mean of $\phi_{y^*,j}$ given $\alpha$
and $\theta_j$. 

When using $k$ features selected from a larger number, $p$, the predictions
above, which are conditional on only $x\trn_{1:k}$ and $y\trn$, are not correct
--- we should also condition on the event,
$\mathcal{S}$, that $|\COR (y\trn,x\trn_j)|\,\le\,\gamma$ for $j=k+1,\ldots,p$.
We need to modify the predictive probability of equation~(\ref{eq-unc-pp})
by replacing $P(\x_{1:k}^*\ |\ y^*,\,\x_{1:k}\trn,\,y\trn)$ with
$P(\x_{1:k}^*\ |\ y^*,\,\x_{1:k}\trn,\,y\trn,\,\mathcal{S})$, which is
proportional to $P(\x_{1:k}^*,\,\x_{1:k}\trn,\,\mathcal{S}\ |\ y^*,\,y\trn)$.
Analogously to equations~(\ref{eq-alpha-int}) and~(\ref{eq-alpha-int2}), 
we obtain
\beq
  \lefteqn{P(\x_{1:k}^*,\,x_{1:k}\trn,\,\mathcal{S}\ |\ y^*,\,y\trn)}\nonumber\\
  & = & 
  \int P(\alpha)\,P(\x_{1:k}^*,\,x_{1:k}\trn,\,\mathcal{S}
    \ |\ \alpha,\,y^*,\,y\trn)\,d\alpha
  \label{eq-alpha-int-mod}\\[4pt]
  & = & 
  \int P(\alpha)\, P(\mathcal{S}\ |\ \alpha,\,y\trn) \prod_{j=1}^k
   P(\x_j^*,\,x_j\trn\ |\ \alpha,\,y^*,\,y\trn)\,d\alpha\ \ \ \
  \label{eq-alpha-int2-mod}
\eeq
The factors for the $k$ retained features are computed as before, using
equation~(\ref{eq-jfact}).  The additional correction factor that is needed
(presented earlier as equation~(\ref{eq-cor-lik})) is
\beq
  P(\mathcal{S}\ |\ \alpha,\,y\trn) & = &
    \prod_{j=k+1}^p P(|\COR (y\trn,x\trn_j)|\,\le\,\gamma\ |\ \alpha,\,y\trn)
  \\[3pt]
  & = &
  \Big[\, 
      P(|\COR (y\trn,x\trn_t)|\,\le\,\gamma\ |\ \alpha,\,y\trn)
    \,\Big]^{p-k}
  \label{eq-corfact}
\eeq
where $t$ is any of the omitted features, all of which have the same
probability of having a small correlation with $y$.  
We discuss how to compute this adjustment factor in the next section.

To see intuitively why this adjustment factor will correct for
selection bias, recall that as discussed in
Section~(\ref{sub-model-b}), when $\alpha$ is small, features will be
more likely to have a strong relationship with the response. If the
likelihood of $\alpha$ is based only on the selected features, which
have shown high correlations with the response in the training
dataset, it will favor values of $\alpha$ that are inappropriately
small.  Multiplying by the adjustment factor, which favors larger
values for $\alpha$, undoes this bias.

We compute the integrals over $\alpha$ in equations~(\ref{eq-alpha-int2})
and~(\ref{eq-alpha-int2-mod}) by numerical quadrature.  We use the midpoint
rule, applied to $u=F(\alpha)$, where $F$ is the cumulative distribution
function for the Inverse-Gamma$(a,b)$ prior for $\alpha$. The prior for $u$ is
uniform over $(0,1)$, and so needn't be explicitly included in the integrand. 
With $K$ points for the midpoint rule, the effect is that we average the value
of the integrand, without the prior factor, for values of $\alpha$ that are the
$0.5/K, 1.5/K, \ldots, 1-0.5/K$ quantiles of its Inverse-Gamma prior. For each
$\alpha$, we use Simpson's Rule to compute the one-dimensional integrals over
$\theta_j$ in equation~(\ref{eq-jfact}).  

\subsection{Computation of the Adjustment Factor for Naive Bayes Models}
\label{sec-adj-b}

Our remaining task is to compute the adjustment factor of
equation~(\ref{eq-corfact}), which depends on the
probability that a feature will have correlation less than $\gamma$ in
absolute value.  Computing this seems difficult ---
we need to sum the probabilities of $\x_t\trn$ given $y\trn$,
$\alpha$ and $\theta_t$ over all configurations of $\x_t\trn$ 
for which $|\COR (y\trn,x\trn_t)|\,\le\,\gamma$ ---
but the computation can be 
simplified by noticing that $\COR(x\trn_t,y\trn)$ can be written in terms of
\mbox{$I_0\,=\,\sum_{i=1}^n I(y^{(i)}=0,\,x^{(i)}_t=1)$} and
$I_1\,=\,\sum_{i=1}^n I(y^{(i)}=1,\,x^{(i)}_t=1)$, as follows:\vspace*{-7pt}
\beq
  \COR(x_t\trn,\,y\trn)
  & = &
  {\displaystyle\sum_{i=1}^n\, \big(y^{(i)}-\bar y\big)\,x^{(i)}_t
  \over
   \sqrt{\sum\limits_{i=1}^n \big(y^{(i)}-\bar y\big)^2}\
   \sqrt{\sum\limits_{i=1}^n \big(x^{(i)}_t-\bar x_t\big)^2} } \\[6pt]
  & = &
  { (0-\ybar)\,I_0\ +\ (1-\ybar)\,I_1 \over
     \sqrt{n \ybar (1\!-\!\ybar)}\,
     \sqrt{I_0+I_1-(I_0+I_1)^2/n} }
  \label{cor-express}
\eeq
We write the above as $\Cor(I_0,I_1,\ybar)$, taking $n$ as known.
This function is defined for $0 \le I_0 \le n(1\!-\!\ybar)$ and
$0 \le I_1 \le n\ybar$.

\begin{figure}

\hspace*{0.2in}\psfig{file=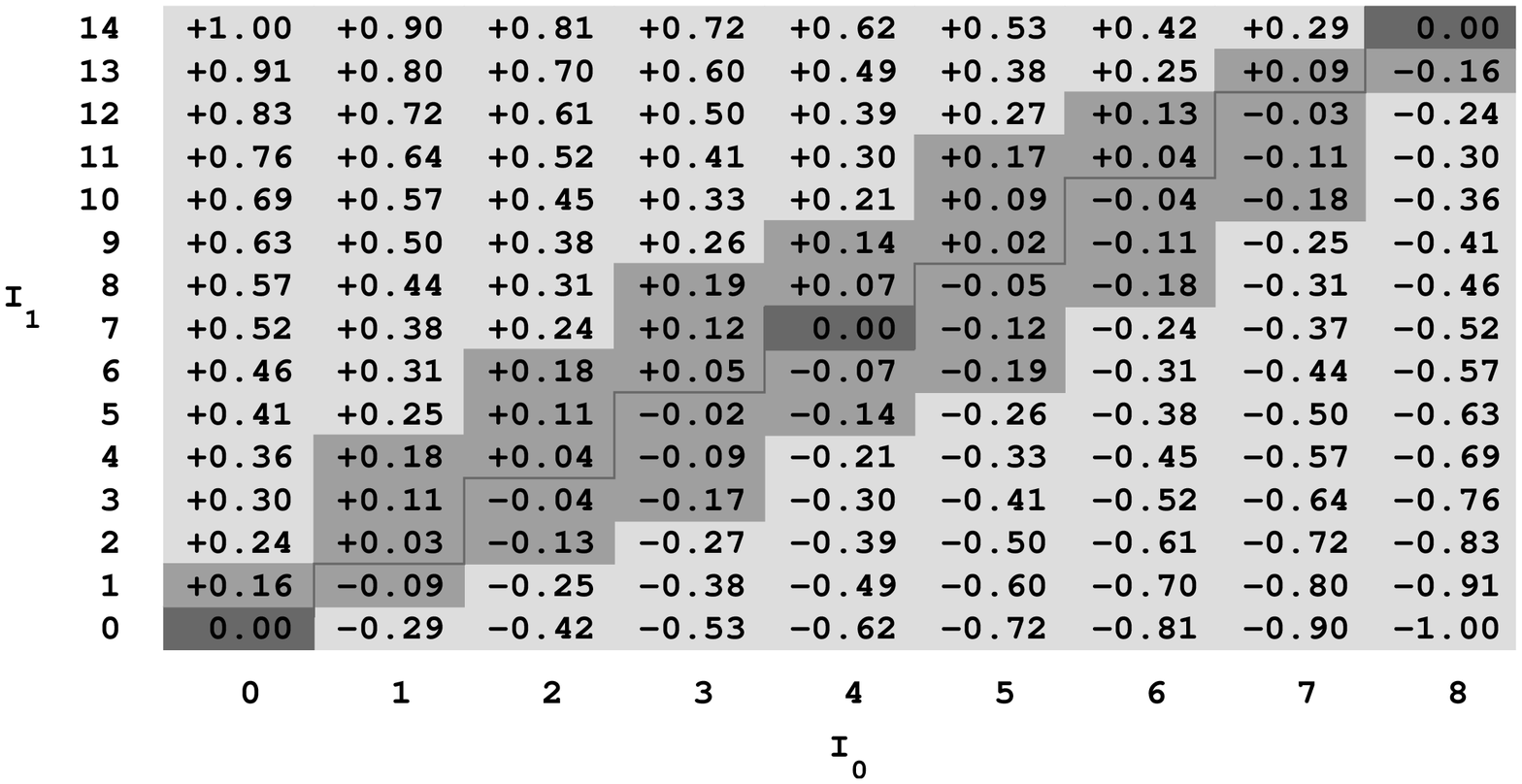,width=6in}

\caption[Display of sample correlations with an example]{The $\Cor$ function for
a dataset with $n=22$ and $\ybar=14/22$. The values of $\Cor(I_0,I_1,\ybar)$ are
shown for the valid range of $I_0$ and $I_1$.  Using $\gamma=0.2$, the values of
($I_0$,$I_1$) in $L_0$ are shown in dark grey, those in $L_-$ or $L_+$ in medium
grey, and those in $H_-$ or $H_+$ in light grey.}\label{fig-sets}

\end{figure}

Fixing $n$, $\ybar$, and $\gamma$, we can define the following sets
of values for $I_0$ and $I_1$ (for some feature $x_t$) in terms of the
resulting correlation with $y$:
\beq
   L_0 & = & \{\,(I_0,I_1)\ :\ \Cor(I_0,I_1,\ybar) = 0 \,\} \\[4pt]
   L_+ & = & \{\,(I_0,I_1)\ :\ 0 < \Cor(I_0,I_1,\ybar) \le \gamma \,\} \\[4pt]
   L_- & = & \{\,(I_0,I_1)\ :\ -\gamma \le \Cor(I_0,I_1,\ybar) < 0 \,\} \\[4pt]
   H_+ & = & \{\,(I_0,I_1)\ :\ \gamma < \Cor(I_0,I_1,\ybar) \,\} \\[4pt]
   H_- & = & \{\,(I_0,I_1)\ :\ \Cor(I_0,I_1,\ybar) < -\gamma \,\}
\eeq
A feature will be discarded if $(I_0,I_1) \,\in\, L_- \cup L_0 \cup L_+$
and retained if $(I_0,I_1) \,\in\, H_- \cup H_+$.  These sets are
illustrated in Figure~\ref{fig-sets}.

We can write the probability needed in equation~(\ref{eq-corfact}) using
either $L_-$, $L_0$, and $L_+$ or $H_-$ and $H_+$.  We will take the latter 
approach here, as follows:
\beq
\ \ \ \
\lefteqn{P(\, |\COR(x\trn_t,y\trn)| \leq \gamma \ |\ \alpha,\,y\trn)}\nonumber\\
 & = &
   1\ -\ P(\,(I_0,I_1)\,\in\, H_- \cup H+\ |\ \alpha,\,y\trn) \\[6pt]
 & = & 1\ -\ \!\!\!\!\!\!\sum_{
   \scriptstyle (I_0,I_1)\, \in \,
                     \scriptstyle H_- \cup H_+ 
   }\!\!\!\!\!\!
   P(I_0,\,I_1\ |\ \alpha,\,y\trn)
 \label{adj-fact-H1}
\eeq

We can now exploit symmetries of the prior and of the $\Cor$ function
to speed up computation.  First, note that
$\Cor(I_0,I_1,\ybar)\, =\, -\Cor(n(1\!-\!\ybar)-I_0,n\ybar-I_1,\ybar)$,
as can be derived from equation~(\ref{cor-express}), or by simply 
noting that exchanging labels for the classes should change only the sign
of the correlation.  The one-to-one mapping $(I_0,I_1) \rightarrow
(n(1\!-\!\ybar)-I_0,n\ybar-I_1)$, which maps $H_-$ and $H_+$ and vice
versa (similarly for $L_-$ and $L_+$), therefore leaves $\Cor$ unchanged.
The priors for
$\theta$ and $\phi$ (see~(\ref{prior-theta-b}) and~(\ref{prior-phi-b}))
are symmetrical with respect to the class labels 0 and 1, so the prior
probability of
($I_0,\,I_1)$ is the same as that of $(n(1\!-\!\ybar)-I_0,\,n\ybar-I_1)$.
We can therefore rewrite equation~(\ref{adj-fact-H1}) as
\beq
P(\, |\COR(x\trn_t,y\trn)| \leq \gamma \ |\ \alpha,\,y\trn) & = & 1\ -\
      2\!\!\!\!\!\!\sum_{(I_0,I_1)\, \in\, H_+}\!\!\!\!\!\!\!
      P(I_0,\,I_1\ |\ \alpha,\,y\trn)
 \label{adj-fact-H2}
\eeq

At this point we write the probabilities for $I_0$ and $I_1$ in
terms of an integral over $\theta_t$, and then swap the order of summation and
integration, obtaining
\beq
  \sum_{(I_0,I_1)\, \in\, H_+}\!\!\!\!\!\!\!
      P(I_0,\,I_1\ |\ \alpha,\,y\trn) & = &
  \int_{0}^{1}\!\!
  \sum_{(I_0,I_1)\, \in\, H_+}\!\!\!\!\!
    P(I_0,\,I_1\ |\ \alpha,\,\theta_t,\,y\trn)\ d\theta_t
\label{eq-adjint}
\eeq
The integral over $\theta_t$ can be approximated using
some one-dimensional numerical quadrature method (we use Simpson's Rule),
provided we can evaluate the integrand.

The sum over $H_+$ can easily be delineated because
$\Cor(I_0,I_1,\ybar)$ is a monotonically decreasing function of $I_0$,
and a monotonically increasing function of $I_1$, as may be confirmed
by differentiating with respect to $I_0$ and $I_1$.  Let $b_0$ be the
smallest value of $I_1$ for which $\Cor(0,I_1,\ybar) > \gamma$.  Taking
the ceiling of the solution of $\Cor(0,I_1,\ybar)=\gamma$, we find
that $b_0\, =\,\lceil 1/(1/n+(1-\bar{y})/(n\bar{y}\gamma^2))\rceil$.
For $b_0 \le I_1 \le n\ybar$, let $r_{I_1}$ be the largest value of
$I_0$ for which $\Cor(I_0,I_1,\ybar) > \gamma$.  We can write
\beq
  \sum_{(I_0,I_1)\, \in\, H_+}\!\!\!\!\!
    P(I_0,\,I_1\ |\ \alpha,\,\theta_t,\,y\trn) & = &
  \sum_{I_1=b_0}^{n\ybar}\, \sum_{I_0=0}^{r_{I_1}}\,
    P(I_0,\,I_1\ |\ \alpha,\,\theta_t,\,y\trn)
\eeq

Given $\alpha$ and $\theta_t$, $I_0$ and $I_1$ are independent, so
we can reduce the computation needed by rewriting the above expression
as follows:
\beq
  \lefteqn{\sum_{(I_0,I_1)\, \in\, H_+}\!\!\!\!\!
    P(I_0,\,I_1\ |\ \alpha,\,\theta_t,\,y\trn)} \nonumber \\
& = &
  \sum_{I_1=b_0}^{n\ybar}\,P(I_1\ |\ \alpha,\,\theta_t,\,y\trn)\,
  \sum_{I_0=0}^{r_{I_1}}\, P(I_0\ |\ \alpha,\,\theta_t,\,y\trn)
\label{eq-sum-H}
\eeq
Note that the inner sum can be
updated from one value of $I_1$ to the next by just adding any additional
terms needed.
This calculation therefore requires $1\!+\!n\ybar\!-\!b_0 \le n$ 
evaluations of $P(I_1\ |\
\alpha,\,\theta_t,\,y\trn)$ and $1\!+\!r_{n\ybar} \le n$ evaluations of
$P(I_0\ |\ \alpha,\,\theta_t,\,y\trn)$.  

To compute $P(I_1\ |\ \alpha,\,\theta_t,\,y\trn)$, we multiply
the probability of any particular value for $x\trn_t$ in which there are $I_1$
cases with $y=1$ and $x_t=1$ by the number of ways this can occur.
The probabilities are found 
by integrating
over $\phi_{0,t}$ and $\phi_{1,t}$, as described in Section~\ref{sub-int-b}.
The result is\vspace*{-6pt}
\beq
  P(I_1\ |\ \alpha,\,\theta_t,\,y\trn) & = &
    \mychoose{n\ybar}{I_1} U(\alpha\theta_t,\,\alpha(1\!-\!\theta_t),\,
                            I_1,\,n\ybar-I_1)
\eeq
Similarly,
\beq
  P(I_0\ |\ \alpha,\,\theta_t,\,y\trn) & = &
    \mychoose{n(1\!-\!\ybar)}{I_0} U(\alpha\theta_t,\,\alpha(1\!-\!\theta_t),\,
                            I_0,\,n(1\!-\!\ybar)-I_0)
\eeq
One can easily derive simple expressions for
$P(I_1\ |\ \alpha,\,\theta_t,\,y\trn)$ and
$P(I_0\ |\ \alpha,\,\theta_t,\,y\trn)$ in
terms of $P(I_1-1\ |\ \alpha,\,\theta_t,\,y\trn)$ and
$P(I_0-1\ |\ \alpha,\,\theta_t,\,y\trn)$, which avoid the need to compute
gamma functions or large products for each value of $I_0$ or $I_1$
when these values are used sequentially, as in equation~(\ref{eq-sum-H}).

\subsection{A Simulation Experiment}
\label{sec-sim-bnaive}

In this section, we use a dataset generated from the naive Bayes model
defined in Section~\ref{sub-model-b} to demonstrate the lack of
calibration that results when only a subset of features is used,
without correcting for selection bias.  We show that our
bias-correction method eliminates this lack of calibration.  We will
also see that for the naive Bayes model only a small amount of extra
computational time is needed to compute the adjustment factor needed
by our method.

\begin{figure}[p]

\centerline{\includegraphics{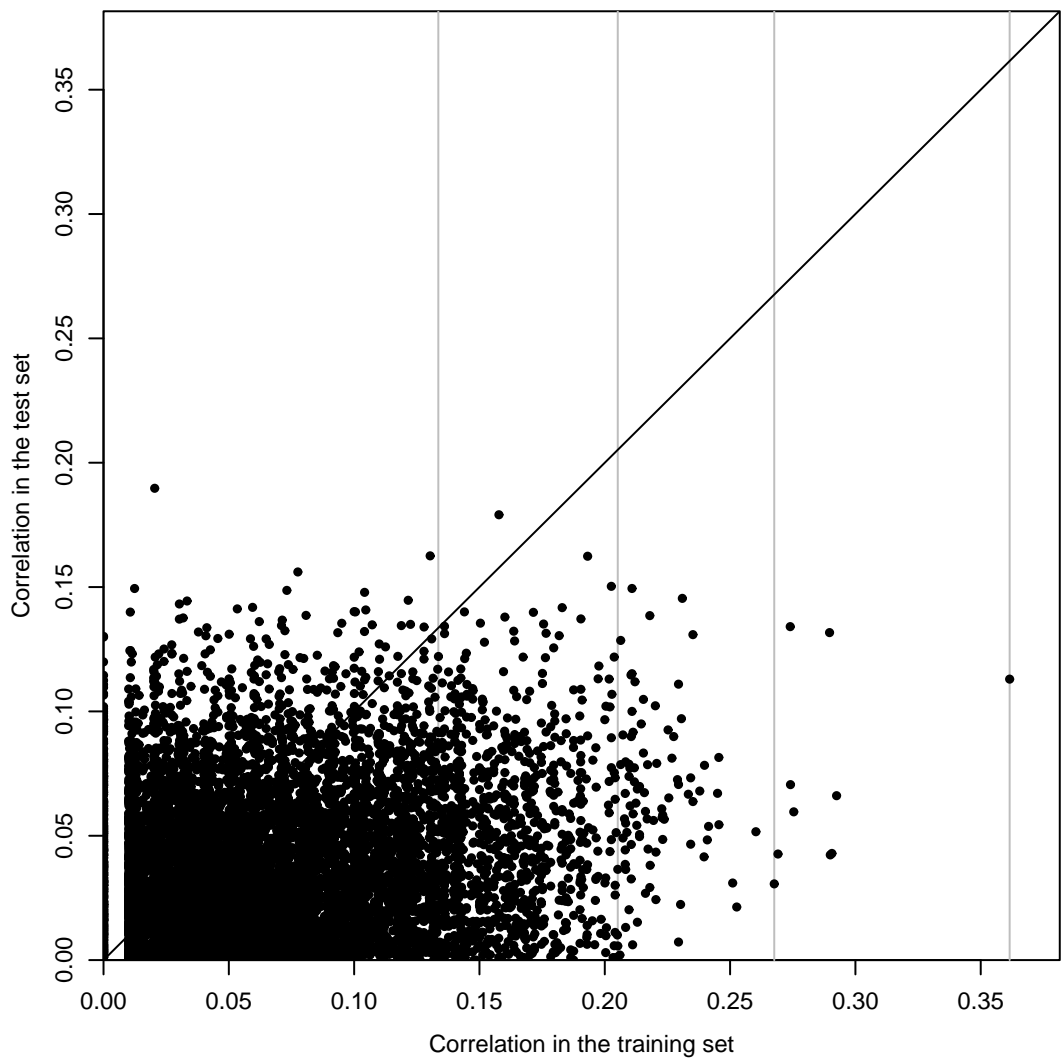}}

\caption[Scatter plot of sample correlations in the training set and the test
set]{The absolute value of the sample correlation of each feature with the
binary response, in the training set, and in the test set.  Each dot represents
one of the 10000 binary features. The training set correlations of the 1st,
10th, 100th, and 1000th most correlated features are marked by vertical
lines.}\label{fig-trn-tst-cor}

\end{figure}

Fixing $\alpha=300$, and $p=10000$, we used equations
(\ref{sample-x-b}), (\ref{prior-theta-b}) and (\ref{prior-phi-b}) to
generate a set of 200 training cases and a set of 2000 test cases,
both having equal numbers of cases with $y=0$ and $y=1$.  We then
selected four subsets of features, containing 1, 10, 100, and 1000
features, based on the absolute values of the sample correlations of
the features with $y$.  The smallest correlation (in absolute value)
of a selected feature with the class was 0.36, 0.27, 0.21, and 0.13
for these four subsets.  These are the values of $\gamma$ used by the
bias correction method when computing the adjustment factor of
equation~(\ref{eq-corfact}).  Figure~\ref{fig-trn-tst-cor} shows the
absolute value of the sample correlation in the training set of all
10000 features, plotted against the sample correlation in the test
set.  As can be seen, the high sample correlation of many selected
features in the training set is partly or wholly a matter of chance,
with the sample correlation in the test set (which is close to the
real correlation) often being much less.  The role of chance is
further illustrated by the fact that the feature with highest sample
correlation in the test set is not even in the top 1000 by sample
correlation in the training set.

\begin{figure}[p]

\includegraphics[height=3in,width=6.4in]{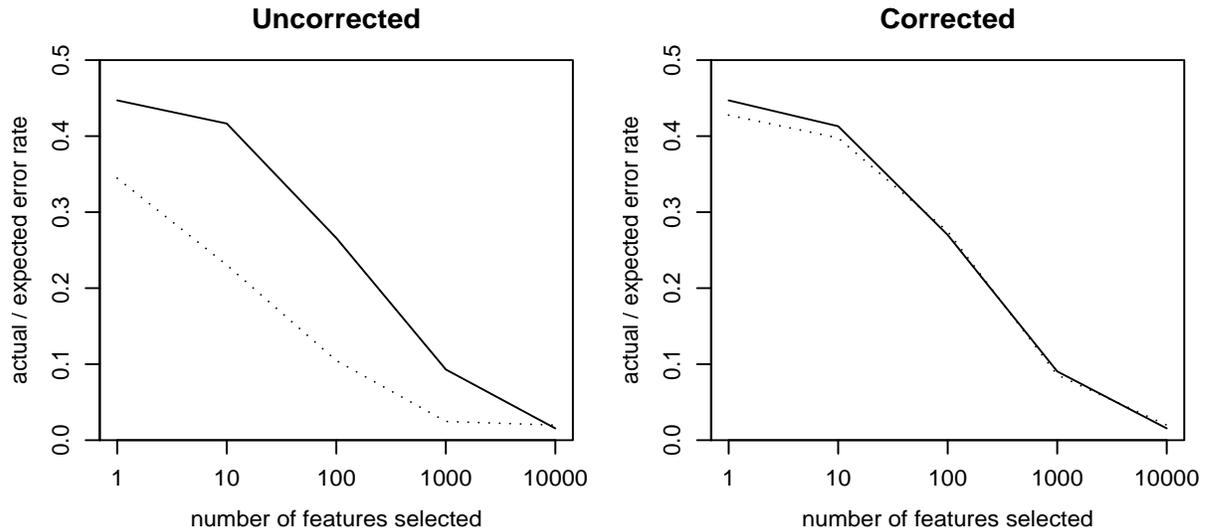}

\vspace*{-6pt}

\caption[Comparison of actual and expected error rates using simulated
data]{Actual and expected error rates with varying numbers of  features
selected, with and without correction for selection bias. The solid line is the
actual error rate on test cases. The dotted line is the error rate that would be
expected based on the  predictive probabilities.}\label{fig-sim-perf1}

\end{figure}

\begin{figure}[p]

\vspace*{-12pt}

\includegraphics[height=3in,width=6.4in]{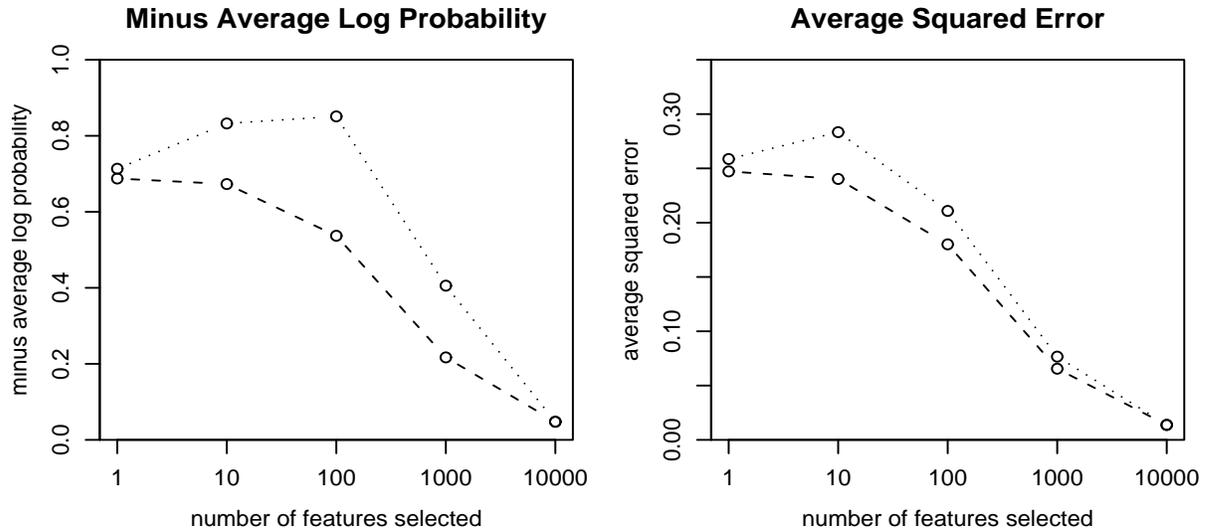}

\vspace*{-7pt}

\caption[Performance in terms of average minus log probability and average
squared error using simulated data]{Performance in terms of average minus log probability and average
squared error, with varying numbers of features selected, with and without
correction for selection bias. The left plot shows minus the average log
probability of the correct class for test cases, with 1, 10, 100, 1000, and all
10000 features selected. The dashed line is with bias correction, the dotted
line without.  The right plot is similar, but shows average squared error on
test cases. Note that when all 10000 features are used, there is no difference
between the corrected and uncorrected methods.}\label{fig-sim-perf2}

\end{figure}

For each number of selected features, we fit this data using the naive
Bayes model with the prior for $\psi$ (equation~(\ref{prior-psi-b}))
having $f_0=f_1=1$ and the prior for $\alpha$
(equation~(\ref{prior-alpha-b})) having shape parameter $a=0.5$ and
rate parameter $b=5$.  We then made predictions for the test cases
using the methods described in Section~\ref{sub-pred-b}.  The
``uncorrected'' method, based on equation~(\ref{eq-unc-pp}), makes no
attempt to correct for the selection bias, whereas the ``corrected''
method, with the modification of equation~(\ref{eq-alpha-int2-mod}),
produces predictions that account for the procedure used to select the
subset of features.  We also made predictions using all 10000
features, for which bias correction is unnecessary.

We compared the predictive performance of the corrected method with
the uncorrected method in several ways.  First, we looked at the error
rate when classifying test cases by thresholding the predictive
probabilities at $1/2$.  As can be seen in Figure~\ref{fig-sim-perf1},
there is little difference in the error rates with and without
correction for bias.  However, the methods differ drastically in terms
of the \textit{expected} error rate --- the error rate we would expect
based on the predictive probabilities for the test cases, equal to
$(1/N) \sum_i\, \hat p^{(i)}\, I(\hat p^{(i)}<0.5) \ +\ (1\!-\!\hat
p^{(i)})\, I(\hat p^{(i)}\ge0.5)$, where $\hat p^{(i)}$ is the
predictive probability of class 1 for test case $i$.  The predictive
probabilities produced by the uncorrected method would lead us to
believe that we would have a much lower error rate than the actual
performance.  In contrast, the expected error rates based on the
predictive probabilities produced using bias correction closely
match the actual error rates.

Two additional measures of predictive performance are shown in
Figure~\ref{fig-sim-perf2}.  One measure of performance is minus the
average log probability of the correct class in the $N$ test cases,
which is $-(1/N)\,\sum_{i=1}^N\,
[y^{(i)}\log(\hat{p}^{(i)})\,+\,(1\!-\!y^{(i)})\log(1\!-\!\hat{p}^{(i)})]$.
This measure heavily penalizes test cases where the actual class has a
predictive probability near zero.  Another measure, less sensitive to
such drastic errors, is the average squared error between the actual
class (0 or 1) and the probability of class 1, given by $(1/N)
\sum_{i=1}^N (y^{(i)}-\hat p ^{(i)})^2$.  The corrected method
outperforms the uncorrected one by both these measures, with the
difference being greater for minus average log probability.
Interestingly, performance of the uncorrected method actually gets
worse when going from 1 feature to 10 features.  This may be because
the single feature with highest sample correlation with the response
does have a strong relationship with the response (as may be likely in
general), whereas some other of the top 10 features by sample
correlation have little or no real relationship.

\begin{table}
\small \input{tab_naive}

\vspace*{-2.325in} 

\hspace*{2.0in}\begin{minipage}{4.2in}

\caption[Comparison of calibration for predictions found with and without
correction for selection bias,on data simulated from the binary naive  Bayes
model]{ Comparison of calibration for predictions found with and without
correction for selection bias, on data simulated from the binary naive  Bayes
model. Results are shown with four subsets of features and with the complete
data (for which no correction is necessary).  The test cases were divided into
10 categories by the first decimal of the predictive probability of class 1,
which is indicated by the 1st column ``C''.  The table shows the number  of test
cases in each category for each method (``\#''), the average predictive
probability of class 1 for cases in that category  (``Pred''), and the actual
fraction of these cases that were in class 1 (``Actual'').}

\label{tab-bnaive} 

\end{minipage}

\end{table}

We also looked in more detail at how well calibrated the predictive
probabilities were.  Table~\ref{tab-bnaive} shows the average
predictive probability for class 1 and the actual fraction of cases in
class 1 for test cases grouped according to the first
decimal of their predictive probabilities, for both the uncorrected
and the corrected method.  Results are shown using subsets of
1, 10, 100, and 1000 features, and using all features.
We see that the uncorrected method produces overconfident
predictive probabilities, either too close to zero or
too close to one.  The corrected method avoids such bias (the values
for ``Pred'' and ``Actual'' are much closer), showing that it is well
calibrated.  

\begin{figure}[hp]
\vspace*{-6pt}

\hspace*{12pt}\includegraphics[width=6.4in,height=4.5in]{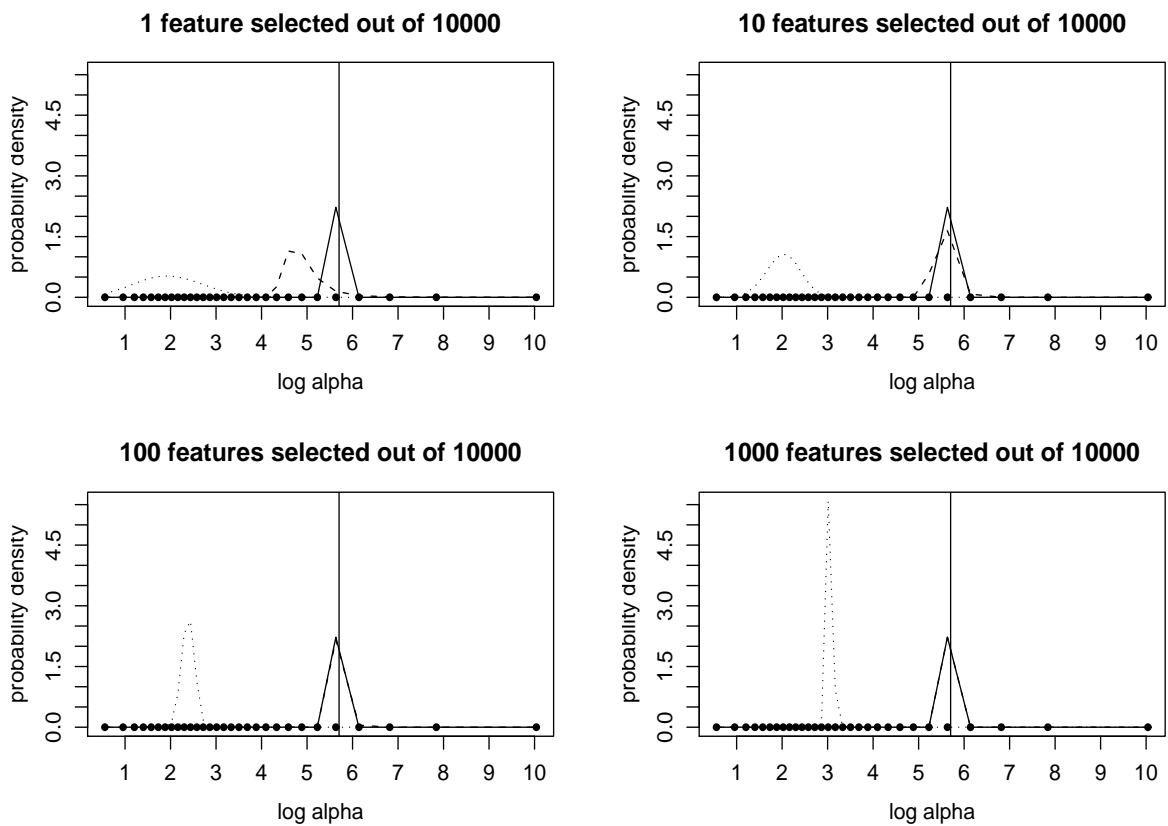}

\vspace*{-10pt}

\caption[Posterior distributions of $\log(\alpha)$ for the
simulated data]{Posterior distributions of $\log(\alpha)$ for the
simulated data, with different numbers of features selected.
The true value of $\log(\alpha)$ is 5.7, shown by the vertical line.
The solid line is the posterior density using all features.  For each number
of selected features, the dashed line is the posterior density
including the factor that corrects for selection bias; the dotted line is
the posterior density without bias correction.  
The dashed and solid lines overlap in the bottom two graphs.
The dots mark the values of $\log(\alpha)$ used to approximate the density, at
the $0.5/K,1.5/K,\ldots,(K\!-\!0.5)/K$ quantiles of the prior distribution
(where $K=30$).
The probabilities of $x\trn$ at each of these values for 
$\alpha$ were computed, rescaled to sum to $K$, and finally 
multiplied by the Jacobian, $\alpha P(\alpha)$,
to obtain the approximation to the posterior density of 
$\log(\alpha)$
}
\label{fig-post-alpha}
\end{figure}

\begin{table}[p]

\begin{center}

\input{naive-time-tab}

\end{center}

\vspace*{-0.2in}\caption{Computation times from simulation experiments with
naive Bayes models }

\label{tab-time-naive}

\end{table}

The biased predictions of the uncorrected method result from an
incorrect posterior distribution for $\alpha$, as illustrated in
Figure~\ref{fig-post-alpha}.  Without bias correction, the posterior
based on only the selected features incorrectly favours values of
$\alpha$ smaller than the true value of 300.  Multiplying by the
adjustment factor corrects this bias in the posterior distribution.

Our software (available from
\texttt{http://www.utstat.utoronto.ca/$\sim$longhai}) is written in the
R language, with some functions for intensive computations such as
numerical integration and computation of the adjustment factor written
in C for speed.  We approximated the integral with respect to $\alpha$
using the midpoint rule with $K=30$ values for $F(\alpha)$, as
discussed at the end of Section~\ref{sub-pred-b}.  The integrals with
respect to $\theta$ in equations~(\ref{eq-jfact})
and~(\ref{eq-adjint}) were approximated using Simpson's Rule,
evaluating $\theta$ at 21 points.

Computation times for each method (on a 1.2 GHz UltraSPARC III processor) are
shown in Table~\ref{tab-time-naive}.  The corrected method is almost as fast as
the uncorrected method, since the time to compute the adjustment factor is
negligible compared to the time spent computing the integrals over $\theta_j$
for the selected features.  Accordingly, considerable time can be saved by
selecting a subset of features, rather than using all of them, without
introducing an optimistic bias, though some accuracy in predictions may of
course be lost when we discard the information contained in the unselected
features.

\subsection{A Test Using Gene Expression Data} \label{sec-gene}

We also tested our method using a publicly available dataset on gene expression
in normal and cancerous human colon tissue.  This dataset contains the
expression levels of 6500 genes in 40 cancerous and 22 normal colon tissues,
measured using the Affymetrix technology. The dataset is available at
\texttt{http://geneexpression.cinj.org/$\sim$notterman/affyindex.html}. We
used only the 2000 genes with highest minimal intensity, as selected by Alon,
Barkai, Notterman, Gish, Mack, and Levine (1999). In order to apply the binary
naive Bayes model to the data, we transformed the real-value data into binary
data by thresholding at the median, separately for each feature.

We divided these 2000 genes randomly into 10 equal groups, producing
10 smaller datasets, each with 200 features.  We applies the corrected
and uncorrected methods separately to each of these 10 datasets,
allowing some assessment of variability when comparing performance.
For each of these 10 datasets, we used leave-one-out cross validation
to obtain the predictive probabilities over the 62 cases.  In this
cross-validation procedure, we left out each of the 62 cases in turn,
selected the five features with the largest sample correlation with
the response (in absolute value), and found the predictive probability
for the left-out case using the binary naive Bayes model, with and
without bias correction.  The absolute value of the correlation of the
last selected feature was around 0.5 in all cases.  We used the same
prior distribution, and the same computational methods, as for the
demonstration in Section~\ref{sec-sim-bnaive}.

\begin{figure}[p]

\hspace*{-0.1in}\includegraphics{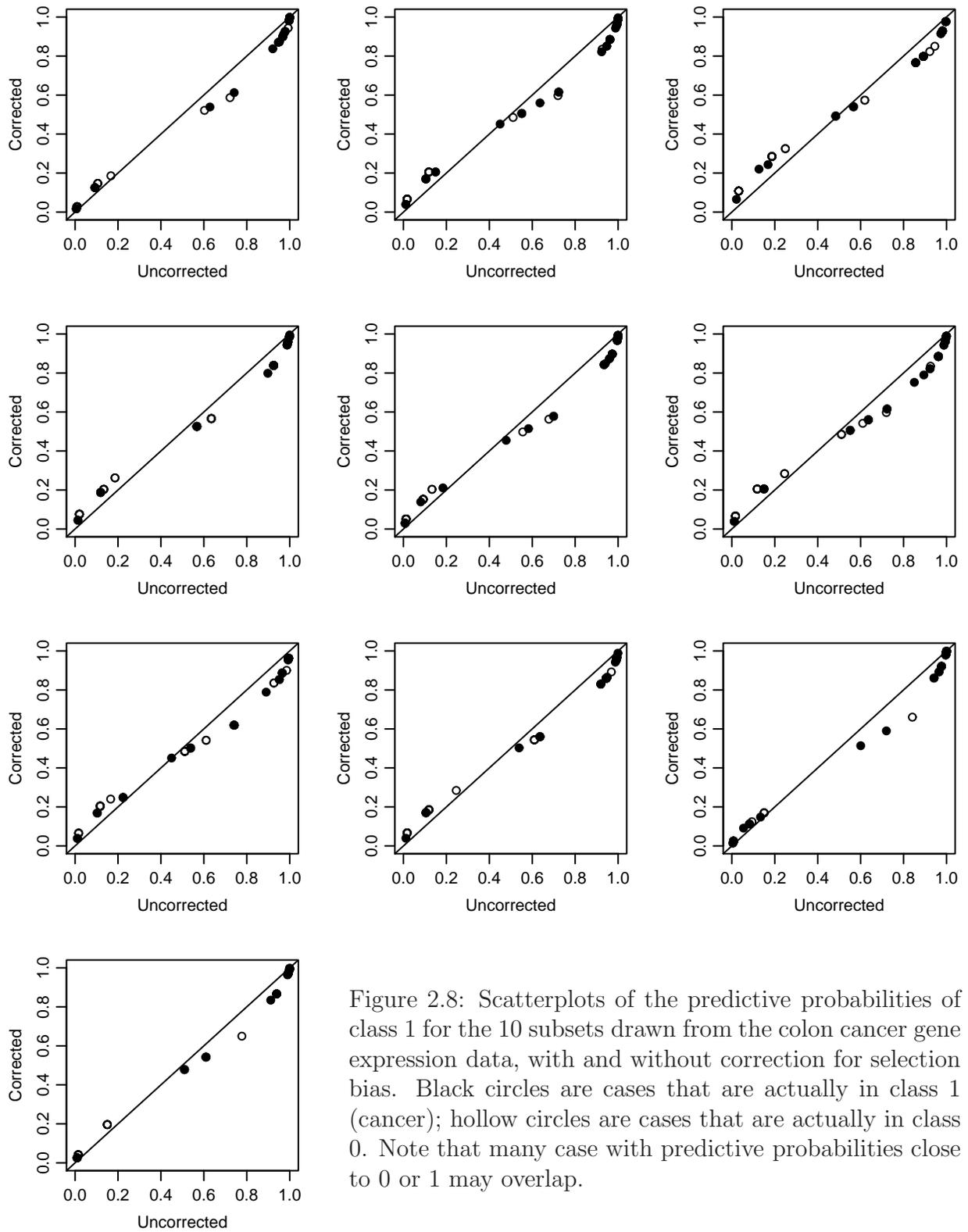}

\vspace*{-2in}

\hspace*{2.3in}\begin{minipage}{4.1in}\caption[Scatterplots of the predictive
probabilities of class 1 for the 10 subsets drawn from the colon cancer gene
expression data]{Scatterplots of the predictive probabilities of class 1 for
the 10 subsets drawn from the colon cancer gene expression data, with and
without correction for selection bias.  Black circles are cases that are
actually in class 1 (cancer); hollow circles are cases that are actually in
class 0. Note that many case with predictive probabilities close to 0 or 1 may 
overlap.}\label{fig-colon-pred-prob}\end{minipage}

\vspace*{0.4in}

\end{figure}

Figure~\ref{fig-colon-pred-prob} plots the predictive probabilities of
class 1 for all cases, with each of the 10 subsets of features.  The
tendency of the uncorrected method to produce more extreme
probabilities (closer to 0 and 1) is clear.  However, when the
predictive probability is close to 0.5, there is little difference
between the corrected and uncorrected methods.  Accordingly, the two
methods almost always classify cases the same way, if prediction is
made by thresholding the predictive probability at 0.5, and have very
similar error rates.  Note, however, that correcting for bias would
have a substantial effect if cases were classified by thresholding the
predictive probability at some value other than 0.5, as would be
appropriate if the consequences of an error are different for the two
classes.

\begin{figure}[p] 

\begin{center}

\vspace*{-0.2in}

\includegraphics[width=6.4in,height=3.1in]{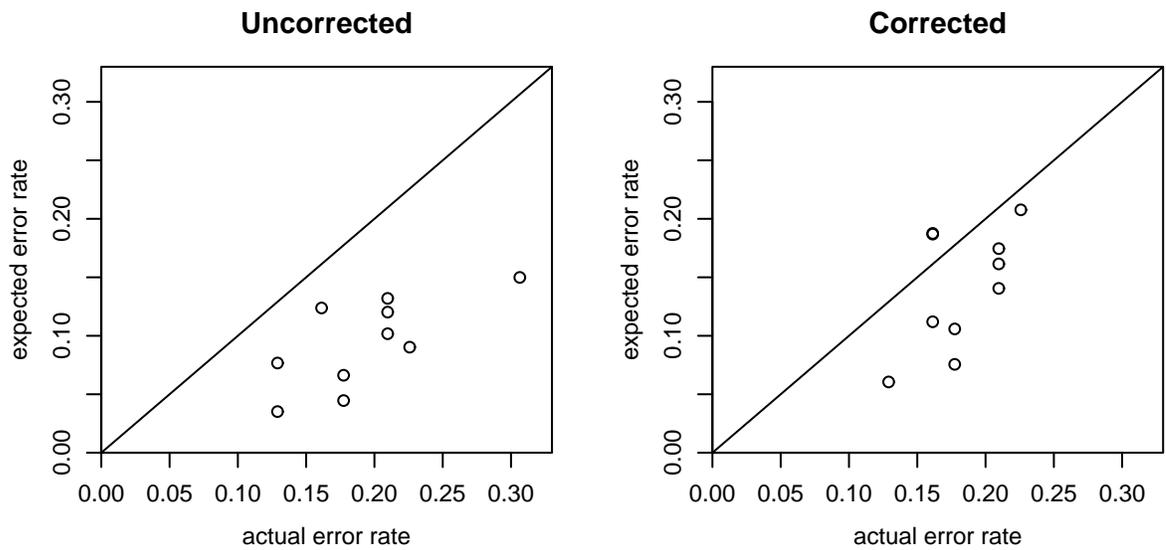}

\vspace*{-0.25in}

\end{center} 

\caption[Actual versus expected error rates on the colon cancer datasets]{Actual
versus expected error rates on the colon cancer datasets, with and without bias
correction.  Points are shown for each of the 10 subsets of features used for
testing. } \label{fig-colon2}

\end{figure}

\begin{figure}[p] 

\begin{center}

\vspace*{-0.2in}

\includegraphics[width=6.4in,height=3in]{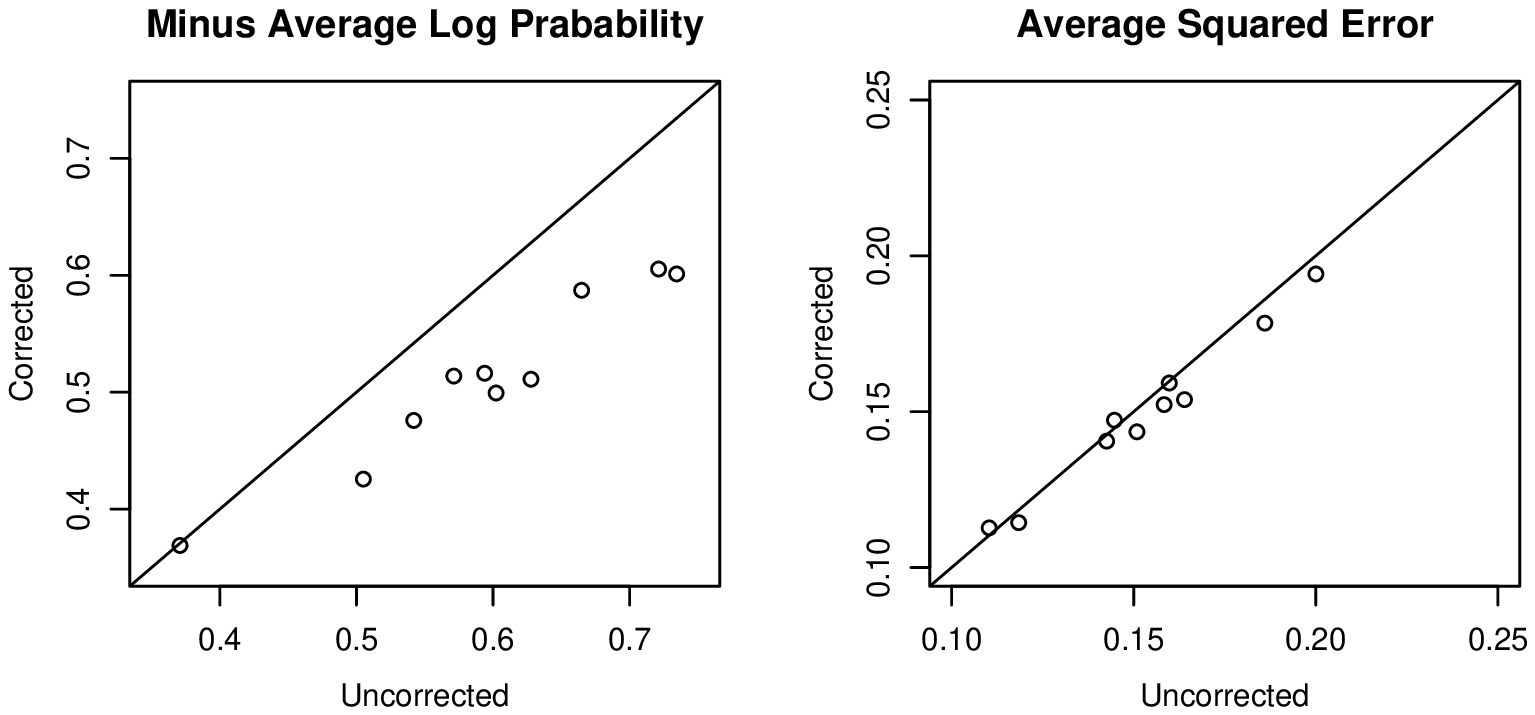}

\vspace*{-0.25in}

\end{center} 

\caption[Scatterplots of the average minus log probability of the correct class
and of the average squared error]{Scatterplots of the average minus log
probability of the correct class and of the average squared error (assessed by
cross validation) when using the 10 subsets of features for the colon cancer
gene expression data, with and without correcting for selection bias.}
\label{fig-colon}

\end{figure}

Figure~\ref{fig-colon} compares the two methods in terms of average
minus log probability of the correct class and in terms of average
squared error.  From these plots it is clear that bias correction
improves the predictive probabilities.  In terms of average minus log
probability, the corrected method is better for all 10 datasets, and
in terms of average squared error, the corrected method is better for
8 out of 10 datasets.  (A paired $t$ test with these two measures
produced $p$-values of $0.00007$ and $0.019$ respectively.)

Finally, Figure~\ref{fig-colon2} shows that our bias correction method
reduces optimistic bias in the predictions.  For each of the 10
datasets, this plot shows the actual error rate (in the leave-one-out
cross-validation assessment) and the error rate expected from the
predictive probabilities.  For all ten datasets, the expected error
rate with the uncorrected method is substantially less than the actual
error rate.  This optimistic bias is reduced in the corrected method,
though it is not eliminated entirely.  The remaining bias presumably
results from the failure in this dataset of the naive Bayes assumption
that features are independent within a class.

\section{Application to Bayesian Mixture Models}

Mixture modelling is another way to model the joint distributions of the
response variable and the predictor variables, from which we can find the
conditional distribution of the response variable given the predictor variables.
In this section we describe the application of the selection bias correction
method to a class of binary mixture models, which is a generalization of the
naive Bayes models.

\subsection{Definition of the Binary Mixture Models}

A complex distribution can be modeled using a mixture of finitely or infinitely
many simple distributions, often called mixture components, for example,
independent Gaussian distributions for real values or independent Bernoulli
distributions for binary values. Mixture models are often applied in density
estimation, classification, and latent class analysis problems, as discussed,
for example, by Everitt and Hand (1981), McLachlan and Basford (1988), and
Titterington, \textit{et al.}\ (1985). For finite mixture models with $K$
components, the density or probability function of the observation $\x$ is
written as \beq f(\x \given \phib_{0},\ldots,\phib_{K-1})=\sum_{k=0}^{K-1}
p_{k}\, f_{k}(\x\ |\ \phib_{k}) \eeq where $p_{k}$ is the mixing proportion of
component $f_k$, and $\phib_{k}$ is the parameter associated with component
$f_k$.

A Bayesian mixture model is often defined by introducing a latent label variable
for each case, written as $z$. Given $z=k$, the conditional distribution of the
observation $\x$ is the distribution for component $k$:

\beq
\x \given z=k,\phib_k \ \ \ \sim \ \ \ f_k(\x\given \phib_k)
\eeq

\begin{figure}[t]

\begin{center}

\includegraphics[scale=0.6]{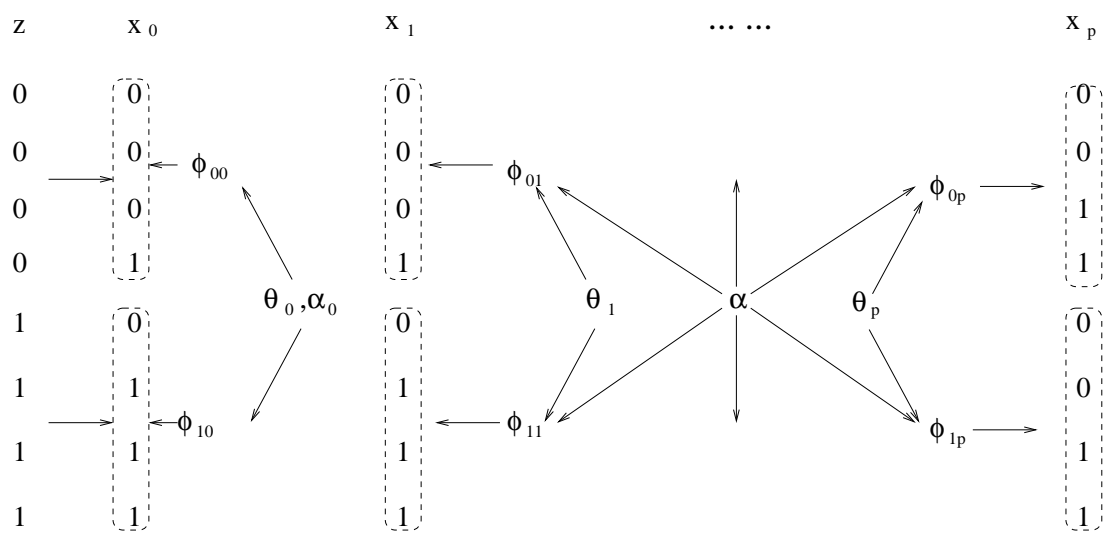}

\caption{A picture of Bayesian binary mixture models}

\label{fig-mix}

\end{center}

\end{figure}

We consider a two-component mixture model in this section, which is a
generalization of the naive Bayes model in Section~\ref{sec-bnaive}. Most of the
notation is therefore the same as in that section, except that we use the $0$th
feature, $x_0$, to represent the response $y$ here (and so $x_0\trn$ is
equivalent to $y\trn$) for convenience of presentation. The parameters of the
Bernoulli distributions are $\phi_{z,0}$ (for the response $x_0$), and
$\phi_{z,1},\cdots,\phi_{z,p}$ (for the features), collectively denoted as
$\phib_z$, for $z=0,1$. The priors for $\phib_0$ and $\phib_1$ are assigned in
the same way as for binary naive Bayes model. The labels of the cases are
assigned a prior in the same way as for $y$ in naive Bayes models. Conditional
on the component labels, all the features and the response are assumed to be
mutually independent. The models are displayed by Figure~\ref{fig-mix}, and are
described formally as follows: 

\beq \psi
&\sim& \betad(f_1,f_0) \label{prior-p} \\ %
\zz \ |\ \psi &\IID& \bern(\psi) \\ %
\alpha_0 
&\sim& \mbox{Inverse-Gamma}(a_0,b_0)\label{prior-alpha0}\\%
\alpha 
&\sim& \mbox{Inverse-Gamma}(a,b)
\label{prior-alpha-mix-cont}\\ %
\alpha_j &=&\Big\{
          \begin{array}{ll}\alpha_0 & \mbox{if }j = 0 \\ 
                           \alpha & \mbox{if }j>0
          \end{array} \\
\theta_0,\theta_1,\cdots,\theta_p 
&\IID& \mbox{Uniform}(0,1) \label{prior-phi}\\%
\phi_{0,j},\phi_{1,j}\ |\ \alpha_j,\theta_j &\IID&
\betad(\alpha_j\theta_j,\alpha_j(1-\theta_j))\ \ \  j=0,1,\cdots,p
\label{prior-theta}\\ %
\x^{(i)}_j\ |\ z^{(i)}, \phi_{z^{(i)},j} &\sim&
\bern(\phi_{z^{(i)},j}) \ \ \  i=1,\cdots,n \label{prob-x} \eeq

The naive Bayes models described in Section~\ref{sec-bnaive} can be seen as a
simpler form of the above mixture models by letting $\zz$ equal the responses
$\x\trn_0$, i.e., fixing  $\phi_{0,0}=1$ and  $\phi_{1,0}=0$. As for Bayesian
naive Bayes models, $\psi$ and $\theta_0,\ldots,\theta_p$ can be integrated away
analytically from Bayesian binary mixture models. The cases are then no longer
independent, but still are exchangeable. This integration reduces the number of
the parameters, therefore improves Markov chain sampling or numerical quadrature
if they are needed, though the resulting model may be harder to manipulate.

\subsection{Predictions for Test Cases using MCMC}\label{sec-mix-tr}

Let us start with no attempt to correct for the selection bias. We want to
predict the response, $\x^*_0$, of a test case for which we know the retained
features $x_1^*,\cdots,x_k^*$ (renumbering the features as necessary), based on
the training data $\x_{0:k}\trn$. For this, we need to calculate the predictive
distribution: 

\beq P(x^*_0 = 1\given \x^*_{1:k},\x_{0:k}\trn) =
    \frac{P(x^*_0 = 1,\x^*_{1:k} \given \x_{0:k}\trn) }
         {P(x^*_0 = 1,\x^*_{1:k} \given \x_{0:k}\trn) + 
          P(x^*_0 = 0,\x^*_{1:k} \given \x_{0:k}\trn)}           
\eeq

We therefore need to calculate $P(\x^*_{0:k} \given \x_{0:k}\trn)$ for $x^*_0 =
1$ and $x^*_0 = 0$. $P(\x^*_{0:k} \given \x_{0:k}\trn)$ can be written as:

\beq  
\lefteqn{P(\x^*_{0:k} \given \x_{0:k}\trn)}\nonumber \\
&=&
\sum_{\z\trn}\int_{\alpha_0}\int_{\alpha}\int_{\scriptsize\thetab} P(\x^*_{0:k} \given
\x_{0:k}\trn,\thetab_{0:k},\alpha_0,\alpha,\z\trn)\,\cdot \nonumber \\
&&\ \ \ \ \ \ \ \ \ \ \ \ \ \ \ \ \ \ \ \ \ \ \ \ \ \ \ \ \ \ \ \ \ \ 
 P(\thetab_{0:k},\alpha_0, \alpha,\z\trn\given
\x_{0:k}\trn)\,d\thetab_{0:k}\,d\alpha\, d\alpha_0  \\
&=&
{1\over P(\x\trn_{0:k})}\,
\sum_{\z\trn}\int_{\alpha_0}\int_{\alpha}\int_{\scriptsize\thetab} P(\x^*_{0:k} \given
\x_{0:k}\trn,\thetab_{0:k},\alpha_0,\alpha,\z\trn)\,\cdot\nonumber \\
&&\ \ \ \ \ \ \ \ \  
 P(\x_{0:k}\trn \given \thetab_{0:k},\alpha_0, \alpha,\z\trn)\,
 P(\thetab_{0:k})\,P(\alpha_0)\, P(\alpha)\,P(z\trn)\,d\thetab_{0:k}\,
 d\alpha\, d\alpha_0
\label{prob-pred-mix}
\eeq

The above integral is intractable analytically. We first approximate the
integrals with respect to $\alpha_0$ and $\alpha$ with the midpoint rule applied
to the transformed variables $u_0 = F_0(\alpha_0)$ and $u = F(\alpha)$, where
$F_0$ and $F$ are the cumulative distribution functions of the priors for
$\alpha_0$ and $\alpha$. Accordingly, the priors for $u_0$ and $u$ are uniform
over $(0,1)$. Suppose the midpoint rule evaluates the integrands with respect to
$u_0$ and $u$ at $K$ points respectively ($K$ can be different for $u_0$ and
$u$, for simplicity of presentation assume the same). After rewriting the
summation with respect to $u_0$ and $u$ over $K$ points,
$(1-0.5)/K,(2-0.5)/K,\ldots,(K-0.5)/K$, in terms of $\alpha_0$ and $\alpha$, the
midpoint rule approximation is equivalent to summing the integrand
in~(\ref{prob-pred-mix}), without the prior distribution for $\alpha_0$ and
$\alpha$, over the quantiles of the priors for $\alpha_0$ and $\alpha$
corresponding to probabilities $(1-0.5)/K,(2-0.5)/K,\ldots,(K-0.5)/K$. Let us
denote the $K$ quantiles of $\alpha_0$ by $\A_0$, and denote the $K$ quantiles
of $\alpha$ by $\A$. The integral in~(\ref{prob-pred-mix}), with $1/
P(\x_{0:k})$ omitted (since it is the same proportionality factor for all
$x^*_0$), is approximated by:

\beq
\lefteqn{\sum_{\z\trn}\sum_{\alpha_0 \in \A_0}\sum_{\alpha \in\A}
\int_{\scriptsize\thetab} P(\x^*_{0:k} \given
\x_{0:k}\trn,\thetab_{0:k},\alpha_0,\alpha,\z\trn)\,\cdot} \nonumber \\
&&\ \ \ \ \ \ \ \ \ \ \ \ \ \ \ \ \ 
P(\x_{0:k}\trn \given \thetab_{0:k},\alpha_0, \alpha,\z\trn)\,
   P(\thetab_{0:k})\,P(\z\trn)\, d\thetab_{0:k}
\label{prob-pred-mix-midrule}
\eeq

Since there are no prior terms in~(\ref{prob-pred-mix-midrule}) for $\alpha_0$
and $\alpha$, the above approximation with midpoint rule applied to $u_0$ and
$u$ can also be seen as approximating the continuous Inverse-Gamma priors for 
$\alpha_0$ and $\alpha$ by the uniform distributions over the finite sets $\A_0$
and $\A$. Based on these discretized priors for $\alpha_0$ and $\alpha$, we use
Gibbs sampling method to draw samples from the posterior distribution of
$\thetab_{0:k},\alpha_0, \alpha,\z\trn$, allowing the integral
in~(\ref{prob-pred-mix-midrule}) to be approximated with the Monte Carlo method.
The reason we use such a discretization for the prior for $\alpha$ is to ease 
the computation of the adjustment factor, which depends on $\alpha$. As will be
discussed later, when $\alpha$ is discrete, we can cache the values of the
adjustment factors for future use when the same $\alpha$ is used again. 

We now start to derive the necessary formulae for performing Gibbs sampling
for estimating~(\ref{prob-pred-mix-midrule}). Using the results from
Section~\ref{sub-int-b}, $P(\x^*_{0:k} \given
\x_{0:k}\trn,\thetab_{0:k},\alpha,\z\trn)$ in~(\ref{prob-pred-mix-midrule}) can
be calculated as follows: 

\beq 
\lefteqn{P(\x^*_{0:k} \given
\x_{0:k}\trn,\thetab_{0:k},\alpha_0,\alpha,\z\trn) }\nonumber \\
&=&\sum_{z^*=0}^1P(z^* \given \z\trn)\,P(\x^*_{0:k} \given
\x_{0:k}\trn,\thetab_{0:k},\alpha_0,\alpha,\z\trn,z^*)\\
&=&\sum_{z^*=0}^1\bern\left(z^*;\hat{\psi}\right)\,\prod_{j=0}^k\bern(x_j^*;
\hat{\phi}_{z^*,j}) \eeq 

\noindent where $\hat{\phi}_{z^*,j}=(\alpha_j\theta_j+I_{z^*,j})\ /\
(\alpha_j+n^{[z^*]})$,
$\hat{\psi}=(f_1+n^{[1]})/(f_0+f_1+n)$, $n^{[z]}=\sum_{i=1}^n
I(z_i=z)$, $I_{z,j}=\sum_{i=1}^n I(z^{(i)}=z,x_j^{(i)}=1)$, and
$O_{z,j}=\sum_{i=1}^n I(z^{(i)}=z,x_j^{(i)}=0)$. 

Again, using the results from Section~\ref{sub-int-b}, the distribution of
$\x_{0:k}^{(i)}$ given other training cases, denoted by $\x_{0:k}^{(-i)}$, which
is needed to update $z^{(i)}$ with Gibbs sampling, can be found:

\beq
  P(\x_{0:k}^{(i)}\given \x_{0:k}^{(-i)}, \z\trn,\thetab_{0:k},\alpha_0,\alpha)
    &=&\prod_{j=0}^k\bern(x^{(i)}_j;\hat{\phi}_{\zi,j}^{(-i)})
    \label{like-ci}
\eeq 

\noindent where $\hat{\phi}_{\zi,j}^{(-i)}=(\alpha_j\theta_j+I_{\zi,
j}^{(-i)})/(\alpha_j+n^{[\zi](-i)}-1)$, $I_{\zi,
j}^{(-i)}=\sum_{s=1}^nI(x_j^{(s)}=1,z^{(s)}=\zi,s \not=i)$ and
$n^{[\zi](-i)}=\sum_{s=1}^nI(z^{(s)}=\zi,s \not=i)$. 

Similarly, the distribution of the whole training data given $\z\trn,\
\thetab_{0:k},\alpha_0, \alpha$ based on the $k$ retained features can be
found: 

\beq
P(\x_{0:k}\trn\ |\ \z\trn,\  \alpha_0,\alpha,\ \thetab_{0:k})
&=&\prod_{j=0}^k\prod_{z=0}^{1}\,U(\alpha_j\theta_j,\,\alpha_j(1\!-\!\theta_j),
\,I_{z,j},\,O_{z,j}) \label{like-part-mix} \eeq 

Integrating away $\psi$ gives the prior for $\zz$: 

\beq P(\zz) &=& U(f_1,f_0,n^{[1]},n^{[0]})\label{dir} \eeq

\noindent From the priors for $\zz$, the conditional distribution of $z^{(i)}$
given all other $z^{(j)}$ except $z^{(i)}$, written as $z^{(-i)}$,  can be
found:

\beq P(z^{(i)}\ |\ z^{(-i)})=\bern\left(\zi;\hat{\psi}^{(-i)}\right) \label{cond-z}\eeq 

\noindent where $\hat{\psi}^{(-i)}=\frac{f_1+n^{[1](-i)}} {f_0+f_1+n-1}$ and
$n^{[1](-i)}=\sum_{s=1}^nI(z_s=1,s\not=i)$

We now can write out the conditional distributions needed for performing Gibbs
sampling. The conditional distribution of $\zi$ is proportional to the product
of~(\ref{cond-z}) and (\ref{like-ci}):

\beq P(z^{(i)}\given \x\trn,z^{(-i)},\thetab_{0:k},\alpha_0,\alpha)\propto
\bern\left(z^{(i)};\hat{\psi}^{(-i)}\right)\,
\prod_{j=0}^k\bern(x^{(i)}_j;\hat{\phi}_{\zi,
j}^{(-i)})\ \label{post-zi}
\eeq

\noindent The conditional distribution of $\theta_j$ is related to only feature
$j$ since the features are independent given $\zz$:

\beq
   P(\theta_j\ |\ \x\trn_j,\alpha_j,\z\trn) &\propto& \prod_{z=0}^{1}
   U(\alpha\theta_j,\,\alpha(1\!-\!\theta_j),\,I_{z,j},\,O_{z,j})
\label{post-phi-mix}
\eeq

\noindent The conditional distribution of $\alpha$ given $\x\trn_{0:k},\
\z\trn,$ and $\thetab_{0:k}$ is proportional to the products of the factors for
$j>0$ in~(\ref{like-part-mix}) since the prior for $\alpha$ is uniform over
$\A$. And the conditional distribution of $\alpha_0$ is proportional to the
factor for $j=0$  in~(\ref{like-part-mix}).

The prediction described above is, however, invalid if the $k$ features are
selected from a large number. It needs to be modified to condition also on the
information $\mathcal{S}$, i.e., we should compute $P(x^*_0\given
\x^*_{1:k},\x_{0:k}\trn,\mathcal{S})$. The calculations are similar to the
above, but with $P(\x_{0:k}\trn \given \thetab_{0:k},\alpha,\z\trn)$  replaced
by $P(\x_{0:k}\trn,\mathcal{S}\given \thetab_{0:k},\alpha,\z\trn)$
in~(\ref{prob-pred-mix}). Accordingly, the conditional distributions of
$\alpha$  and $\zz$  are multiplied by the following adjustment factor:

\beq P(\mathcal{S}\given x_0\trn,\z\trn,\alpha) = \left( \int_{0}^{1} P(\ |\COR(x\trn_t,x_0\trn)|
\leq \gamma \ |\ x_0\trn,\z\trn,\theta_t,\alpha)d\theta_t \right)^{p-k}
\label{adjust-factor-mix} \eeq

Compared with the adjustment factor for the Bayesian naive Bayes models, the
adjustment factor~(\ref{adjust-factor-mix}) is more difficult to calculate, as
we will discuss in the next section. Furthermore, this adjustment factor depends
on both $\alpha$ and the unknown latent label variables $\zz$, for which we need
to sample using Markov chain sampling method.  We therefore need to recompute
the adjustment factor whenever we change $\zz$ during Markov chain sampling run.
But we still need only to calculate the probability of one feature being
discarded then raise it to the power of $p-k$.

\subsection{Computation of the Adjustment Factor for Mixture Models}
\label{sec-mix-modify}

Computing $P(\mathcal{S}\given x_0\trn,\z\trn,\alpha)$ is similar to what was
done for Bayesian naive Bayes models in Section~\ref{sec-adj-b}, with the
difference that we condition on both $x_0\trn$ and $\z\trn$. The region 
$|\COR(x\trn_t,x_0\trn)| \leq \gamma$ can still be seen from
Figure~\ref{fig-sets}. The $P(\mathcal{S}\given x_0\trn,\z\trn,\alpha)$ is equal
to the sum of $P(I_0,I_1\given x_0\trn,\z\trn,\alpha)$ over $L_{+}\cup L_{-}\cup
L_0$, or equivalently 1 minus the sum over $H_+\cup H_-$. The probability over
$H_+$ is equal to the probability over $H_-$ since the prior for $\theta_t$ is
symmetrical about $1/2$. We therefore need to compute the probability for each
point only in either $H_+$ or $H_-$. We then exchange the summation over $H_+$
with the integration with respect to $\theta_t$. Next we discuss only how to
calculate $P(I_0,I_1\given x_0\trn,\z\trn,\theta_t,\alpha)$ for each point
$(I_0,I_1)$.

\begin{figure}[t]

\begin{center}

\begin{picture}(230,155)(0,0)

\put(20,155){$z$}

\put(-20,70){$n^{[1]}$}

\put(0,70){ $\left\{ 
             \begin{array}{c}  
             1 \\ 1 \\ 1 \\ 1 \\ [18pt] 1 \\ 1 \\ 1 \\ 1 
             \end{array} 
             \right.$ }
\put(125,155){$\x_0\trn$}

\put(90,110){$n_{0}^{[1]}$}
\put(110,110){ $\left\{ \begin{array}{c}  0 \\ 0 \\0 \\0 \end{array} \right.$ } 

\put(90,30){$n_{1}^{[1]}$}
\put(110,30){ $\left\{ \begin{array}{c}  1 \\ 1 \\ 1 \\1 \end{array} \right.$ }

\put(230,155){$\x_t\trn$}

\put(228,124.5){$\begin{array}{c}  0 \\ 0 \end{array}$ }
\put(215,94){ $\left\{ \begin{array}{c}  1 \\ 1 \end{array} \right.$ }
\put(200,94){$I_0^{[1]}$}

\put(228,45){$ \begin{array}{c}  0 \\ 0 \end{array} $ }
\put(215,15){ $\left\{\begin{array}{c}  1 \\ 1 \end{array}\right.$ }
\put(200,15){$I_1^{[1]}$}

\end{picture}

\end{center}

\vspace*{-0.2in}\caption{Notations used in deriving the adjustment factor of
Bayesian mixture models}

\label{gr}

\end{figure}

We divide the training cases according to $\z\trn$ into two groups, and let
$I_0^{[z]}= \sum_{i=1}^n I(z^{(i)}=z,x_0^{(i)}=0, x_t^{(i)}=1)$, and
$I_1^{[z]}=I(z^{(i)}=z,x_0^{(i)}=1,x_t^{(i)}=1)$, where $z=0,1$. The probability
of $(I_0^{[z]}, I_1^{[z]})$ is found by summing over all configurations of
feature $t$ that have $z^{(i)}=z$ and results in $(I_0^{[z]}, I_1^{[z]})$:

\beq
 \lefteqn{P(I_0^{[z]},I_1^{[z]} \given x_0\trn, \z\trn,\theta_t,\alpha)} \nonumber \\
 & = &\mychoose{n_0^{[z]}}{I_0^{[z]}}\mychoose{n_{1}^{[z]}}{I_1^{[z]}}
 U(\alpha\theta_t,\alpha(1-\theta_t),I_0^{[z]}+I_1^{[z]},
 n^{[z]}-(I_0^{[z]}+I_1^{[z]})\label{prob-n01-n11-z}
\eeq where $n_0^{[z]}=\sum_{i=1}^n I(z^{(i)}=z,x_0^{(i)}=0),
n_{1}^{[z]}=\sum_{i=1}^n I(z^{(i)}=z,x_0^{(i)}=1)$ and
$n^{[z]}=\sum_{i=1}^nI(z_i=z)$.

Then, the joint probability function of $(I_0,I_1)$ is found by summing over all
possible combinations of $(I_0^{[0]}, I_1^{[0]})$ and $(I_0^{[1]},I_1^{[1]})$
that result in $I_0^{[0]}+I_0^{[1]}=I_0,I_1^{[0]}+I_1^{[1]}=I_1$:

\beq
  P(I_0,I_1\given x_0\trn,\z\trn,\theta_t,\alpha) &=&
  \sum_{\scriptsize{\begin{array}{c}
    I_0^{[0]}+I_0^{[1]}=I_0\\
    I_1^{[0]}+I_1^{[1]}=I_1
       \end{array}}
      }
  \prod_{z=0}^1 P(I_0^{[z]},I_1^{[z]}\ |\  x_0\trn,\z\trn,\theta_t,\alpha)
  \label{prob-mix-n01-n11}
\eeq

The way of finding the combinations of $(I_0^{[0]}, I_1^{[0]})$ and
$(I_0^{[1]},I_1^{[1]})$ that satisfy $I_0^{[0]}+I_0^{[1]}=I_0$ and
$I_1^{[0]}+I_1^{[1]}=I_1$ is given in the Appendix $2$ to this Chapter.

\subsection{A Simulation Experiment}\label{sim}


\begin{table}[p]

\small \input{mix300-tab}


\vspace*{0.3in}\caption[Comparison of calibration for predictions found with and
without correction for selection bias, on data simulated from a binary mixture
model] {Comparison of calibration for predictions found with and without
correction for selection bias, on data simulated from a binary mixture model.
The test cases were divided into 10 categories by the first decimal of the
predictive probability of class 1, which is indicated by the 1st column ``C''. 
The table shows the number  of test cases in each category for each method
(``\#''), the average predictive probability of class 1 for cases in that
category  (``Pred''), and the actual fraction of these cases that were in class
1 (``Actual'').}

\label{tab-mix} 


\end{table}


\begin{figure}[p]

\begin{center} \includegraphics{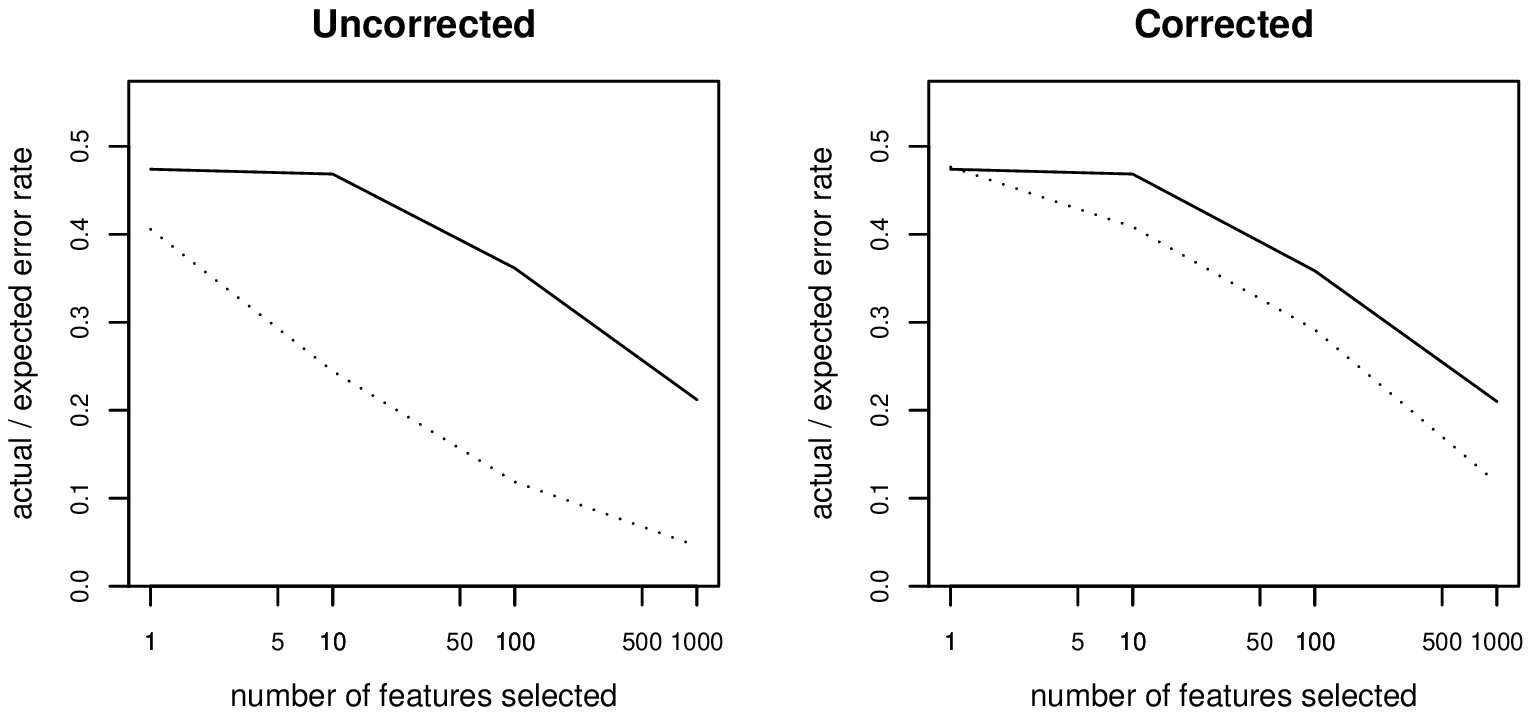} \end{center}

\vspace*{-7pt}

\caption[Actual and expected error rates with varying numbers of  features
selected]{Actual and expected error rates with varying numbers (in log scale) of  features
selected, with and without correction for selection bias. The solid line is the
actual error rate on test cases. The dotted line is the error rate that would be
expected based on the  predictive probabilities.}

\label{fig-mix-er}

\end{figure}


\begin{figure}[p]

\vspace*{-12pt}

\begin{center} \includegraphics{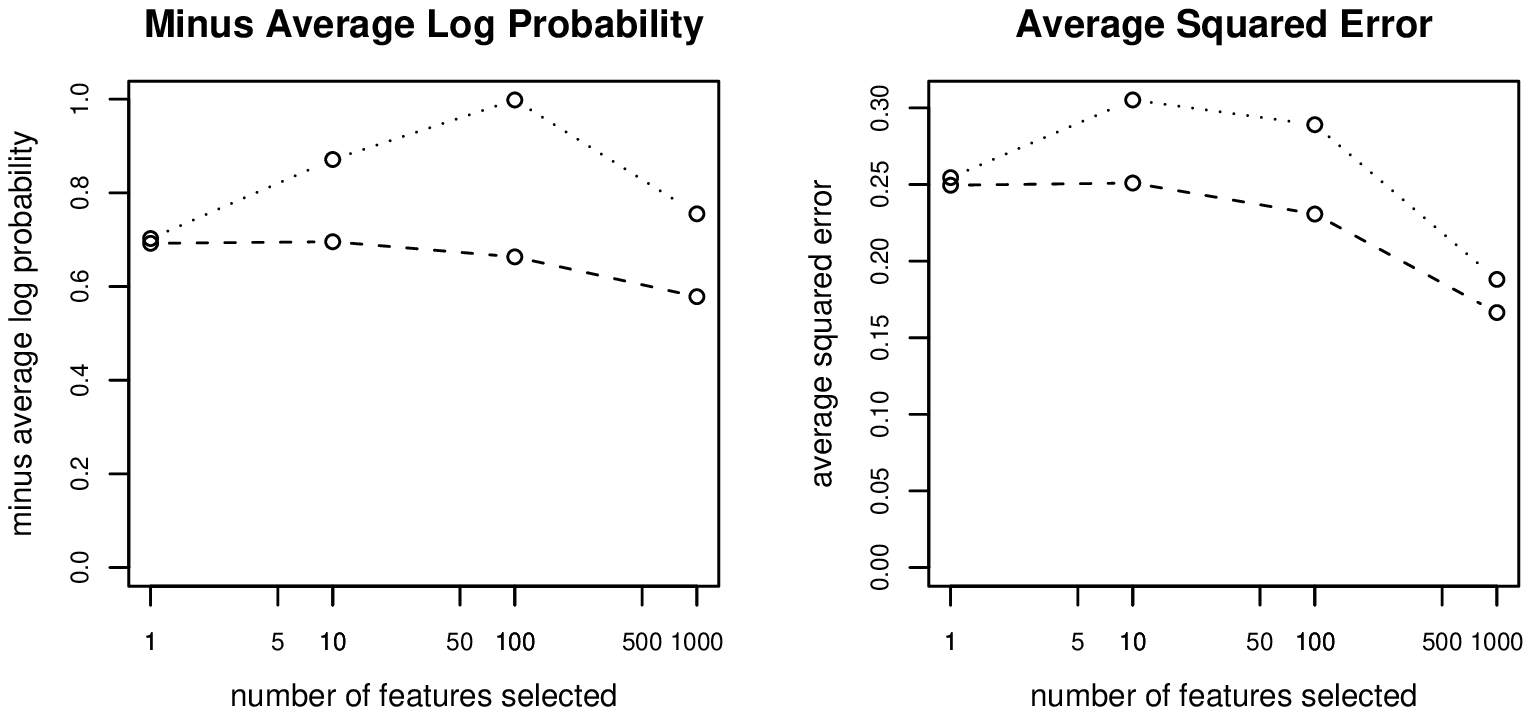} \end{center}

\vspace*{-7pt}

\caption[Performance in terms of average minus log probability and average
squared error by simulated data]{Performance in terms of average minus log
probability and average squared error, with varying numbers (in log scale) of features
selected, with and without correction for selection bias. The left plot shows
minus the average log probability of the correct class for test cases, with 1,
10, 100, and 1000 features selected. The dashed line is with bias correction,
the dotted line without.  The right plot is similar, but shows average squared
error on test cases. }\label{fig-mix-perf}

\end{figure}

We tested our method using a data with $200$ training cases and $2000$ test
cases, which are generated from a Bayesian mixture model, by setting
$\alpha=300$, $\phi_{00}=0.1$, and $\phi_{10}=0.9$, and letting the number of
$z=1$ and $z=0$ be equal in both training and test sets.

We then selected four subsets of features, containing 1, 10, 100, and 1000
features, based on the absolute values of the sample correlations of the
features with $y$.  The smallest correlation (in absolute value) of a selected
feature with the class was 0.30, 0.24, 0.18, and 0.12 for these four subsets. 
These are the values of $\gamma$ used by the bias correction method when
computing the adjustment factor.

For each number of selected features, we fit this data using the Bayesian
mixture model with the prior for $\psi$ (equation~(\ref{prior-p})) having
$f_0=f_1=1$ and the Inverse-Gamma prior for both $\alpha_0$ and $\alpha$
(equation~(\ref{prior-alpha0}) and~(\ref{prior-alpha-mix-cont})) both having
shape parameter $a=0.5$ and rate parameter $b=5$. After using Gibbs sampling to
train the model, with and without correction for the selection bias, we made
predictions for the test cases.  

We compared the predictive performance of the methods with and without
correction for selection bias in several ways. Table~\ref{tab-mix} shows how
well calibrated the predictive probabilities were, by looking at the actual
fraction of class $1$ of the test cases with predictive probabilities within each
of the ten intervals evenly spaced in $(0,1)$. This shows that the methods with
correction for selection bias are better calibrated than without correction. For
the methods with correction for selection bias, the actual fractions are closer
to predictive probabilities, whereas for the methods without such correction,
they are more different, with  predictive probabilities incorrectly close to $0$
or $1$ for many cases. But we have seen that some bias, although less severe,
still exists for the methods with correction, which we will explain later.

The calibration can also be illustrated by comparing the actual error rate, from
making predictions by thresholding the predictive probabilities at $0.5$, to the
expected error rate, equal to $(1/N) \sum_i \hat p^{(i)}I(\hat p^{(i)} < 0.5)  +
(1-\hat p^{(i)}) I(\hat p^{(i)}\ge0.5)$, where $\hat p^{(i)}$ is the predictive
probability for test case $i$. As shown by Figure~\ref{fig-mix-er}, the expected
error rates for the methods without correction for selection bias are much lower
than the actual error rates, showing that the predictions are overconfident. In
contrast, the expected error rates and the actual error rates are much closer
for the methods with correction for selection bias, though there are still gaps
between them. From Figure~\ref{fig-mix-er}, we also see that the actual error
rates for the methods with and without correction for selection bias are almost
the same. 

The remaining bias for the methods with correction for selection bias presumably
results from  Markov chain Monte Carlo method. Since the two groups are not very
apart with $\alpha=300$, the latent values $\zz$ temporarily converge to the
responses $x^{(1)}_0\,\ldots,x^{(n)}_0$, even with correction for selection
bias. The estimates of $\phi_{00}$ and $\phi_{10}$ are therefore very close to
$0$ and $1$. But the traces of $\alpha$ with correction for selection bias still
move around much bigger values in $\A$ than without correction. The
probabilities of a test case belonging to two groups with correction are
therefore closer than without correction, making it difficult to decide the
label of the test case. The correction method, implemented with simple Gibbs
sampling, reduces the selection bias, but does not eliminate it entirely. This
remaining bias will be eliminated entirely if one uses a more sophisticated
Markov chain sampling method that allows the Markov chains to explore more
thoroughly in the space of $\zz$. Note that, however, this is a matter of
computation rather than of theory. In theory, the bias will be eliminated
entirely by conditioning on all information available in making Bayesian
inference if the data sets are generated from the Bayesian model. 

The methods with and without correction for selection bias are also compared in
terms of average minus log probability and average squared error, as shown in
Figure~\ref{fig-mix-perf}. In both measures, the methods with the selection bias
corrected are superior over the methods without correction.
 
The predictive performance using the complete data was not shown in previous
table and figures, because it is ironically worse than using selected features.
This is because the Markov chain, using the simple Gibbs sampling, converges
temporarily to only one group if the initial labels are drawn randomly from the
permutations of $1,\ldots,n$. Again, a more sophisticated Markov chain sampling
method will solve this problem.

We used Gibbs sampling to train the model. In each iteration of Gibbs sampling,
we update $\alpha$ and $\alpha_0$ once, and repeat $5$ time the combination of
updating the labels $\zz$ once and updating each of $\thetab_{1:k}$ $20$ times.
In approximating the continuous priors for $\alpha_0$ and $\alpha$  with the
uniform distribution over the quantiles of the priors
(equation~(\ref{prob-pred-mix-midrule})), we chose $K=10$, giving
$\A=\A_0=\{2.60, 4.83,    7.56,   11.45,   17.52,   27.99,   48.57,   98.49, 
279.60, 2543.14 \}$. Simpson's Rule, which is used to approximate the integral
with respect to $\theta_t$ for computing the adjustment factor, evaluates the
integrand at $11$ points.

Our software (available from
\texttt{http://www.utstat.utoronto.ca/$\sim$longhai}) is written entirely in R
language. Computation times for each method (on a 2.2 GHz Opteron processor,
running $50$ iterations of Gibbs sampling as described above) are shown in
Table~\ref{mix-tab-time}. The computation of adjustment factor takes a large
amount of extra time. This is because the computation of a single adjustment
factor is more complex than naive Bayes models and the computation needs to be
redone whenever the latent values $\zz$ change. The current method for computing
the adjustment factor can still be improved. However, the methods using
selecting features and correcting for the selection bias still work faster than
using complete data, which takes about $40000$ seconds for updating $50$
iterations.


\begin{table}[t]

\begin{center}

\input{mix300-time-tab}

\end{center}

\vspace*{-0.25in}\caption{Computation times from simulation experiments with
mixture models.}

\label{mix-tab-time}

\end{table}

\section{Conclusion and Discussion}

We have proposed a Bayesian method for making well-calibrated
predictions for a response variable when using a subset of features
selected from a larger number based on some measure of dependency
between the feature and the response.  Our method results from
applying the basic principle that predictive probabilities should be
conditional on all available information --- in this case, including
the information that some features were discarded because they appear
weakly related to the response variable.  This information can only be
utilized when using a model for the joint distribution of the response
and the features, even though we are interested only in the
conditional distribution of the response given the features.

We applied this method to naive Bayes models with binary features that are
assumed to be independent conditional on the value of the binary response
(class) variable. With these models, we can compute the adjustment factor needed
to correct for selection bias. Crucially, we need only compute the probability
that a single feature will exhibit low correlation with the response, and then
raise this probability to the number of discarded features. Due to the
simplicity of naive Bayes models, the methods with the selection bias corrected
work as fast as the methods without considering this corrrection. Substantial
computation time can therefore be saved by discarding features that appear to
have little relationship with the response.  

We also applied this method to mixture models for binary data. The computation
of the adjustment factor is more complex than for naive Bayes models, and it
needs to be computed many times. But the method is still feasible, and will be
faster than using all features when the number of available features is huge.

The practical utility of the bias correction method we describe would be much
improved if methods for more efficiently computing the required adjustment
factor could be found, which could be applied to a wide class of models.

\section*{Appendix 1:\\ Proof of the well-calibration of the Bayesian
Prediction}

\sectionmark{Appendix}

\setcounter{subsubsection}{0}

\addcontentsline{toc}{section}{Appendix 1: Proof of the well-calibration of the
Bayesian Prediction}

Suppose we are interested in predicting whether a random vector $\mb Y$ is in a
set $\A$ if {\it we} know the value of another random vector $\mb X$. Here, $\mb
X$ is all the information we know for predicting $\mb Y$, such as the
information from the training data and the feature values of a test case. And
$\mb Y$ could be any unknown quantity, for example a model parameters or the
unknown response of a test case. For discrete $\mb Y$, $\A$ may contain only a
single value; for continuous $\mb Y$, it is a set such that the probability of
$\mb Y \in \A$ is not $0$ (otherwise any predictive method giving predictive
probability $0$ is well-calibrated). From a Bayesian model for $\mb X$ and $\mb
Y$, we can derive a marginal joint distribution for $\mb X$ and $\mb Y$, $P(\mb
X,\mb Y)$ (which may be a probability function or a density function, or a
combination of probability and density function), by integrating over the prior
for the model parameters.

Let us denote a series of independent experiments from $P(\mb X,\mb Y)$ as $(\mb
X_i,\mb Y_i)$, for $i=1,2,\ldots$ . Suppose a predictive method  predicts that
event $\mb Y\in\A$ will occur with probability $\hat Y(\x)$ after seeing $\mb
X=\x$. $\hat Y(\x)$ is said to be well-calibrated if, for any two numbers
$c_1,c_2\in(0,1)$ (assuming $c_1<c_2$) such that $P(\,\hat Y(\mb X_i)\in
(c_1,c_2)\,)\not=0$, the fraction of $\mb Y_i\in \A$ among those experiments
with predictive probability, $\hat Y(\mb X_i)$, between $c_1$ and $c_2$, will be
equal to the average of the predictive proabilities (with $P$-probability $1$), 
when the number of experiments, $k$, goes to $\infty$, that is,

\beq \frac{\sum_{i=1}^k I(\,\mb Y_i\in \A \mbox{ and } \hat Y(\mb X_i)\in
(c_1,c_2)\,)}{\sum_{i=1}^k I(\,\hat Y(\mb X_i)\in (c_1,c_2)\,)} -
\frac{\sum_{i=1}^k \hat Y(\mb X_i)\,I(\,\hat Y(\mb X_i)\in
(c_1,c_2)\,)}{\sum_{i=1}^k I(\,\hat Y(\mb X_i)\in (c_1,c_2)\,)} \longrightarrow
0 \label{eqn-cal} \eeq 

\noindent This definition of well-calibration is a special case for $iid$
experiments of what is defined in~(Dawid 1982). Note that this concept of
calibration is with respect to averaging over both the data and the parameters
drawn from the prior.

We will show that under the above definition of calibration, the Bayesian
predictive function $\hat Y(\x)=P(\mb Y \in \A \given X=\x)$ is well-calibrated.

First, from the strong law of large numbers, the left-hand of~(\ref{eqn-cal})
converges to:

\beq \frac{P(\mb Y\in\A\mbox{ and } \hat Y(\mb X)\in (c_1,c_2))}{P(\hat Y(\mb
X)\in (c_1,c_2))}-\frac{E(\,\hat Y(\mb
X)\,I(\hat Y(\mb X)\in (c_1,c_2)\,)\,)}{P(\hat
Y(\mb X)\in (c_1,c_2))} \label{eqn-cal-prob}\eeq 

\noindent We then need only show that the expression~(\ref{eqn-cal-prob}) is
actually equal to $0$, i.e., the numerators in two terms are the same. This
equality can be shown as follows:

\beq \lefteqn{P(\mb Y\in\A\mbox{ and } \hat Y(\mb X)\in (c_1,c_2))}\nonumber \\ 
&=&\int I(\,\hat Y(\mb
x)\in (c_1,c_2))\,P(\mb Y\in\A\given \mb X=\x)P_{\scriptsize \mb X}(\x)\,d\x
\label{eqn-proof1} \\
&=&\int I(\,\hat Y(\mb
x)\in (c_1,c_2))\,\hat Y(\x)\,P_{\scriptsize \mb X}(\x)\,d\x 
\label{eqn-proof2}\\
&=&E(\,\hat Y(\mb
X)\,I(\hat Y(\mb X)\in (c_1,c_2)\,)\,)
\eeq

\noindent What is essential from~(\ref{eqn-proof1}) to~(\ref{eqn-proof2}) is
that the Bayesian predictive function~$\hat Y(\x)$ is just the conditional
probability $P(\mb Y\in\A\given \mb X=\x)$.

The Bayesian predictive function $\hat Y(\x)=P(\mb Y \in \A \given X=\x)$ also
has the following property, which is helpful in understanding the concept of
well-calibration: 

\beq  P(\mb Y\in\A\given \hat Y(\mb X)\in (c_1,c_2)) = E(\hat Y(\mb X)\given \hat Y(\mb X)\in (c_1,c_2))&\in&
(c_1,c_2)\label{cal-interval}\eeq

\noindent  $P(\mb Y\in\A\given \hat Y(\mb X)\in (c_1,c_2))$ is just the first
term in~(\ref{eqn-cal-prob}), and is equal to the second term
in~(\ref{eqn-cal-prob}), which can be written as $E(\hat Y(\mb X)\given \hat
Y(\mb X)\in (c_1,c_2))$. This conditional expectation is obviously between $c_1$
and $c_2$. 

\section*{Appendix 2:\\ Details of the Computation of the Adjustment Factor for
Binary Mixture Models}\vspace*{-4pt}

\addcontentsline{toc}{section}{Appendix 2: Details of the Computation of the
Adjustment Factor for Binary Mixture Models}

\sectionmark{Appendix}

\setcounter{subsubsection}{0}

\subsubsection {\bf Delineating $H_+$}

With the monotonicity of $\Cor(I_1,I_0;x_0\trn)$ with respect to either $I_1$ or
$I_0$, we can easily determine the bound of $I_1$ for each $I_0$ that satisfies
$\Cor(I_1,I_0;x) \leq \gamma$ by solving the equation: \beq \
|\Cor(I_0,I_1;x_0\trn)|\  = \gamma \label{coreta} \eeq then rounding to the
appropriate integers and truncating them by $0$ and $n_{1}$.  I.e. the lower
bound is $\max(0,\ceil{l})$ and the upper bound is $\min(n_{1},\floor{u})$,
where $l$ and $u$ are the two solutions of equation (\ref{coreta}). When $I_0
=0$, if $I_1$ is also 0, we assume the correlation of $x_0\trn$ and $x_t\trn$ is
0, therefore the lower bound is set to be 0. Similarly, when $I_0=n_{0}$, the
upper bound is set to be $n_{1}$.

\subsubsection {Computation of $P(I_0^{[z]},I_1^{[z]}\ |\ x_0\trn, \z\trn,
\theta_t,\alpha)$ with formula (\ref{prob-n01-n11-z})}
\label{sec-appendix-table}

We need to calculate $P(I_0^{[z]},I_1^{[z]}\ |\ x_0\trn, \z\trn,
\theta_t,\alpha)$ for all $I_0^{[z]} \in\{0,\cdots,n_0^{[z]}\}$ and $I_1^{[z]}
\in\{0,\cdots,n_{1}^{[z]}\}$. These values are saved with a matrix $T^{[z]}$ for
convenience. We don't have to evaluate each element of $T^{[z]}$ by noting that
the third factor of equation (\ref{prob-mix-n01-n11}) depends only on
$I_0^{[z]}+I_1^{[z]}$, i.e. the number of $x_t\trn=1$. For each $k
\in\{0,1,\cdots,n_0^{[z]} + n_{1}^{[z]} \}$, we need to evaluate it only once.
Then go along the diagonal line $I_0^{[z]}+I_1^{[z]} =k$ to obtain the elements
of $T^{[z]}$. The lower bound of $I_1^{[z]}$ on this line is
$\max(0,k-n_0^{[z]})$ and the upper bound is $\min(k,n_{1}^{[z]})$. For each
$I_1^{[z]}$ between the lower and upper bounds, correspondingly
$I_0^{[z]}=k-I_1^{[z]}$.

For each line associated with $k$, only when $I_1^{[z]}=\max(0,k-n_0^{[z]})$ we
need to evaluate (\ref{prob-mix-n01-n11}), then we can obtain the remaining
values using the following relation between two successive elements on the
line: 

\beq
 \frac{P(I_0^{[z]},I_1^{[z]}\ |\ x_0\trn,\z\trn, \alpha,\theta_t)}
    {P(I_0^{[z]}+1,I_1^{[z]}-1\ |\ x_0\trn,\z\trn, \alpha,\theta_t)}
 & = & \frac{ \mychoose{n_0^{[z]} } {I_0^{[z]}}
            \mychoose{n_{1}^{[z]}}{I_1^{[z]}}
      }
      {\mychoose{n_0^{[z]}}{I_0^{[z]}+1 }
            \mychoose{n_{1}^{[z]}}{I_1^{[z]}-1 }
      }\\
 & =& \frac{(I_0^{[z]}+1)(n_{1}^{[z]}-I_1^{[z]}+1)}
     {(n_0^{[z]}-I_0^{[z]})I_1^{[z]}}
\eeq

\subsubsection {\bf Computation of $P(I_0,I_1\given
x_0\trn,\z\trn,\theta_t,\alpha)$ with formula (\ref{prob-mix-n01-n11})}

For each $(I_0,I_1) \in H_+$, we need find all the pairs of
$(I_0^{[0]},I_1^{[0]})$ and $( I_0^{[1]},I_1^{[1]})$ that satisfy 

\beq
  I_0^{[0]}+I_0^{[1]} & = & I_0,
  \mbox{and }0 \leq I_0^{[0]} \leq n_{0}^{[0]}, 0 \leq I_0^{[1]} \leq n_{0}^{[1]}
  \label{dn01}\\
  I_1^{[0]}+I_1^{[1]} & = & I_1,
  \mbox{and }0 \leq I_1^{[0]} \leq n_{1}^{[0]}, 0 \leq I_1^{[1]} \leq n_{1}^{[1]}
  \label{dn11}
\eeq

The decompositions of $I_0$ and $I_1$ are independent and the methods are
identical. Taking $I_0$ as example, we determine the bound of $I_0^{[0]}$ and
$I_0^{[1]}$ by truncating the straight line of $I_0^{[0]}+I_0^{[1]}=I_0$ with
the square determined by $(0,n_0^{[0]})\times (0,n_0^{[1]})$. By this way, we
obtain the decompositions as follows:

 \beq
  I_0^{[0]} \in \{\max(0,I_0-n_{0}^{[1]}),\cdots,\min(n_{0}^{[0]},I_0)\}
  \equiv B_{01}^{[0]} \mbox{, and } I_0^{[1]} = I_0-I_0^{[0]} \label{dn012}
  \\
  I_1^{[0]} \in \{\max(0,I_1-n_{1}^{[1]}),\cdots,\min(n_{1}^{[0]},I_1)\}
  \equiv B_{11}^{[0]} \mbox{, and } I_1^{[1]} =
  I_1-I_1^{[0]}\label{dn112}
\eeq

We make a sub-matrix $S^{[0]}$ from $T^{[0]}$, which has been computed in
Section \ref{sec-appendix-table}, by taking the rows in $B_{01}^{[0]}$ and the
columns in $B_{11}^{[0]}$, and accordingly make $S^{[1]}$ from $T^{[1]}$ by
taking the rows in $I_0- B_{01}^{[0]}$ and the columns in $I_1-B_{11}^{[0]}$.
Then multiplying $S^{[0]}$ and $S^{[1]}$ element by element (i.e. the
corresponding elements of $S^{[0]}$ and $S^{[1]}$ are multiplied together) makes
matrix $S$. Summing all the elements of $S$ together yields $P(I_0,I_1\given
x_0\trn,\z\trn,\theta_t,\alpha)$.

%% file: tab_naive.tex
\begin{tabular}
{@{}c|@{}rcc|@{}rcc|rcc|rcc@{}}
         & \multicolumn{6}{c|}{\bf  1 feature selected out of 10000 } & \multicolumn{6}{c}{\bf  10 features selected out of 10000 } \\[5pt] 
         & \multicolumn{3}{c}{Corrected} & \multicolumn{3}{c|}{Uncorrected} & \multicolumn{3}{c}{Corrected} & \multicolumn{3}{c}{Uncorrected} \\[5pt] 
C & \mbox{~~~}\#\mbox{~} & \small{Pred} & \small{Actual} & \mbox{~~~}\#\mbox{~} & \small{Pred} & \small{Actual} & \mbox{~~}\#\mbox{~} & \small{Pred} & \small{Actual} & \mbox{~~}\#\mbox{~} & \small{Pred} & \small{Actual} \\[5pt] 
0 &    0 &    -- &    -- &    0 &    -- &    -- &    0 &    -- &    -- &  237 & 0.046 & 0.312 \\ 
1 &    0 &    -- &    -- &    0 &    -- &    -- &    3 & 0.174 & 0.000 &  349 & 0.149 & 0.444 \\ 
2 &    0 &    -- &    -- &    0 &    -- &    -- &  126 & 0.270 & 0.294 &   68 & 0.249 & 0.500 \\ 
3 &    0 &    -- &    -- & 1346 & 0.384 & 0.461 &  467 & 0.360 & 0.420 &  300 & 0.360 & 0.443 \\ 
4 & 1346 & 0.446 & 0.461 &    0 &    -- &    -- &  566 & 0.462 & 0.461 &  189 & 0.443 & 0.487 \\ 
5 &    0 &    -- &    -- &    0 &    -- &    -- &  461 & 0.554 & 0.566 &   48 & 0.546 & 0.417 \\ 
6 &  654 & 0.611 & 0.581 &    0 &    -- &    -- &  276 & 0.643 & 0.616 &  238 & 0.650 & 0.588 \\ 
7 &    0 &    -- &    -- &  654 & 0.736 & 0.581 &   97 & 0.733 & 0.742 &  180 & 0.737 & 0.567 \\ 
8 &    0 &    -- &    -- &    0 &    -- &    -- &    4 & 0.825 & 0.750 &  192 & 0.864 & 0.609 \\ 
9 &    0 &    -- &    -- &    0 &    -- &    -- &    0 &    -- &    -- &  199 & 0.943 & 0.668 \\ 
\end{tabular}

\vspace*{0.45in}
\begin{tabular}
{@{}c|@{}rcc|@{}rcc|rcc|rcc@{}}
         & \multicolumn{6}{c|}{\bf  100 features selected out of 10000 } & \multicolumn{6}{c}{\bf  1000 features selected out of 10000 } \\[5pt] 
         & \multicolumn{3}{c}{Corrected} & \multicolumn{3}{c|}{Uncorrected} & \multicolumn{3}{c}{Corrected} & \multicolumn{3}{c}{Uncorrected} \\[5pt] 
C & \mbox{~~~}\#\mbox{~} & \small{Pred} & \small{Actual} & \mbox{~~~}\#\mbox{~} & \small{Pred} & \small{Actual} & \mbox{~~}\#\mbox{~} & \small{Pred} & \small{Actual} & \mbox{~~}\#\mbox{~} & \small{Pred} & \small{Actual} \\[5pt] 
0 &  155 & 0.067 & 0.077 &  717 & 0.017 & 0.199 &  774 & 0.018 & 0.027 &  954 & 0.004 & 0.066 \\ 
1 &  247 & 0.151 & 0.162 &  133 & 0.150 & 0.391 &   97 & 0.143 & 0.165 &   28 & 0.149 & 0.500 \\ 
2 &  220 & 0.247 & 0.286 &   70 & 0.251 & 0.429 &   63 & 0.243 & 0.302 &   13 & 0.248 & 0.846 \\ 
3 &  225 & 0.352 & 0.356 &   68 & 0.351 & 0.515 &   48 & 0.346 & 0.438 &   17 & 0.349 & 0.412 \\ 
4 &  237 & 0.450 & 0.494 &   58 & 0.451 & 0.500 &   45 & 0.446 & 0.600 &   14 & 0.449 & 0.786 \\ 
5 &  227 & 0.545 & 0.586 &   78 & 0.552 & 0.603 &   44 & 0.547 & 0.614 &   16 & 0.546 & 0.375 \\ 
6 &  202 & 0.650 & 0.728 &   77 & 0.654 & 0.532 &   53 & 0.647 & 0.698 &   16 & 0.667 & 0.812 \\ 
7 &  214 & 0.749 & 0.785 &   80 & 0.746 & 0.662 &   81 & 0.755 & 0.815 &   22 & 0.751 & 0.636 \\ 
8 &  182 & 0.847 & 0.857 &   98 & 0.852 & 0.633 &  124 & 0.854 & 0.863 &   25 & 0.865 & 0.560 \\ 
9 &   91 & 0.935 & 0.923 &  621 & 0.979 & 0.818 &  671 & 0.977 & 0.982 &  895 & 0.995 & 0.946 \\ 
\end{tabular}

\vspace*{0.45in}
\begin{tabular}[t]
{@{}c|@{}rcc@{}}
         & \multicolumn{3}{c}{\bf Complete data} \\[5pt] 
C & \mbox{~~~}\#\mbox{~} & \small{Pred} & \small{Actual} \\[5pt] 
0 &  964 & 0.004 & 0.006 \\ 
1 &   21 & 0.145 & 0.238 \\ 
2 &    8 & 0.246 & 0.375 \\ 
3 &   10 & 0.342 & 0.300 \\ 
4 &   12 & 0.436 & 0.500 \\ 
5 &    7 & 0.544 & 1.000 \\ 
6 &   20 & 0.656 & 1.000 \\ 
7 &   13 & 0.743 & 0.846 \\ 
8 &   22 & 0.851 & 0.818 \\ 
9 &  923 & 0.994 & 0.998 \\ 
\end{tabular}

%% file: naive-time-tab.tex
\begin{tabular}{c@{\ \ }@{\ \ }ccccc}
Number of Features Selected&  1      &  10   & 100   &1000   & 
Complete data \\[5pt]
\hline \\
Uncorrected Method & 11 & 19 & 107 & 1057 & 10639 \\[5pt]
Corrected  Method& 12 & 19 & 107 & 1057 & 10639 \\[5pt]
\end{tabular}

%% file: mix300-tab.tex
\begin{tabular}
{@{}c|@{}rcc|@{}rcc|rcc|rcc@{}}
         & \multicolumn{6}{c|}{\bf  1 feature selected out of 10000 } & \multicolumn{6}{c}{\bf  10 features selected out of 10000 } \\[5pt] 
         & \multicolumn{3}{c}{Corrected} & \multicolumn{3}{c|}{Uncorrected} & \multicolumn{3}{c}{Corrected} & \multicolumn{3}{c}{Uncorrected} \\[5pt] 
C & \mbox{~~~}\#\mbox{~} & \small{Pred} & \small{Actual} & \mbox{~~~}\#\mbox{~} & \small{Pred} & \small{Actual} & \mbox{~~}\#\mbox{~} & \small{Pred} & \small{Actual} & \mbox{~~}\#\mbox{~} & \small{Pred} & \small{Actual} \\[5pt] 
0 &    0 &    -- &    -- &    0 &    -- &    -- &    0 &    -- &    -- &   19 & 0.089 & 0.105 \\ 
1 &    0 &    -- &    -- &    0 &    -- &    -- &    0 &    -- &    -- &  340 & 0.155 & 0.415 \\ 
2 &    0 &    -- &    -- &    0 &    -- &    -- &   19 & 0.259 & 0.105 &   45 & 0.269 & 0.622 \\ 
3 &    0 &    -- &    -- &    0 &    -- &    -- &  317 & 0.368 & 0.420 &  406 & 0.359 & 0.480 \\ 
4 & 1083 & 0.488 & 0.472 & 1083 & 0.424 & 0.472 &  530 & 0.466 & 0.494 &   52 & 0.424 & 0.558 \\ 
5 &  917 & 0.536 & 0.523 &    0 &    -- &    -- &  552 & 0.549 & 0.500 &   54 & 0.560 & 0.407 \\ 
6 &    0 &    -- &    -- &  917 & 0.617 & 0.523 &  480 & 0.639 & 0.529 &  443 & 0.649 & 0.519 \\ 
7 &    0 &    -- &    -- &    0 &    -- &    -- &  100 & 0.735 & 0.620 &   49 & 0.735 & 0.449 \\ 
8 &    0 &    -- &    -- &    0 &    -- &    -- &    2 & 0.832 & 1.000 &  329 & 0.854 & 0.505 \\ 
9 &    0 &    -- &    -- &    0 &    -- &    -- &    0 &    -- &    -- &  263 & 0.945 & 0.593 \\ 
\end{tabular}

\vspace*{0.45in}
\begin{tabular}
{@{}c|@{}rcc|@{}rcc|rcc|rcc@{}}
         & \multicolumn{6}{c|}{\bf  100 features selected out of 10000 } & \multicolumn{6}{c}{\bf  1000 features selected out of 10000 } \\[5pt] 
         & \multicolumn{3}{c}{Corrected} & \multicolumn{3}{c|}{Uncorrected} & \multicolumn{3}{c}{Corrected} & \multicolumn{3}{c}{Uncorrected} \\[5pt] 
C & \mbox{~~~}\#\mbox{~} & \small{Pred} & \small{Actual} & \mbox{~~~}\#\mbox{~} & \small{Pred} & \small{Actual} & \mbox{~~}\#\mbox{~} & \small{Pred} & \small{Actual} & \mbox{~~}\#\mbox{~} & \small{Pred} & \small{Actual} \\[5pt] 
0 &   71 & 0.072 & 0.183 &  605 & 0.033 & 0.286 &  692 & 0.033 & 0.140 &  919 & 0.016 & 0.185 \\ 
1 &  195 & 0.154 & 0.308 &  122 & 0.144 & 0.361 &  129 & 0.148 & 0.271 &   33 & 0.145 & 0.455 \\ 
2 &  237 & 0.250 & 0.300 &   86 & 0.246 & 0.465 &   88 & 0.247 & 0.375 &   28 & 0.240 & 0.429 \\ 
3 &  229 & 0.349 & 0.328 &   63 & 0.350 & 0.508 &   64 & 0.350 & 0.500 &   21 & 0.350 & 0.619 \\ 
4 &  234 & 0.454 & 0.504 &   62 & 0.448 & 0.597 &   68 & 0.454 & 0.426 &   20 & 0.441 & 0.400 \\ 
5 &  259 & 0.549 & 0.556 &   90 & 0.551 & 0.589 &   71 & 0.548 & 0.634 &   25 & 0.552 & 0.480 \\ 
6 &  253 & 0.652 & 0.565 &   66 & 0.650 & 0.530 &   69 & 0.646 & 0.580 &   20 & 0.648 & 0.700 \\ 
7 &  251 & 0.748 & 0.673 &   87 & 0.749 & 0.529 &   83 & 0.744 & 0.795 &   31 & 0.749 & 0.645 \\ 
8 &  192 & 0.848 & 0.729 &  140 & 0.856 & 0.564 &  143 & 0.857 & 0.804 &   44 & 0.856 & 0.545 \\ 
9 &   79 & 0.928 & 0.734 &  679 & 0.965 & 0.666 &  593 & 0.966 & 0.841 &  859 & 0.980 & 0.818 \\ 
\end{tabular}

%% file: mix300-time-tab.tex
\begin{tabular}{c@{\ \ }@{\ \ }rrrr}
Number of Features Selected&  1      &  10   & 100   &1000\\[5pt]
\hline \\
Uncorrected Method & 70 & 82 & 193 & 1277 \\[5pt]
Corrected  Method  & 3324 & 2648 & 2881 & 4300 \\[5pt]
\end{tabular}

%% file: chapter3.tex
\renewcommand{\chaptername}{Chapter}
\renewcommand{\thechapter}{3}

\chapter{Compressing Parameters in Bayesian Models  with High-order
Interactions}

\renewcommand{\chaptername}{Compressing Parameters in Bayesian Models with
High-order Interactions}

\doublespacing \vspace*{-0.4in} {\bf Abstract.} Bayesian regression and
classification with high order interactions is largely infeasible because Markov
chain Monte Carlo (MCMC) would need to be applied with a huge number of
parameters, which typically increases exponentially with the order. In this
chapter we show how to make it feasible by effectively reducing the number of
parameters, exploiting the fact that many interactions have the same values for
all  training cases. Our method uses a single ``compressed'' parameter to
represent the sum of all parameters associated with a set of patterns that have
the same value for all training cases. Using symmetric stable distributions as
the priors of  the original parameters, we can easily find the priors of these
compressed parameters. We therefore need to deal only  with a much smaller
number of compressed parameters when training the model with MCMC. The number of
compressed parameters may have converged before considering the highest possible
order. After training the model, we can split these compressed parameters into
the original ones as needed to make predictions for test cases. We show in
detail how to compress parameters for logistic sequence prediction and logistic
classification models. Experiments on both simulated and real data demonstrate
that a huge number of parameters can  indeed be reduced by our compression
method.

\newpage

\doublespacing

\section{Introduction}

In many regression and classification problems, the response variable $y$
depends on  high-order interactions of ``features'' (also called ``covariates'',
``inputs'', ``predictor variables'', or ``explanatory variables''). Some complex
human diseases are found to be related to high-order interactions of
susceptibility genes and environmental exposures (Ritchie et. al. 2001).  The
prediction of the next character in English text  is improved by using a large
number of preceding characters (Bell, Cleary and Witten 1990). Many biological
sequences have long-memory properties.

When the features are discrete, we can employ high-order interactions in
regression and classification models by introducing, as additional predictor
variables, the indicators for each possible interaction pattern, equal to $1$ if
the pattern  occurs for a subject and $0$ otherwise. In this chapter we will use
``features'' for the original discrete measurements and ``predictor variables''
for these derived variables, to distinguish them. The number of such predictor
variables increases exponentially with the order of interactions. The total
number of order-$k$ interaction patterns with $k$ binary (0/1) features is
$2^k$, accordingly we will have $2^k$ predictor variables. A model with
interactions of even a moderate order is prohibitive in real applications, 
primarily for computational reasons. People are often forced to use a model with
very small order, say only $1$ or $2$, which, however, may omit useful
high-order predictor variables.

Besides the computational considerations, regression and classification with a
great many predictor variables may ``overfit'' the data. Unless the number of
training cases is much larger than the number of predictor variables the model
may fit the noise instead of the signal in the data, with the result that
predictions for new test cases are poor. This problem can be solved by using
Bayesian modeling with appropriate prior distributions. In a Bayesian model, we
use a probability distribution over parameters to express our prior belief about
which configurations of parameters may be appropriate. One such prior belief is
that a parsimonious model can approximate the reality well. In particular, we
may believe that most high-order interactions are largely irrelevant to
predicting the response. We express such a prior by assigning each regression
coefficient a distribution with mode $0$, such as a Gaussian or Cauchy
distribution centered at $0$. Due to its heavy tail, a Cauchy distribution may
be more appropriate than a Gaussian distribution to express the prior belief
that almost all coefficients of high order interactions are close to $0$, with a
very small number of exceptions. Additionally, the priors we use for the widths
of Gaussian or Cauchy distributions for higher order interaction should favor
small values. The resulting joint prior for all coefficients favors a model with
most coefficients close to $0$, that is, a model emphasizing  low order
interactions. By incorporating such prior information into our inference, we
will not overfit the data with an unnecessarily complex model. 

However, the computational difficulty with a huge number of parameters is even
more pronounced for a Bayesian approach than other approaches, if we have to use
Markov chain Monte Carlo methods to sample from the posterior distribution,
which is computationally burdensome even for a moderate number of parameters.
With more parameters, a Markov chain sampler will take longer for each iteration
and require more memory, and may need more iterations to converge or get trapped
more easily in local modes. Applying Markov chain Monte Carlo methods to
regression and classification with high-order interactions therefore seems
infeasible.

In this chapter, we show how these problems can be solved by effectively
reducing the number of parameters in a Bayesian model with high-order
interactions, using the fact that in a model that uses all interaction patterns,
from a low order to a high order, many predictor variables have the same values
for all training cases. For example, if an interaction pattern occurs in only
one training case, all the interaction patterns of higher order contained in it
will also occur in only that case and have the same values for all training
cases --- $1$ for that training case and $0$ for all others. Consequently, only
the sum of the coefficients associated with these predictor variables matters in
the likelihood function. We can therefore use a single ``compressed'' parameter
to represent the sum of the regression coefficients for a group of predictor
variables that have the same values in training cases. For models with very high
order of interactions, the number of such compressed parameters will be much
smaller than the number of original parameters. If the priors for the original
parameters are symmetric stable distributions, such as Gaussian or Cauchy, we
can easily find the prior distributions of these compressed parameters, as they
are also  symmetric stable distributions of the same type. In training the model
with Markov chain Monte Carlo methods we need to deal only with these compressed
parameters.  After training the model, the compressed parameters can be  split
into the original ones as needed to  make predictions for test cases.  Using our
method for compressing parameters, one can handle Bayesian regression and
classification problems with very high order of interactions in a reasonable
amount of time.

This chapter will be organized as follows. We first describe Bayesian logistic sequence prediction models and Bayesian logistic classification models to which our compression method can be applied. Then, in Section~\ref{sec-all-comp} we describe in general terms the method of compressing parameters, and how to split them to make predictions for test cases. We then apply the method to logistic sequence models in Section~\ref{sec-blsm}, and to logistic classification models in Section~\ref{sec-blcm}. There, we will describe the specific schemes for compressing parameters for these models, and use simulated data and real data to demonstrate our method. We draw conclusions and discuss future work in Section~\ref{sec-comp-conclude}.

\section{Two Models with High-order Interactions}\label{sec-models}

\subsection{Bayesian Logistic Sequence Prediction Models}
\label{sec-def-blsm}

\begin{figure}[t]

\begin{center}

\includegraphics[scale=0.85]{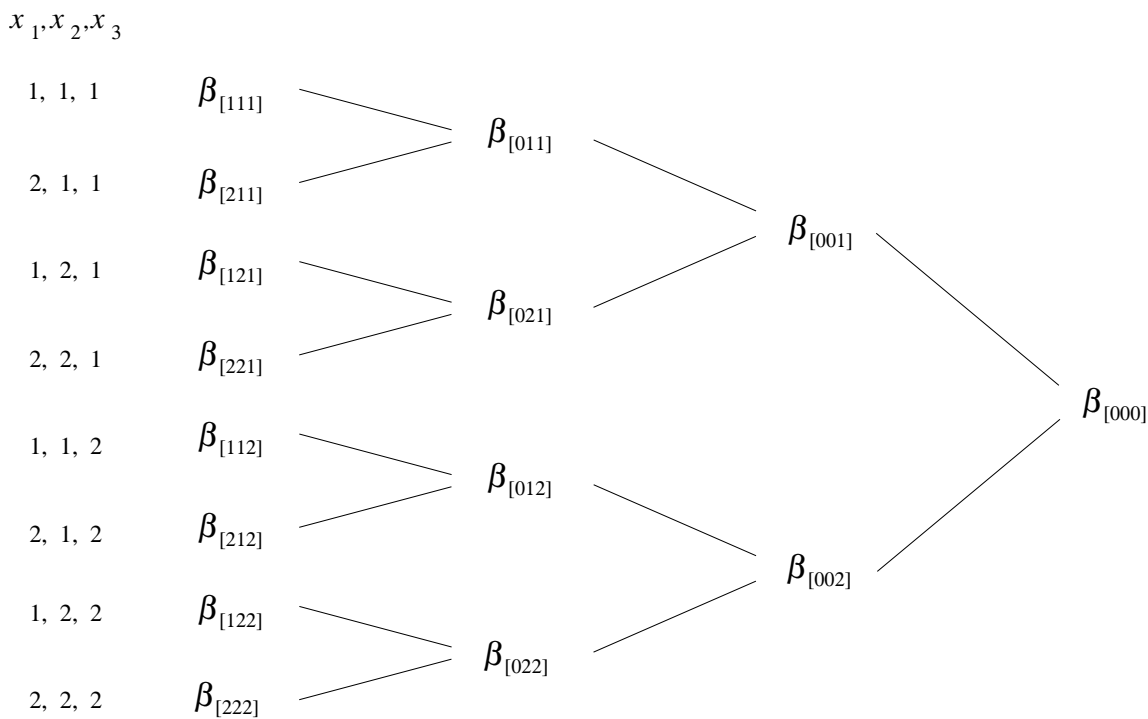}

\end{center}

\caption[A picture of the coefficients, $\mb \beta$, for all patterns  in 
binary sequences of length $O=3$.]{ A picture of the coefficients, $\mb \beta$,
for all patterns  in  binary sequences of length $O=3$. $\beta_{[A_1A_2A_3]}$ is
associated with the pattern written as $[A_1A_2A_3]$, with $A_t=0$ meaning that
$x_t$ is allowed to be either $1$ or $2$, in other words, $x_t$ is ignored in
defining this pattern. For example, $\beta_{[000]}$ is the intercept term. These
coefficients are used in defining the linear function $l\,((x_1,x_2,x_3),\mb
\beta)$ in the logistic model~(\ref{eqn-softmax}). For each combination of
$(x_1,x_2,x_3)$ on the left column, $l\,((x_1,x_2,x_3),\mb \beta)$ is equal to
the sum of $\beta$'s along the path linked by lines, from $\beta_{[x_1 x_2
x_3]}$ to $\beta_{[000]}$. }

\label{fig-seq}

\end{figure}

We often need to predict the next state of a sequence given its preceding states, for example in speech recognition (Jelinek 1998), in text compression (Bell, Cleary, and Witten 1990), and in many others. We write a sequence of length $O+1$ as $x_1,\ldots,x_O,x_{O+1}$, where $x_t$ takes values from $1$ to $K_t$, for $t=1,\ldots,O$, and $x_{O+1}$ takes values from $1$ to $K$. We call $x_1,\ldots,x_O = \x_{1:O}$ the historic sequence. For subject $i$ we write its historic sequence and response as $\mb x_{1:O}^{(i)}$ and $x^{(i)}_{O+1}$. We are interested in modelling the conditional distribution $P(x_{O+1}\given \mb x_{1:O})$.

An interaction pattern $\P$ is written as $[A_1A_2\ldots A_O]$, where $A_t$ can
be from $0$ to $K_t$, with $A_t=0$ meaning that $x_t$ can be any value from $1$
to $K_t$. For example, $[0\ldots 01]$ denotes the pattern that fixes $x_O=1$ and
allows $x_1,\ldots,x_{O-1}$ to be any values in their ranges.  When all nonzero
elements of $\P$ are equal to the corresponding elements of a historic sequence,
$\mb x_{1:O}$, we say that pattern $\P$ occurs in $\mb x_{1:O}$, or pattern $\P$
is expressed by $\mb x_{1:O}$, denoted by $\mb x_{1:O}\in \P$. We will use  the
indicator $I(x_{1:O}\in \P)$ as a predictor variable, whose coefficient is
denoted by $\beta_{\P}$. For example, $\beta_{[0\cdots0]}$ is the intercept
term. A logistic model assigns each possible value of the response a linear
function of the predictor variables. We use $\beta^{(k)}_{\P}$ to denote the
coefficient associated with pattern $\P$ and used in the linear function for 
$x_{O+1} = k$.

For modeling sequences, we consider only the patterns where all zeros (if any)
are at the start.  Let us denote all such patterns by $\mb {\mathcal{S}}$. We
write all coefficients for $x_{O+1}=k$, i.e.,  $\left\{\beta^{(k)}_{\P} \given
\P \in \mb {\mathcal{S}}\right\}$, collectively as $\mb \beta^{(k)}$.
Figure~(\ref{fig-seq}) displays $\mb \beta^{(k)}$  for binary sequence of length
$O=3$, for some $k$, placed in a tree-shape.

Conditional on $\mb \beta^{(1)},\ldots,\mb \beta^{(K)}$ and $\mb x_{1:O}$, the
distribution of $x_{O+1}$ is defined as

\beq
P(x_{O+1} = k \given \mb x_{1:O}, \mb \beta^{(1)},\ldots,\mb \beta^{(K)} ) =
\frac{\exp( l\,(\x_{1:O},\mb\beta^{(k)}) ) }
     {\sum_{j=1}^K \exp( l\,(\x_{1:O},\mb\beta^{(j)}) ) }
\label{eqn-softmax}
\eeq

\noindent where

\beq
l\,(\x_{1:O},\mb\beta^{(k)})
&=& \sum_{\P \in \mb {\mathcal{S}}}\beta^{(k)}_{\P}\ I(\x_{1:O}\in \P)=
     \beta^{(k)}_{[0\cdots0]} +
    \sum_{t=1}^O \beta^{(k)}_{[0\cdots x_t\cdots x_O]}
\label{eqn-seq-linear}
\eeq

In Figure~\ref{fig-seq}, we display the linear functions for each possible
combination of $(x_1,x_2,x_3)$ on the left column, by linking together all
$\beta$'s in the summation~(\ref{eqn-seq-linear})  with lines, from
$\beta_{[x_1x_2x_3]}$ to $\beta_{[000]}$.

The prior for each $\beta_{\P}^{(k)}$ is a Gaussian or Cauchy distribution
centered at $0$, whose width depends on the order, $o(\P)$, of $\P$, which is
the number of nonzero elements of $\P$. There are $O+1$ such width parameters,
denoted by $\sigma_0,\ldots,\sigma_O$.  The $\sigma_o$'s are treated as
hyperparameters, assigned Inverse Gamma prior distributions with some shape
and rate parameters, leaving their values to be determined by the data. In
summary, the hierarchy of the priors is:

\beq \begin{array}{rcl} \sigma_o & \sim &
\mbox{Inverse-Gamma}(\alpha_o\,,(\alpha_o+1)\,w_o), \mbox{ for } o=0,\ldots,O\\
\beta^{(k)}_{\P}\given \sigma_{o(\P)} &\sim& \cc(0,\sigma_{o(\P)}) \mbox{ or }
N(0,\sigma_{o(\P)}^2), \mbox{ for } \P \in \mb {\mathcal{S}} \end{array}
\label{eqn-cc-prior} \eeq

\noindent where Inverse-Gamma$(\alpha,\lambda)$ denotes an Inverse Gamma
distribution with density function
$x^{-\alpha-1}\,\lambda^\alpha\,\exp(-\lambda/x)/\Gamma(\alpha)$. We express
$\alpha$ and $\lambda$  in~(\ref{eqn-cc-prior}) so that the mode of the prior is
$w_o$.

\subsection{Remarks on the Sequence Prediction Models} \label{sec-remark-blsm}

The Inverse Gamma distributions have heavy upward tails when $\alpha$ is small, and particularly when $\alpha \leq 1$, they have infinite means. An Inverse Gamma distribution with $\alpha_o\leq 1$ and small $w_o$, favors small values around $w_o$, but still allows $\sigma_o$  to be exceptionally large, as needed by the data. Similarly, the Cauchy distributions have heavy two-sided tails. The absolute value of a Cauchy random variable has infinite mean. When a Cauchy distribution with center $0$ and a small width is used as the prior for a group of parameters, such as all $\beta$'s of the interaction patterns with the same order in~(\ref{eqn-cc-prior}), a few parameters may be much larger in absolute value than others in this group. As the priors for the coefficients of high-order interaction patterns, the Cauchy distributions can therefore express more accurately than the Gaussian distributions the prior belief that most high-order interaction patterns are useless in predicting the response, but a small number may be important.

It seems redundant to use a $\mb \beta^{(k)}$ for each $k=1,\ldots,K$
in~(\ref{eqn-softmax}) since only the differences between $\mb\beta^{(k)}$
matter in~(\ref{eqn-softmax}). A non-Bayesian model could fix one of them, say
$\mb \beta^{(1)}$, all equal to $0$, so as to make the parameters identifiable.
However, when $K\not=2$, forcing $\mb \beta^{(1)}=0$ in a Bayesian model will
result in a prior that is not symmetric for all $k$, which we may not be able to
justify. When $K=2$, we do require that $\mb \beta^{(1)}$ are all equal to $0$,
as there is no asymmetry problem.

Inclusion of $\beta_\P$ other than the highest order is also a redundancy, which
facilitates the expression of appropriate prior beliefs. The prior distributions
of linear functions of similar historic sequences $x_{1:O}$ are positively
correlated since they share some common $\beta$'s, for example, in the model
displayed by Figure~\ref{fig-seq}, $l\,((1,1,1),\mb \beta)$ and $l\,((2,1,1),\mb
\beta)$ share $\beta_{[011]},\beta_{[001]}$ and $\beta_{[000]}$. Consequently,
the predictive distributions of $x_O$ are similar given similar $x_{1:O}$.  By
incorporating such a prior belief into our inference, we borrow ``statistical
strength'' for those historic sequences with few replications in the training
cases from other similar sequences with more replications, avoiding making an
unreasonably extreme conclusion due to a small number of replications.

\subsection{Bayesian Logistic Classification Models} \label{sec-def-blcm}

\begin{figure}[p]

\begin{center}
\includegraphics{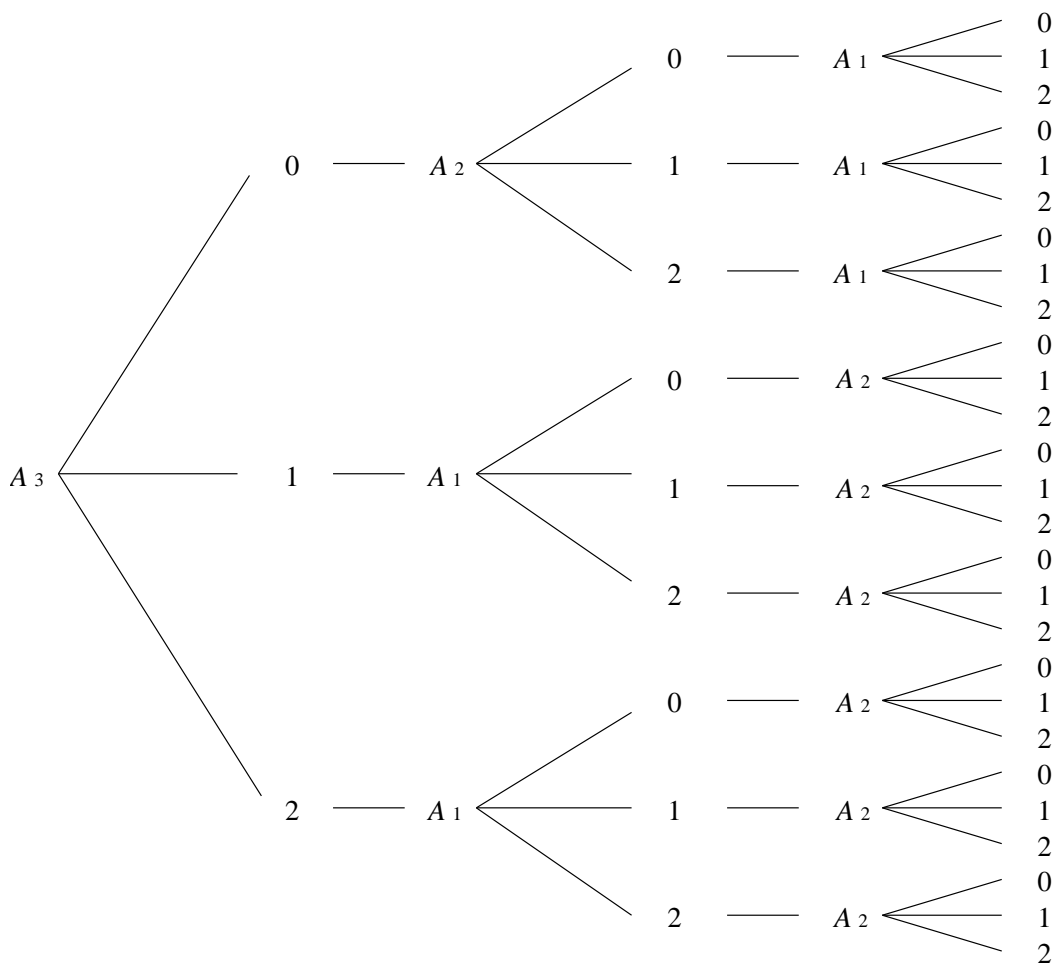}
\end{center}

\caption[A picture displaying all the interaction patterns of classification
models]{ A picture displaying all the interaction patterns, from order $0$ to
order $3$, of $3$ features $x_1,x_2,x_3$, where $x_t$ is either $1$ or $2$. A
pattern, written as $[A_1A_2A_3]$, is shown by $3$ numbers linked by lines,
where $A_t$ can be an integer from $0$ to $2$, with $A_t=0$ meaning $x_t$ could
be either $1$ or $2$. The order of $A_1,A_2,A_3$ on the above graph can be
changed to any permutation of $A_1,A_2,A_3$.} 

\label{fig-ptn-cls}

\end{figure}

In this section we define the general Bayesian classification models with high-order interactions. We write the feature vector of dimension $p$ as $(x_1,\ldots,x_p)$, or collectively $\mb x_{1:p}$, where $x_t$ takes values from $1$ to $K_t$, for $t=1,\ldots,p$. In this thesis, we consider only classification problems in which the response $y$ is discrete, assumed to take values from $1$ to $K$. But our compression method could be applied to regression problems without any difficulty. The features and response for subject $i$ are written as $\mb x_{1:p}^{(i)}$ and $y^{(i)}$. We are interested in modeling the conditional distribution $P(y\given \mb x_{1:p})$.

A pattern $\P$ is written as $[A_1A_2\ldots A_p]$, where $A_t$ can be from $0$ to $K_t$, with $A_t=0$ meaning that $x_t$ can be any value from $1$ to $K_t$. For example, $[0\ldots 01]$ denotes the pattern that fixes $x_p=1$ and allows $x_1,\ldots,x_{p-1}$ to be any values in their ranges.  When all nonzero elements of $\P$ are equal to the corresponding elements of a feature vector, $\mb x_{1:p}$, we say that pattern $\P$ occurs in $\mb x_{1:p}$, or pattern $\P$ is expressed by $\mb x_{1:p}$, denoted by $\mb x_{1:p}\in \P$. The number of nonzero elements of a pattern $\P$ is called the order of $\P$, denoted by $o(\P)$. All the patterns of order $o$ are denoted by $\mb \P^o$, and all the patterns from  order $0$ to order $O$ are denoted by $\mb \P^{0:O}=\bigcup_{o=0}^O\P^o$. All the patterns that are of order $o$ and expressed by a feature vector $\mb x_{1:p}$, are denoted by $\mb \P^o_{\mb x_{1:p}}=\{\,[A_1,\ldots,A_p] \given A_t=0 \mbox{ or } x_t\mbox{ and } \sum_{t=1}^pI(A_t\not=0)=o\}$. There are totally $\mychoose{p}{o}$ patterns in $\P^o_{\mb x_{1:p}}$. For example, $\mb \P^2_{(1,2,1)} =\{[0,2,1],\,[1,0,1],\,[1,2,0]\}$.  Figure~\ref{fig-ptn-cls} displays $\mb \P^{0:3}$ for $3$ binary ($1\,/\,2$) features $x_1,x_2,x_3$. The patterns expressed by a feature vector $(x_1,x_2,x_3)$ can be found from such a graph, by searching from the root along the lines pointing to $A_t=0$ and $A_t=x_t$. 

We will use the indicator $I(x_{1:p}\in \P)$ as a predictor variable, with
coefficient denoted by $\beta_{\P}$. For example, $\beta_{[0\cdots 0]}$ is the
intercept term. A logistic model assigns each possible value of the response a
linear function of the predictor variables. We use $\beta^{(k)}_{\P}$ to denote
the coefficient associated with pattern $\P$ and used in the linear function
for  $y = k$. All the coefficients for $y=k$ are written as
$\mb\beta^{(k)}=\{\beta^{(k)}_\P \given \P \in \mb\P^{0:O}\}$. 

A Bayesian logistic classification model using all interaction patterns from
order $0$ to order $O$ is defined as follows:

\beq
P(y=k\given \x_{1:p},\mb\beta^{(1)},\ldots,\mb\beta^{(K)}) = 
\frac{\exp( l\,(\x_{1:p},\mb\beta^{(k)}) ) }
     {\sum_{j=1}^K \exp( l\,(\x_{1:p},\mb\beta^{(j)}) ) }
\label{eqn-softmax-class}
\eeq

\noindent where

\beq l\,(\mb x_{1:p},\mb\beta^{(k)})=\sum_{\P\in
\mb\P^{0:O}}\beta_\P^{(k)}\,I(\x_{1:p}\in \P)
=\sum_{o=0}^O\sum_{ \P\in\mb\P^o_{\x_{1:p}} }\beta_\P^{(k)}\eeq

\noindent The priors for $\beta_\P^{(k)}$ are given in the same way as
in~(\ref{eqn-cc-prior}).

The remarks regarding the Bayesian sequence prediction models in
Section~\ref{sec-remark-blsm} still apply to the above classification models.
Compared with the classification models, the Bayesian sequence prediction models
are more restrictive models, using only the interaction patterns with all zeros
at the start as predictor variables.

\section{Our Method for Compressing Parameters}\label{sec-all-comp}

In this section we describe in general terms our method for compressing parameters in Bayesian models, and how the original parameters can later be retrieved as needed for use in making predictions for test cases.

\subsection{Compressing Parameters} \label{sec-comp}

In the above high-order models, the regression parameters of the likelihood function can be divided into a number of groups such that the likelihood function depends only on the sums over these groups, as shown by equation \eqref{eqn-like-train} below. We first use  the Bayesian logistic sequence models to illustrate this fact. The likelihood function of $\mb\beta^{(k)}$, for $k=1,\ldots,K$, is the product of probabilities in  \eqref{eqn-softmax} applied to the training cases, $\x^{(i)}_{1:O}, x^{(i)}_{O+1}$, for $i=1,\ldots,N$ (collectively denoted by $\trainingdata$). It can be written as follows:

\beq
L^\beta(\mb\beta^{(1)},\ldots,\mb\beta^{(K)} \given \trainingdata) =
\prod_{i=1}^N
     \frac{\exp( l\,(\x^{(i)}_{1:O},\mb\beta^{(x^{(i)}_{O+1})}) ) }
     {\sum_{j=1}^K \exp( l\,(\x^{(i)}_{1:O},\mb\beta^{(j)}) ) }
\label{eqn-seq-like}
\eeq

\noindent (When $K=2$, $\mb\beta^{(1)}$ is fixed at $0$, and therefore not included in the above likelihood function. But for simplicity, we do not write another expression for $K=2$.)

As can be seen in (\ref{eqn-seq-linear}), the function $l\,(\x_{1:O},\mb\beta)$ is the sum of the $\beta$'s associated with the interaction patterns expressed by $\x_{1:O}$. If a group of interaction patterns are expressed by the same training cases, the associated $\beta$'s will appear \textit{simultaneously} in the same factors of~(\ref{eqn-seq-like}). The likelihood function~(\ref{eqn-seq-like}) therefore depends only on the sum of these $\beta$'s, rather than the individual ones. Suppose the number of such groups is $G$. The parameters
in group $g$ are rewritten as $\beta_{g1},\ldots,\beta_{g,n_g}$, and the sum of
them is denoted by $s_g$:

\beq s_g=\sum_{k=1}^{n_g}\beta_{gk},\ \ \ \ \ \mbox{for }g=1,\ldots,G
\label{eqn-sg}
\eeq

\noindent The likelihood function can then be rewritten as:

\beq  \lefteqn{L^\beta(\beta_{11},\ldots,\beta_{1,n_1},\lldots,
               \beta_{G1},\ldots,\beta_{G,n_G})}\nonumber\\
&=&
L\left(\sum_{k=1}^{n_1}\beta_{1k},\,\ldots,\, \sum_{k=1}^{n_G}\beta_{Gk}
\right)
= L(s_1,\lldots,s_G) \label{eqn-like-train}
\eeq

\noindent  (The above $\beta$'s are only the regression coefficients for the interaction patterns occurring in training cases. The predictive distribution for a test case may use extra regression coefficients, whose distributions depend only on the priors given relevant hyperparameters.)

We need to define priors for the $\beta_{gk}$ in a way that lets us  easily find the priors of the $s_g$. For this purpose, we could assign each $\beta_{gk}$ a symmetric stable distribution centered at $0$ with  width parameter $\sigma_{gk}$. Symmetric stable distributions (Feller 1966) have the following additive property: If random variables $X_1,\ldots,X_n$ are independent and have symmetric stable distributions of index $\alpha$,  with location parameters $0$ and width parameters $\sigma_1,\ldots,\sigma_n$, then the sum of these random variables, $\sum_{i=1}^n X_i$, also has a symmetric stable distribution of index $\alpha$, with location parameter $0$ and width parameter $(\sum_{i=1}^n \sigma_i^\alpha)^{1/\alpha}$. Symmetric stable distributions exist and are unique for $\alpha\in (0,2]$. The symmetric stable distributions with $\alpha=1$  are Cauchy distributions. The density function of a Cauchy distribution with location parameter $0$ and width parameter $\sigma$ is $[\pi\sigma(1+x^2/\sigma^2)]^{-1}$. The symmetric stable distributions with $\alpha=2$ are Gaussian distributions, for which the width parameter is the standard deviation. Since the symmetric stable distributions with $\alpha$ other than $1$ or $2$ do not have closed form density functions,  we will use only Gaussian or Cauchy priors. That is, each parameter $\beta_{gk}$ has a Gaussian or Cauchy distribution with location parameter $0$ and width parameter $\sigma_{gk}$:

\beq \beta_{gk} \sim N(0,\sigma_{gk}^2)\ \ \ \mbox{or} \ \ \
\beta_{gk} \sim \cc(0,\sigma_{gk})  \label{eqn-comp-prior}
\eeq

\noindent As can been seen from the definitions of the priors (equation \eqref{eqn-cc-prior}),  some $\sigma_{gk}$ may be common for different $\beta_{gk}$, but for simplicity  we denote them individually. We might also treat the $\sigma_{gk}$'s as unknown hyperparameters, but again we assume them fixed for the moment.

If the prior distributions for the $\beta_{gk}$'s are as
in~(\ref{eqn-comp-prior}), the prior distribution of $s_g$ can be found using
the property of symmetric stable distributions:

\beq
s_g \sim N\left(0,\ \sum_{k=1}^{n_g}\sigma_{gk}^2\right)\ \ \ \mbox{or}\ \ \
s_g \sim \cc\left(0,\ \sum_{k=1}^{n_g}\sigma_{gk}\right)
\label{eqn-dist-sg}
\eeq

Let us denote the density of $s_g$ in~(\ref{eqn-dist-sg}) by $P_g^s$ (either a
Gaussian or Cauchy), and denote $s_1,\ldots,s_G$ collectively by $\mb s$. The
posterior distribution can be written as follows:

\beq P(\mb s \given \trainingdata)= {1\over
c(\trainingdata)}\,L(s_1,\lldots,s_G)\ P_1\sp{s}(s_1)\ \cdots \ P_g\sp{s}(s_G)
\label{eqn-post-sum} \eeq

\noindent where $\trainingdata$ is the training data, and $c(\trainingdata)$ is
the marginal probability or density function of $\trainingdata$.

Since the likelihood function $L(s_1,\lldots,s_G)$ typically depends on
$s_1,\ldots,s_G$ in a complicated way, we may have to use some Markov chain
sampling method to sample for $\mb s$ from distribution~(\ref{eqn-post-sum}). 

\subsection{Splitting Compressed Parameters}\label{sec-split}

\begin{figure}[t]

\begin{center} \includegraphics[width=5.0in,height=1in]{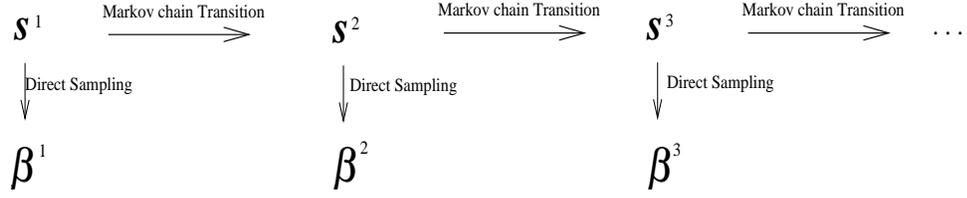}
\end{center}

\vspace*{-0.2in}\caption{A picture depicting the sampling procedure after 
compressing parameters.}

\label{fig-sample-comp}

\end{figure}

After we have obtained samples of $s_g$,  we may need to split them into their original components $\beta_{g1},\ldots,\beta_{g,n_g}$ to make predictions for test cases. This ``splitting'' distribution depends only on the prior distributions, and is independent of the training data $\trainingdata$. In other words, the splitting distribution is just the conditional distribution of $\beta_{g1},\ldots,\beta_{gn_g}$ given $\sum_{k=1}^{n_g}\beta_{gk}=s_g$, whose density function is:

\beq
P(\beta_{g1},\ldots,\beta_{g,n_g-1}\given s_g) =
\left[\prod_{k=1}^{n_g-1}\ P_{gk}(\beta_{gk})\right]\
P_{g,n_g}\left(s_g - \sum_{k=1}^{n_g-1}\beta_{gk}\right)\,/\,P_g^s(s_g)
\label{eqn-split1}
\eeq

\noindent where $P_{gk}$ is the density function of the prior for $\beta_{gk}$.
Note that $\beta_{g,n_g}$ is omitted since it is equal to $s_g -
\sum_{k=1}^{n_g-1}\beta_{gk}$. 

As will be discussed in the Section~\ref{sec-split-pred}, sampling
from~(\ref{eqn-split1}) can be done efficiently by a direct sampling method,
which does not involve costly evaluations of the likelihood function. We need to
use Markov chain sampling methods and evaluate the likelihood function only when
sampling for $\mb s$. Figure~\ref{fig-sample-comp} shows the sampling procedure
after compressing parameters, where $\mb \beta$ is a collective representation
of $\beta_{gk}$, for $g=1,\ldots,G, k = 1,\ldots,n_g-1$. When we consider
high-order interactions, the number of groups, $G$, will be much smaller than
the number of $\beta_{gk}$'s. This procedure is therefore much more efficient
than  applying Markov chain sampling methods to all the original $\beta_{gk}$
parameters.

Furthermore, when making predictions for a particular test case, we actually do
not need to sample from the distribution~(\ref{eqn-split1}), of dimension
$n_g-1$, but only from a derived 1-dimensional distribution, which saves a huge
amount of space. 

Before discussing how to sample from~(\ref{eqn-split1}), we first phrase this
compressing-splitting procedure more formally in the next section to show its
correctness. 

\subsection{Correctness of the Compressing-Splitting Procedure}
\label{sec-correctness}

The above procedure of compressing and splitting parameters can be seen in terms
of a transformation of the original parameters $\beta_{gk}$  to a new set of
parameters containing $s_g$'s, as defined in~(\ref{eqn-sg}), in light of the
training data. The posterior distribution~(\ref{eqn-post-sum}) of $\mb s$ and
the splitting distribution~(\ref{eqn-split1}) can be derived from the joint
posterior distribution of the new parameters. 

The invertible mappings from the original parameters to the new parameters are
shown as follows, for $g=1,\ldots,G$,

\beq 
(\beta_{g1},\ldots,\beta_{g,n_g-1},\beta_{g,n_g})\ \ \Longrightarrow \ \ 
(\beta_{g1},\ldots,\beta_{g,n_g-1},\sum_{k=1}^{n_g}\beta_{gk}) =
 (\beta_{g1},\ldots,\beta_{g,n_g-1},s_g)
\label{eqn-transform}
\eeq

\noindent In words, the first $n_g-1$ original parameters $\beta_{gk}$'s are
mapped to themselves (we might use another set of symbols, for example $b_{gk}$,
to denote the new parameters, but here we still use the old ones for simplicity
of presentation while making no confusion), and the sum of all $\beta_{g,k}$'s,
is mapped to $s_g$.  Let us denote the new parameters $\beta_{gk}$, for
$g=1,\ldots,G, k = 1,\ldots,n_g-1$, collectively by $\mb \beta$, and denote
$s_1,\ldots,s_g$ by $\mb s$. (Note that $\mb \beta$ does not include
$\beta_{g,n_g}$, for $g=1,\ldots,G$. Once we have obtained the samples of $\mb
s$ and $\mb \beta$ we can use $\beta_{g,n_g}=s_g - \sum_{k=1}^{n_g-1}\beta_{gk}$
to obtain the samples of $\beta_{g,n_g}$.)

The posterior distribution of the original parameters, $\beta_{gk}$, is:

\beq
P(\beta_{11},\lldots,\beta_{G,n_G}\given \trainingdata)
=
{1\over c(\trainingdata)}
L\left(\sum_{k=1}^{n_1}\beta_{1k},\,\ldots,\,
  \sum_{k=1}^{n_G}\beta_{Gk}\right)
\prod_{g=1}^G\prod_{k=1}^{n_g}\ P_{gk}(\beta_{gk})
\label{eqn-original-post}
\eeq

\noindent By applying the standard formula for the density function of
transformed random variables, we can obtain from~(\ref{eqn-original-post}) the
posterior distribution of the $\mb s$ and $\mb \beta$:

\beq
P(\mb s,\mb \beta \given \trainingdata ) =
{1\over c(\trainingdata)}\,
L\left(s_1,\,\ldots,\,s_G\right)
\prod_{g=1}^G\left[
\prod_{k=1}^{n_g-1}P_{gk}(\beta_{gk})\right]
P_{g,n_g}\left(s_g - \sum_{k=1}^{n_g-1}\beta_{gk}\right)
\,|\det(J)|
\label{eqn-transform-post}
\eeq

\noindent where the $|\det(J)|$ is absolute value of the determinant of the
Jacobian matrix, $J$, of the mapping~(\ref{eqn-transform}), which can be shown
to be $1$.

Using the additive property of symmetric stable distributions, which is stated
in section~\ref{sec-comp}, we can analytically integrate out $\mb \beta$ in
$P(\mb s,\mb \beta \given \trainingdata)$, resulting in the marginal
distribution $P(\mb s \given \trainingdata)$:

\beq
P(\mb s \given \trainingdata)
&=& \int\,P(\mb s,\mb \beta \given \trainingdata )\,d\mb\beta \\
&=& {1\over c(\trainingdata)}\,
L\left(s_1,\,\ldots,\,s_G\right)\nonumber \cdot\\
&&\ \ \ \prod_{g=1}^G\int\cdots\int\,\left[
\prod_{k=1}^{n_g-1}P_{gk}(\beta_{gk})\right]
P_{g,n_g}\left(s_g - \sum_{k=1}^{n_g-1}\beta_{gk}
\right)d\beta_{g1}\cdots d\beta_{g,n_g-1}\\
&=& {1\over c(\trainingdata)}\,L\left(s_1,\,\ldots,\,s_G\right)\,
    P^s_1(s_1)\ \cdots\ P^s_G(s_G)
\label{eqn-dist-s}
\eeq

The conditional distribution of $\mb\beta$ given $\trainingdata$ and $\mb s$ can
then be obtained as follows:

\beq
P(\mb \beta \given \mb s,\trainingdata)
& = & P(\mb s,\mb \beta \given \trainingdata)\,/\,P(\mb
s \given \trainingdata) \\
& = & \prod_{g=1}^G\,\left[
\prod_{k=1}^{n_g-1}P_{gk}(\beta_{gk})\right]
P_{g,n_g}\left(s_g - \sum_{k=1}^{n_g-1}\beta_{gk}\right)\,/\,P^s_g(s_g)
\label{eqn-cond-beta-s}
\eeq

\noindent From the above expression, it is clear that $P(\mb \beta \given \mb s,\trainingdata)$ is unrelated to $\trainingdata$, i.e., $P(\mb \beta\given \mb s,\trainingdata)=P(\mb \beta \given \mb s)$, and is independent for different groups. Equation~(\ref{eqn-split1}) gives this distribution only for one group $g$.

\subsection{Sampling from the Splitting Distribution} \label{sec-split-pred}

In this section, we discuss how to sample from the splitting distribution~(\ref{eqn-split1}) to make predictions for test cases after we have obtained samples of $s_1,\ldots,s_G$. 

If we sampled for all the $\beta_{gk}$'s, storing them would require a huge amount of space when the number of parameters in each group is huge. We therefore sample for $\mb \beta$ conditional on $s_1,\ldots,s_G$ only temporarily, for a particular test case. As can be seen in Section~\ref{sec-def-blsm} and~\ref{sec-def-blcm}, the predictive function needed to make prediction for a particular test case, for example the probability that a test case is in a certain class, depends only on the sums of subsets of $\beta_{gk}$'s in  groups. After re-indexing the $\beta_{gk}$'s in each group such that the $\beta_{g1},\ldots,\beta_{g,t_g}$ are those needed by the test case, the variables needed for making a prediction for the test case are:

\beq  
s^t_g &=& \sum_{k=1}^{t_g}\beta_{gk}\, , \mbox{ for } g=1,\ldots,G,
\eeq

\noindent When $t_g=0$, $s^t_g=0$, and when $t_g=n_g$, $s^t_g = s_g$. The predictive function may also use some sums of  extra regression coefficients associated with the interaction patterns that occur in this test case but not in training cases. Suppose  the extra regression coefficients need to be divided into $Z$ groups, as required by the form of the predictive function, which we denote by  $\beta_{11}^*,\ldots,\beta_{1,n^*_1}^*,\ldots,\beta_{Z,1}^*, \ldots,\beta_{Z,n^*_Z}^*$. The variables needed for making prediction for the test cases are:

\beq
s^*_z
&=& \sum_{k=1}^{n^*_z}\beta_{zk}^*\, , \mbox{ for } z =1,\ldots,Z \eeq

In terms of the above variables, the function needed to make a prediction for a
test case can be written as

\beq a\left(\sum_{k=1}^{t_1}\beta_{1k},\ \ldots \ ,\sum_{k=1}^{t_G}\beta_{Gk},\
\sum_{k=1}^{n^*_1}\beta_{1k}^*,\lldots,\sum_{k=1}^{n^*_Z}\beta_{Zk}^* \right)
\label{eqn-predfun} = a(s_1^t,\lldots,s_G^t,s_1^*,\lldots,s_Z^*) 
\label{eqn-blsm-pred} \eeq

\noindent 

Let us write $s^t_1,\ldots,s^t_G$ collectively as $\mb s^t$, and write
$s^*_1,\ldots,s^*_Z$ as $\mb s^*$. The integral required to make a prediction
for this test case is

\beq \int\ a(\mb s^t,\mb s^*)\ P(\mb s^*)\ P(\mb s \given \trainingdata)\
\prod_{g=1}^G\ P(s^t_g\given s_g)\ d\mb s\ d\mb s^t d\mb s^*.
\label{eqn-int-pred} \eeq

The integral over $\mb s^t$ is done by MCMC. We also need to sample for $\mb
s^*$ from $P(\mb s^*)$, which is the prior distribution of $\mb s^*$ given some
hyperparameters (from the current MCMC iteration) and can therefore be sampled
easily. Finally, we need to sample from $P(s^t_g\given s_g)$, which can be
derived from~(\ref{eqn-split1}), shown as follows: 

\beq P(s^t_g \given s_g) = P^{(1)}_g(s_g^t)\
P^{(2)}_g(s_g-s_g^t)\,/\,P^s_g(s_g) \label{eqn-split2} \eeq

\noindent where $P^{(1)}_g$ and  $P^{(2)}_g$ are the priors (either Gaussian or
Cauchy) of $\sum_{1}^{t_g}\beta_{gk}$ and $\sum_{t_g+1}^{n_g}\beta_{gk}$,
respectively. We can obtain~(\ref{eqn-split2}) from~(\ref{eqn-split1})
analogously as we obtained the density of $s_g$, that is, by first mapping $\mb
\beta$ and $\mb s$ to a set of new parameters containing $\mb s$ and $\mb s^t$,
then integrating away other parameters, using the additive property of
symmetric stable distributions. The distribution~(\ref{eqn-split2}) splits 
$s_g$ into two components.

When the priors for the $\beta_{gk}$'s are Gaussian distributions, the
distribution~(\ref{eqn-split2}) is also a Gaussian distribution, given as
follows:

\beq s^t_g \given s_g \ \sim \  N\left(s_g\
\frac{\sigmaa^2}{\sigmaa^2 +
\sigmab^2}\ ,\  \sigmaa^2\left(1\ -\
\frac{\sigmaa^2}{\sigmaa^2 +
\sigmab^2}\right) \right)
\label{eqn-cond-gs}
\eeq

\noindent where $\sigmaa^2 = \sum_{k=1}^{t_g}\sigma_{gk}^2$
and $\sigmab^2 = \sum_{t_g+1}^{n_g}\sigma_{gk}^2$.
Since~(\ref{eqn-cond-gs}) is a Gaussian distribution,  we can  sample from it
by standard methods. 

When we use Cauchy distributions as the priors for the $\beta_{gk}$'s, the
density function of~(\ref{eqn-split2}) is:

\beq
P(s_g^t \given s_g) = {1\over C}\,
 \frac{1}{ \sigmaa^2 + (s_g^t)^2}\
 \frac{1}{ \sigmab^2 + (s_g^t-s_g)^2}
\label{eqn-cond-cauchy}
\eeq

\noindent where $\sigmaa = \sum_{k=1}^{t_g}\sigma_{gk}$, $\sigmab  =
\sum_{t_g+1}^{n_g}\sigma_{gk}$, and $C$ is the normalizing constant given below
by~(\ref{eqn-C-cauchy}).

When $s_g=0$ and $\sigmaa=\sigmab$, the distribution~(\ref{eqn-cond-cauchy}) is
a t-distribution with $3$ degrees of freedom, mean $0$ and width
$\sigmaa/\sqrt{3}$, from which we can sample by standard methods. Otherwise, the
cumulative distribution function (CDF) of~(\ref{eqn-cond-cauchy}) can be shown
to be:

\beq
F(\st\,;\,\s,\sigmaa,\sigmab)
& = &\frac{1}{C}\,
     \left[r
           \log\left(\frac{(\st)^2+\sigmaa^2}
                          {(\st-\s)^2+\sigmab^2}
               \right) + \right.
     \nonumber\\
&   &\ \ \ \ \ \ \
      p_0\,\left(\arctan\left(\frac{\st}{\sigmaa}\right)+\frac{\pi}{2}\right)+
     \nonumber\\
&   &\ \ \ \ \ \ \
     \left.
           p_s\,\left(\arctan\left(\frac{\st-\s}{\sigmab}\right)+\frac{\pi}{2}
                \right)
     \right]
\label{eqn-cdf-split-cc1}
\eeq

\noindent where

\beq
C  &=&{\pi\,(\sigmaa+\sigmab)
       \over \sigmaa\sigmab\,(\s^2+(\sigmaa+\sigmab)^2)}\,,
   \label{eqn-C-cauchy}\\
r  &=&\frac{\s}{\s^4+2\left(\sigmaa^2+\sigmab^2\right)\,\s^2+
      \left(\sigmaa^2-\sigmab^2\right)^2}\,, \\
p_0&=&{1\over \sigmaa}\,\frac{\s^2 -
      \left(\sigmaa^2-\sigmab^2\right)}
      {\s^4+2\left(\sigmaa^2+\sigmab^2\right)\,\s^2+
      \left(\sigmaa^2-\sigmab^2\right)^2},\\
p_s&=&{1\over \sigmab}\,\frac{\s^2 +
      \left(\sigmaa^2-\sigmab^2\right)}
      {\s^4+2\left(\sigmaa^2+\sigmab^2\right)\,\s^2+
      \left(\sigmaa^2-\sigmab^2\right)^2}
\eeq

When $s_g\not=0$, the derivation of~(\ref{eqn-cdf-split-cc1})  uses the
equations below from~(\ref{eqn-decomp1-b}) to~(\ref{eqn-decomp1-e}) as follows,
where $p=(a^2-c)/b,q=b+q,r=pc-a^2q$, and we assume $4c-b^2>0$, 

\beq
\frac{1}{x^2+a^2}\,\frac{1}{x^2+bx+c}
&\hspace*{-0.1in}=&\hspace*{-0.1in}
{1\over r}\,\left(\frac{x+p}{x^2+a^2} - \frac{x+q}{x^2+bx+c}
        \right) \label{eqn-decomp1-b}\\
\int_{-\infty}^x\frac{u+p}{u^2+a^2}du
&\hspace*{-0.1in}=&\hspace*{-0.1in}
{1\over2}\,\log(x^2 + a^2) + {p\over a}\,\arctan\left({x\over a}\right)
     +{\pi\over 2}
\\
\int_{-\infty}^x{u+q \over u^2+bu+c}du
&\hspace*{-0.1in}=&\hspace*{-0.1in}
{1\over2}\,\log(x^2 + bx + c) +
{2q-b \over \sqrt{4c-b^2}}\,\arctan\left({2x+b \over \sqrt{4c-b^2}}\right)
+{\pi\over 2} \label{eqn-decomp1-e}
\eeq

When $s_g=0$, the derivation of~(\ref{eqn-cdf-split-cc1}) uses the following
equations:

\beq
\frac{1}{x^2+a^2}\,\frac{1}{x^2+b^2} &=& 
\frac{1}{b^2-a^2}\,\left(\frac{1}{x^2+a^2}-\frac{1}{x^2+b^2}\right)\\
\int_{-\infty}^x \frac{1}{u^2+a^2}\,du &=& 
\frac{1}{a}\left(\arctan\left({x\over a}\right)+\frac{\pi}{2}\right)
\eeq

Since we can compute the CDF of~(\ref{eqn-cond-cauchy}) with 
~(\ref{eqn-cdf-split-cc1}) explicitly, we can use the inversion method to sample
from~(\ref{eqn-cond-cauchy}), with the inverse CDF computed by some numerical
method. We chose the Illinois method (Thisted 1988, Page 171), which is robust
and fairly fast.

When sampling for $s^t_1,\ldots,s^t_G$ temporarily for each test case is not
desired, for example, when we need to make predictions for a huge number of test
cases at a time, we can still apply  the above method that splits a Gaussian or
Cauchy random variable into two parts $n_g-1$ times to split $s_g$  into $n_g$
parts. Our method for compressing parameters is still useful because sampling
from the splitting distributions uses direct sampling methods, which are much
more efficient than applying Markov chain sampling method to the original
parameters. However, we will not save space if we take this approach of sampling
for all $\beta$'s.

\section{Application to Sequence Prediction Models} \label{sec-blsm}

In this section, we show how to compress parameters of logistic sequence prediction models in which states of a sequence  are discrete, as defined in Section \ref{sec-def-blsm}.  To demonstrate our method, we use a binary data set generated using a hidden Markov model, and a data set created from English text, in which each state has 3 possibilities (consonant, vowel, and others). These experiments show that our compression method produces a large reduction in the number of parameters needed for training the model, when the prediction for the next state of a sequence is based on a long  preceding sequence, i.e., a high-order model. We also show that good predictions on test cases result from being able to use a high-order model.

\subsection{Grouping Parameters of Sequence Prediction
Models}\label{compress-seq}

In this section, we describe a scheme for dividing the $\beta$'s  into a number of groups, based on the training data, such that the likelihood function depends only on the sums in groups, as shown by~(\ref{eqn-like-train}). 

Since the linear functions for different values of response have the same form except the superscript, the way we divide $\mb\beta^{(k)}$ into groups is the same for all $k$. Our task is to find the groups of interaction patterns expressed by the same training cases.

\begin{figure}[p]

\begin{center}

\includegraphics[scale=0.72]{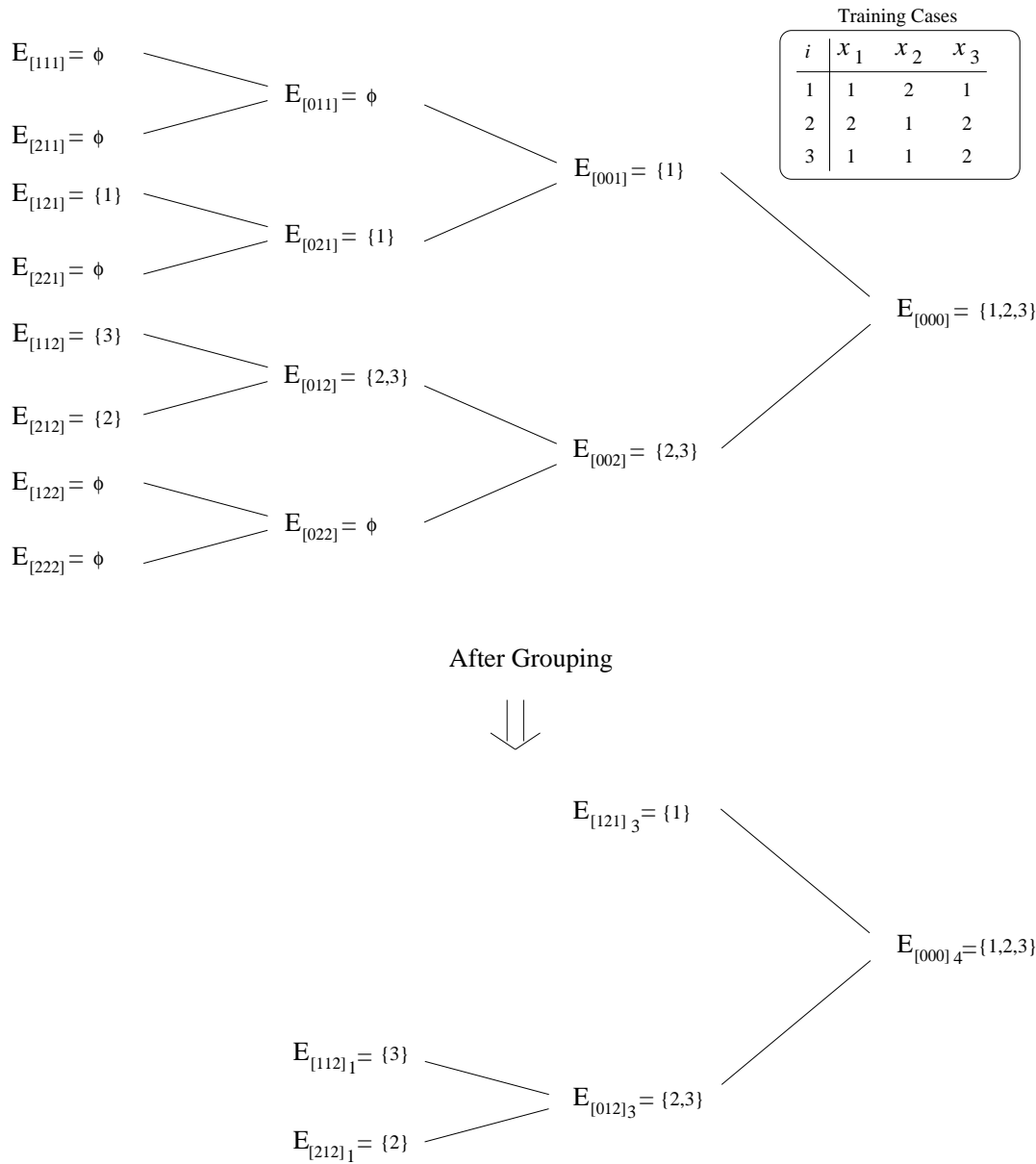}

\end{center}

\caption[A picture showing that the interaction patterns in logistic sequence prediction models can be grouped, illustrated with binary sequences of length $O=3$, based on $3$ training cases]{A picture showing that the interaction patterns in logistic sequence prediction models can be grouped, illustrated with binary sequences of length $O=3$, based on $3$ training cases shown in the upper-right box. $E_\P$ is the expression of the pattern (or superpattern) $\P$ --- the indices of the training cases in which the $\P$ is expressed, with $\phi$ meaning the empty set. We group the patterns with the same expression together, re-represented collectively by a ``superpattern'', written as $[0\cdots 0A_b\cdots A_O]_f$,  meaning $\bigcup_{t=b}^{f}\,[0\cdots 0A_t\cdots A_O]$, where $1 \leq b \leq f \leq O+1$, and in particular when $t=O+1$, $[0\cdots 0A_t\cdots A_O] = [0\cdots0]$. We also remove the patterns not expressed by any of the training cases. Only $5$ superpatterns with unique expressions are left in the lower picture.}

\label{fig-seq-group}

\end{figure}

Let us use $E_{\P}$ to denote the ``expression'' of the pattern $\P$ --- the indices of training cases in which $\P$ is expressed, a subset of $1,\ldots,N$. For example, $E_{[0\cdots 0]}=\{1,\ldots,N\}$. In other words, the indicator for pattern $\P$ has value $1$ for the training cases in $E_{\P}$, and $0$ for others. We can display $E_\P$ in a tree-shape, as we displayed $\beta_\P$. The upper part of Figure~\ref{fig-seq-group} shows such expressions for each pattern of binary sequence of length $O=3$, based on $3$ training cases: $\x_{1:3}^{(1)}=(1,2,1)$,$\x_{1:3}^{(2)}=(2,1,2)$ and $\x_{1:3}^{(3)}=(1,1,2)$. From Figure~\ref{fig-seq-group}, we can see that the expression of a ``stem'' pattern is equal to the union of the expressions of its ``leaf'' patterns, for example, $E_{[000]}=E_{[001]}\bigcup E_{[002]}$ . 

\begin{figure}[p]

\begin{center}

\includegraphics[scale=0.9]{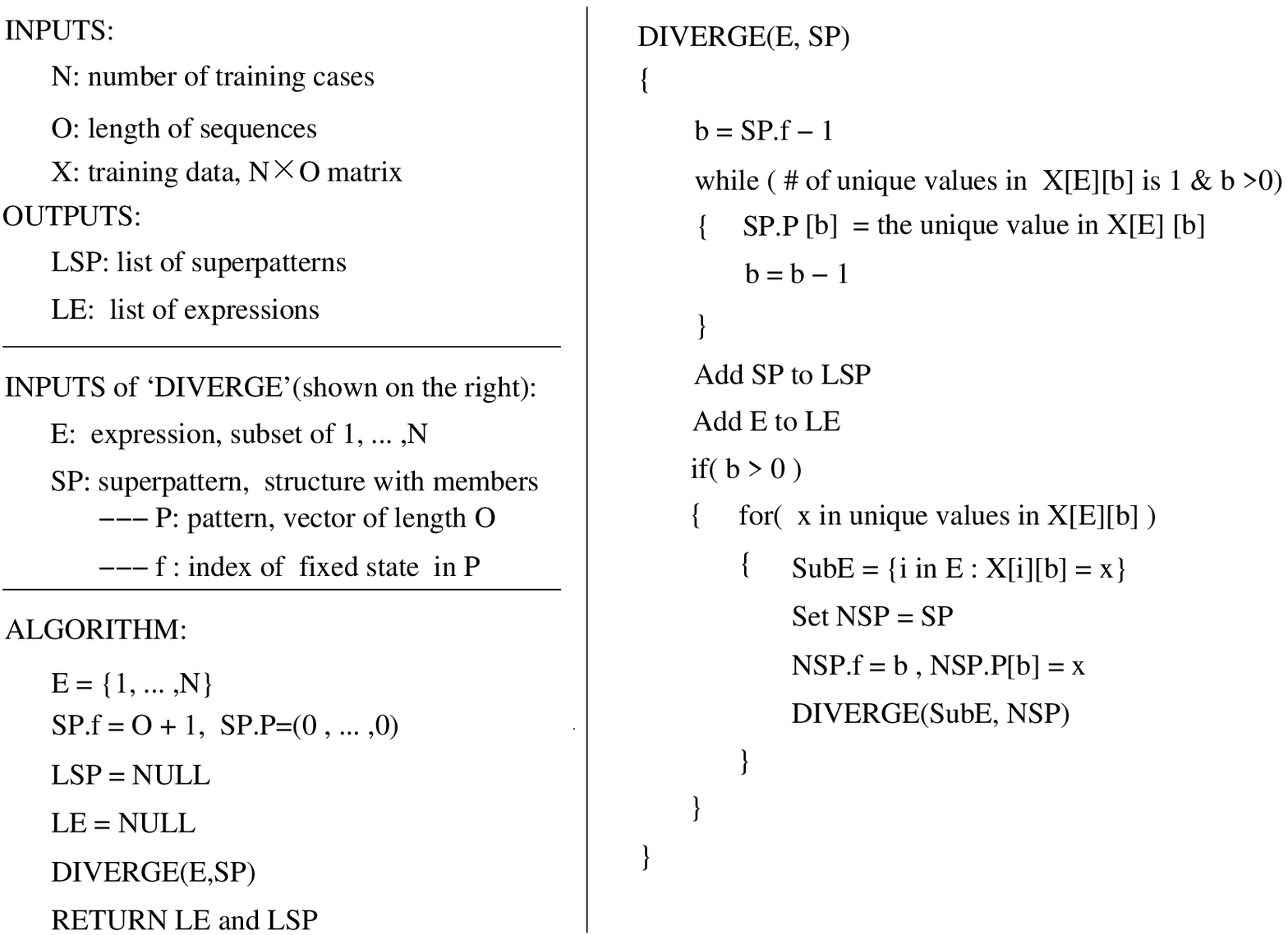}

\end{center}

\caption[The algorithm for grouping parameters of Bayesian logistic sequence
prediction models.]{The algorithm for grouping parameters of Bayesian logistic
sequence prediction models. To group parameters, we call function ``DIVERGE''
with the initial values of expression $E=\{1,\ldots,N\}$ and superpattern
$SP=[0\ldots 0]_{O+1}$, as shown in above picture, resulting in two lists of the
same length, LE and LSP, respectively storing the expressions and the
corresponding superpatterns. Note that the first index of an array is assumed to
be $1$, and that the $X[E][b]$ means a 1-dimension subarray  of $X$ in which the
row indices are in $E$ and the column index equals $b$. }

\label{fig-alg-seq-group}

\end{figure}

When a stem pattern has only one leaf pattern with non-empty expression, the stem and leaf patterns have the same expression, and can therefore be grouped together. This grouping procedure will continue by taking the leaf pattern as the new stem pattern, until encountering a stem pattern that ``splits'', i.e. has more than one leaf pattern with non-empty expression.  For example, $E_{[001]},E_{[021]}$ and $E_{[121]}$ in Figure~\ref{fig-seq-group} can be grouped together. All such patterns must be linked by lines, and can be represented collectively with a ``superpattern'' $SP$, written as $[0\cdots 0A_b\cdots A_O]_f=\bigcup_{t=f}^{b}\,[0\cdots 0A_t\cdots A_O]$, where $1\leq b \leq f \leq O+1$, and in particular when $t=O+1$, $[0\cdots 0A_t\cdots A_O] = [0\cdots0]$. One can easily translate the above discussion into a  computer algorithm. Figure~\ref{fig-alg-seq-group} describes the algorithm for grouping parameters of Bayesian logistic sequence prediction models, in a C-like language, using a recursive function.

\label{sec-comp-blsm}

An important property of our method for compressing parameters of sequence prediction models is that given $N$ sequences as training data, conceivably of infinite length, denoted by $x^{(i)}_{-\infty},\ldots,x^{(i)}_{-1}$, for $i=1,\ldots,N$, the number of superpatterns with unique expressions, and accordingly the number of compressed parameters, will converge to a finite number as $O$ increases. The justification of this claim is that if we keep splitting the expressions following the tree shown in Figure~\ref{fig-seq-group}, at a certain time, say $t$, every expression will be an expression with only 1 element (suppose we in advance remove the sequences that are identical with another one). When considering further smaller $t$, no more new superpattern with different expressions will be introduced, and the number of superpatterns will not grow. The number of the \textit{compressed parameters}, the regression coefficients for the superpatterns, will therefore not grow after the time $t$.

In contrast, after the time $t$ when each interaction pattern is expressed by only $1$ training case, if the order is increased by $1$, the number of interaction patterns is increased by the number of training cases. The regression coefficients associated with these original interaction patterns, called \textit{the original parameters} thereafter, will grow linearly with the order considered. Note that these original parameters do not include the regression coefficients for those interaction patterns not expressed by any training case. The total number of regression coefficients defined by  the model grows exponentially with the order considered.

\subsection{Making Prediction for a Test Case}\label{sec-blsm-pred}

Given $\beta^{(1)},\ldots,\mb \beta^{(K)}$, the predictive probability for the next state $\x^*_{O+1}$ of a test case for which we know the historic sequence $\x^*_{1:O}$ can be computed using equation~(\ref{eqn-softmax}), applied to $\x^*_{1:O}$. A Monte Carlo estimate of $P(x^*_{O+1} = k\given \x^*_{1:O},\trainingdata)$ can be obtained by averaging~(\ref{eqn-softmax}) over the Markov chain samples from the posterior distribution of $\mb\beta^{(1)},\ldots,\mb\beta^{(K)}$. 

Each of the $O+1$ patterns expressed by the  test case $\x^*_{1:O}$ is either expressed by some training case (and therefore belongs to one of the superpatterns), or is a new pattern (not expressed by any training case). Suppose we have found $\gamma$ superpatterns. The $O+1$ $\beta$'s in the linear function $l(\x_{1:O}^*,\beta^{(k)})$ can accordingly be divided into $\gamma+1$ groups (some groups may be empty). The function $l(\x_{1:O}^*,\beta^{(k)})$ can be written as the sum of the sums of the $\beta$'s over these $\gamma+1$ groups. Consequently, $P(x^*_{O+1} = k\given \x^*_{1:O})$ can be written in the form of~(\ref{eqn-blsm-pred}). As discussed in Section~\ref{sec-split-pred}, we need to only split the sum of the $\beta$'s associated with a superpattern, i.e., a compressed parameter $s_g$, into two parts, such that one of them is the sum of those $\beta$ expressed by the  test case $\x^*_{1:O}$, using the splitting distribution~(\ref{eqn-split2}). 

It is easy to identify the patterns that are also expressed by $\x^*_{1:O}$ from a superpattern $[0\cdots A_b\cdots A_O]_f$. If $(x^*_f,\ldots,x^*_O)\not=(A_f,\ldots,A_O)$, none of the patterns in $[0\cdots A_b\cdots A_O]_f$ are expressed by $\x^*_{1:O}$, otherwise, if $(x^*_{b'},\ldots,x^*_{O})=(A_{b'},\ldots,A_O)$ for some $b'$ ($b\leq b' \leq f$), all patterns in $[0\cdots A_{b'}\cdots A_O]_f$ are expressed by $\x^*_{1:O}$. 

\subsection{Experiments with a Hidden Markov Model} \label{sec-sim-blsm}

In this section we apply Bayesian logistic sequence prediction modeling, with or without our compression method, to data sets generated using a Hidden Markov model, to demonstrate our method for compressing parameters. The experiments show that when the considered length of the sequence $O$ is increased, the number of compressed parameters will converge to a fixed number, whereas the number of original parameters will increase linearly. Our compression method also improves the quality of Markov chain sampling in terms of autocorrelation. We therefore obtain good predictive performances in a small amount of time using long historic sequences.

\subsubsection{The Hidden Markov Model Used to Generate the Data}

\begin{figure}[ht]

\begin{center}

\includegraphics[scale=1]{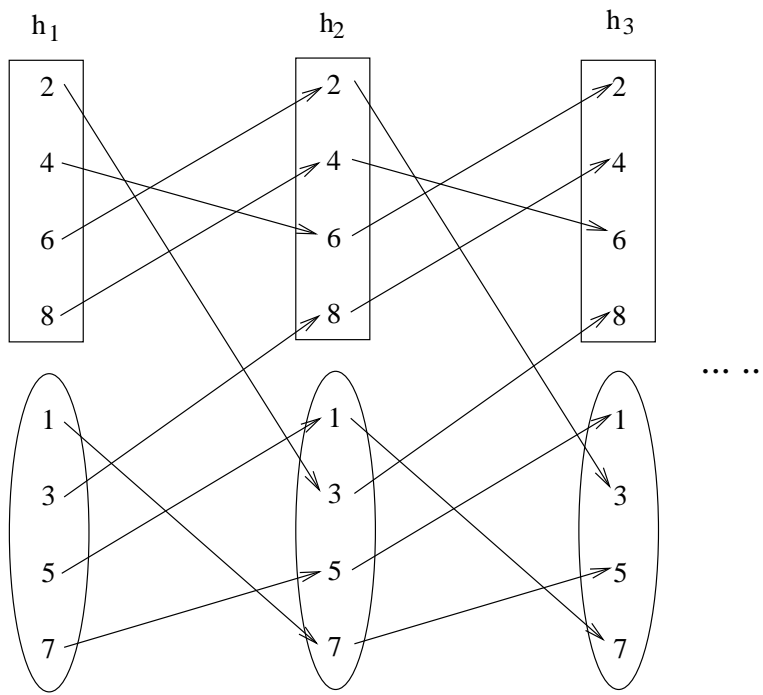}

\end{center}

\caption[A picture showing a Hidden Markov Model, which is used to generate sequences to demonstrate  Bayesian logistic sequence prediction models]{ A picture showing a Hidden Markov Model, which is used to generate sequences to demonstrate  Bayesian logistic sequence prediction models. Only the dominating transition probabilities of 0.95 are shown using arrows in the above graph, while from any state the hidden Markov chain can also move to any other state with a small probability. When $h_t$ is in a rectangle, $x_t$ is equal to $1$ with probability 0.95, and $2$ with probability 0.05, otherwise, when $h_t$ is in an oval, $x_t$ is equal to $2$ with probability 0.95, and $1$ with probability 0.05. }

\label{fig-hmm}

\end{figure}

Hidden markov models (HMM) are applied widely in many areas, for example, speech recognition (Baker 1975), image analysis (Romberg et.al. 2001), computational biology (Sun 2006). In a simple hidden Markov model, the observable sequence $\{x_t\given t=1,2,\ldots\}$ is modeled as a noisy representation  of a hidden sequence $\{h_t \given t=1,2,\ldots\}$ that has the Markov property (the distribution of $h_t$ given $h_{t-1}$ is independent with the previous states before $h_{t-1}$). Figure~\ref{fig-hmm} displays the hidden Markov model used to generate our data sets, showing the transitions of three successive states. The hidden sequence $h_t$ is an Markov chain with state space $\{1,\ldots,8\}$, whose dominating transition probabilities are shown by the arrows in Figure~\ref{fig-hmm}, each of which is 0.95. However, the hidden Markov chain can move from any state to any other state as well, with some small probabilities. If $h_t$ is an even number, $x_t$ will be equal to $1$ with probability 0.95 and $2$ with probability 0.05, otherwise, $x_t$ will be equal to $2$ with probability 0.95 and $1$ with probability 0.05. The sequence $\{x_t\given t=1,2,\ldots\}$ generated by this exhibits high-order dependency, though the hidden sequence is only a Markov chain. We can see this by looking at the transitions of observable $x_t$ in Figure~\ref{fig-hmm}. For example, if $x_1=1$ (rectangle) and $x_2=2$ (oval), it is most likely to be generated by $h_1=2$ and $h_2=3$, since this is the only strong connection from the rectangle to the oval, consequently, $h_3=8$ is most likely to to be the next, and $x_3$ is therefore most likely to be $1$ (rectangle). 

\subsection{Specifications of the Priors and Computation Methods}

\subsubsection{The Priors for the Hyperprameters}
\label{sec-priors-seq}

We fix $\sigma_0$ at $5$ for the Cauchy models and $10$ for the Gaussian models.
For $o>0$, the prior for $\sigma_o$ is Inverse
Gamma$(\alpha_o,(\alpha_o+1)w_o)$, where $\alpha_o$ and $w_o$ are:

\beq
\alpha_o=0.25,\ \ \ w_o = 0.1/o,\ \ \ \ \mbox{for }o =1,\ldots,O
\eeq

\noindent  The quantiles of Inverse-Gamma$(0.25,1.25\times 0.1)$, the prior of
$\sigma_1$, are shown as follows:

\noindent$$
\begin{array}{l|lllllllllll}
p & 0.01&0.1 & 0.2 & 0.3 &  0.4 &  0.5 &  0.6 & 0.7  & 0.8&0.9&0.99\\
\hline
q & 0.05&0.17& 0.34& 0.67&  1.33&  2.86&  7.13& 22.76& 115.65 &1851.83
  &1.85\times 10^7
\end{array}
$$

\noindent The quantiles of other $\sigma_o$ can be obtained by multiplying those
of $\sigma_1$ by $1/o$.

\subsubsection{The Markov Chain Sampling Method}

We use Gibbs sampling to sample for both the $s_g$'s (or the $\beta_{gk}$'s when not applying our compression method) and the hyperparameters, $\sigma_o$. These 1-dimensional conditional distributions are sampled using the slice sampling method (Neal 2003), summarized as follows.  In order to sample from a 1-dimensional distribution with density $f(x)$, we can draw points $(x,y)$ from the uniform distribution over the set $\{(x,y)\given 0 < y < f(x)\}$, i.e., the region of the 2-dimensional plane between the x-axis and the curve of $f(x)$. One can show that the marginal distribution of $x$ drawn this way is $f(x)$.  We can use Gibbs sampling scheme to sample from the uniform distribution over $\{(x,y)\given 0 < y < f(x)\}$. Given $x$, we can draw $y$ from the uniform distribution over $\{y\given 0< y < f(x)\}$. Given $y$, we need to draw $x$ from the uniform distribution over the ``slice'', $S=\{x\given f(x) > y\}$. However, it is generally infeasible to draw a point directly from the uniform distribution over $S$. (Neal 2003) devises several Markov chain sampling schemes that leave this uniform distribution over $S$ invariant. One can show that this updating of $x$ along with the previous updating of $y$ leaves $f(x)$ invariant. Particularly we chose the ``stepping out'' plus ``shrinkage'' procedures. The ``stepping out'' scheme first steps out from the point in the previous iteration, say $x_0$, which is in $S$, by expanding an initial interval, $I$, of size $w$ around $x_0$ on both sides with intervals of size $w$, until the ends of $I$ are outside $S$, or the number of steps has reached a pre-specified number, $m$. To guarantee correctness, the initial interval $I$ is positioned randomly around $x_0$, and $m$ is randomly aportioned for the times of stepping right and stepping left.  We then keep drawing a point uniformly from $I$ until obtaining an $x$ in $S$. To facilitate the process of obtaining an $x$ in $S$, we shrink the interval $I$ if we obtain an $x$ not in $S$ by cutting off the left part or right part of $I$  depending on whether $x<x_0$ or $x>x_0$. 

We set $w=20$ when sampling for $\beta$'s if we use Cauchy priors, considering that there might be two modes in this case, and set $w=10$ if we use Gaussian priors. We set $w=1$ when sampling for $\sigma_o$. The value of $m$ is $50$ for all cases. We trained the Bayesian logistic sequence model, with the compressed or the original parameters, by running the Markov chain 2000 iterations, each updating the $\beta$'s $1$ time, and updating the $\sigma$'s $10$ times, both using slice sampling. The first $750$ iterations were discarded, and every $5$th iteration afterward was used to predict for the test cases.  The number of $750$ was chosen empirically after looking at many trial runs of Markov chains for many different circumstances.

The above specification of Markov chain sampling and the priors for the hyperparameters will be used for all experiments in this chapter, including the experiments on classification models discussed in Section~\ref{sec-blcm}.

\subsubsection{Experiment Results}

We used the HMM in Figure~\ref{fig-hmm} to generate $5500$ sequences with length $21$. We used $5000$ sequences as test cases,  and the remaining $500$ as the training cases. We tested the prediction methods by predicting $x_{21}$ based on varying numbers of preceding states, $O$, chosen from the set $\{1,2,3,4,5,7,12,15,17,20\}$. 

\begin{figure}[p]

\begin{center}

\includegraphics[scale=0.8]{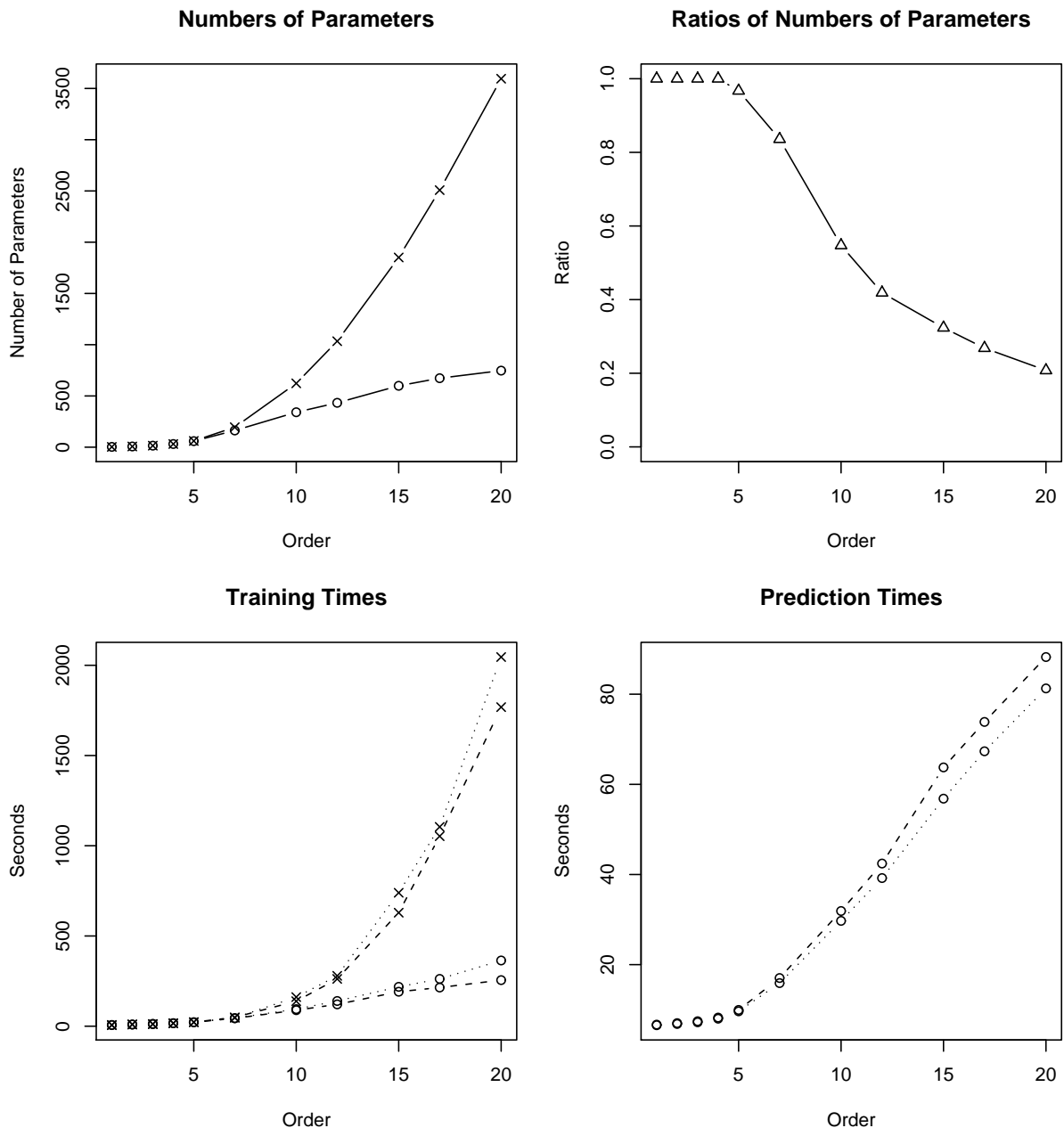}

\end{center}

\vspace*{-0.3in}\caption[Plots showing the reductions of the number of parameters and the training time with our compression method using the experiments on a data set generated by a HMM]{Plots showing the reductions of the number of parameters and the training time with our compression method using the experiments on a data set generated by a HMM. The upper-left plot shows the number of the compressed and the original parameters based on $500$ training sequences for $O=1,2,3,4,5,7,10,12,15,17,20$, their ratios are shown in the upper-right plot. In the above plots, the lines with $\circ$ are for the methods with parameters compressed, the lines with $\times$ are for the methods without parameters compressed, the dashed lines are for the methods with Gaussian priors, and the dotted lines are for the methods with Cauchy priors. The lower-left plot shows the training times for the methods with and without parameters compressed. The lower-right plot shows the prediction time only for the methods with parameters compressed. }

\label{fig-hmm500-comp}

\end{figure}

Figure~\ref{fig-hmm500-comp} compares the number of parameters and the times used to train the model, with and without our compression method.  It is clear that our method for compressing parameters reduces greatly the number of parameters. The ratio of the number of compressed parameters to the number of the original ones decreases with the number of preceding states, $O$. For example, the ratio reaches $0.207$ when $O=20$. This ratio will reduce to $0$ when considering even bigger $O$, since the number of original parameters will grow with $O$ while the number of compressed parameters will converge to a finite number, as discussed in Section~\ref{sec-comp-blsm}. There are similar reductions for the training times with our compression method. But the training time with compressed parameters will not converge to a finite amount, since the time used to update the hyperparameters ($\sigma_o$'s) grows with order, $O$. Figure~\ref{fig-hmm500-comp} also shows the prediction times for $5000$ training cases. The small prediction times show that the methods for splitting Gaussian and Cauchy variables are very fast. The prediction times grow with $O$ because the time used to identify the patterns in a superpattern expressed by a test case grows with $O$. The prediction times with the original parameters are not shown in Figure~\ref{fig-hmm500-comp}, since we do not claim that our compression method saves prediction time. (If we used the time-optimal programming method for each method, the prediction times with compressed parameters should be more than without compressing parameters since the method with compression should include times for identifying the patterns from the superpattern for test cases. With our software, however, prediction times with compression are less than without compression, which is not shown in Figure~\ref{fig-hmm500-comp}, because the method without compression needs to repeatedly read a huge number of the original parameters into memory from disk.)

\begin{figure}[p]

\begin{center}

\includegraphics[scale=0.65]{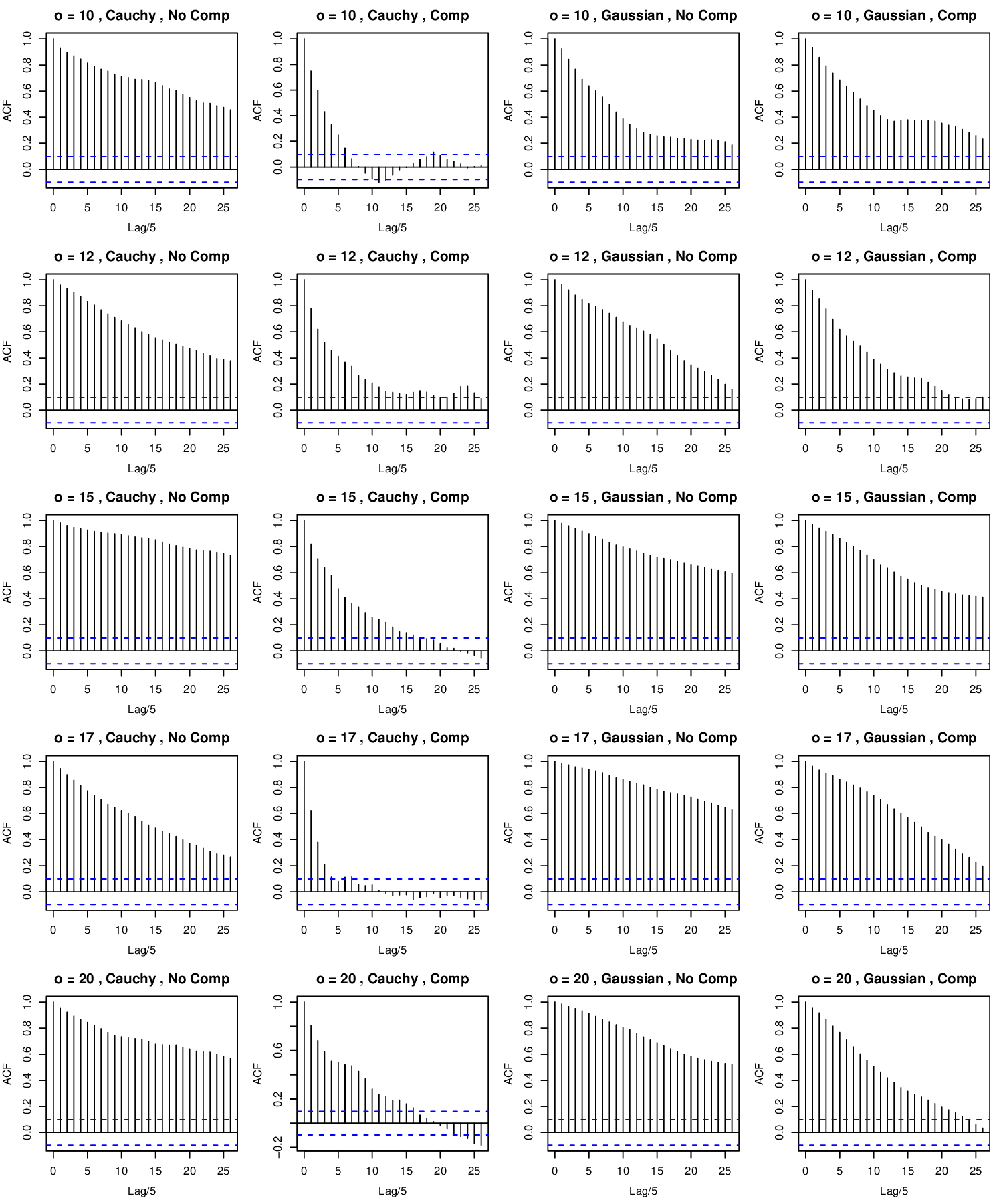}

\end{center}

\vspace*{-0.3in}\caption[The autocorrelation plots of the Markov chains of $\sigma_o$'s for the experiments on a data set generated by a HMM]{The autocorrelation plots of $\sigma_o$'s for the experiments on a data set generated by a HMM, when the length of the preceding sequence $O=20$. We show the autocorrelations of $\sigma_o$, for $o=10,12,15,17,20$. In the above plots, ``Gaussian'' in the titles indicates the methods with Gaussian priors, ``Cauchy'' indicates with Cauchy priors, ``comp'' indicates with parameters compressed, ``no comp'' indicates without parameters compressed.}

\label{fig-hmm500-acf}

\end{figure}

\begin{figure}[p]

\begin{center}

\includegraphics[height=6in,width=6.4in]{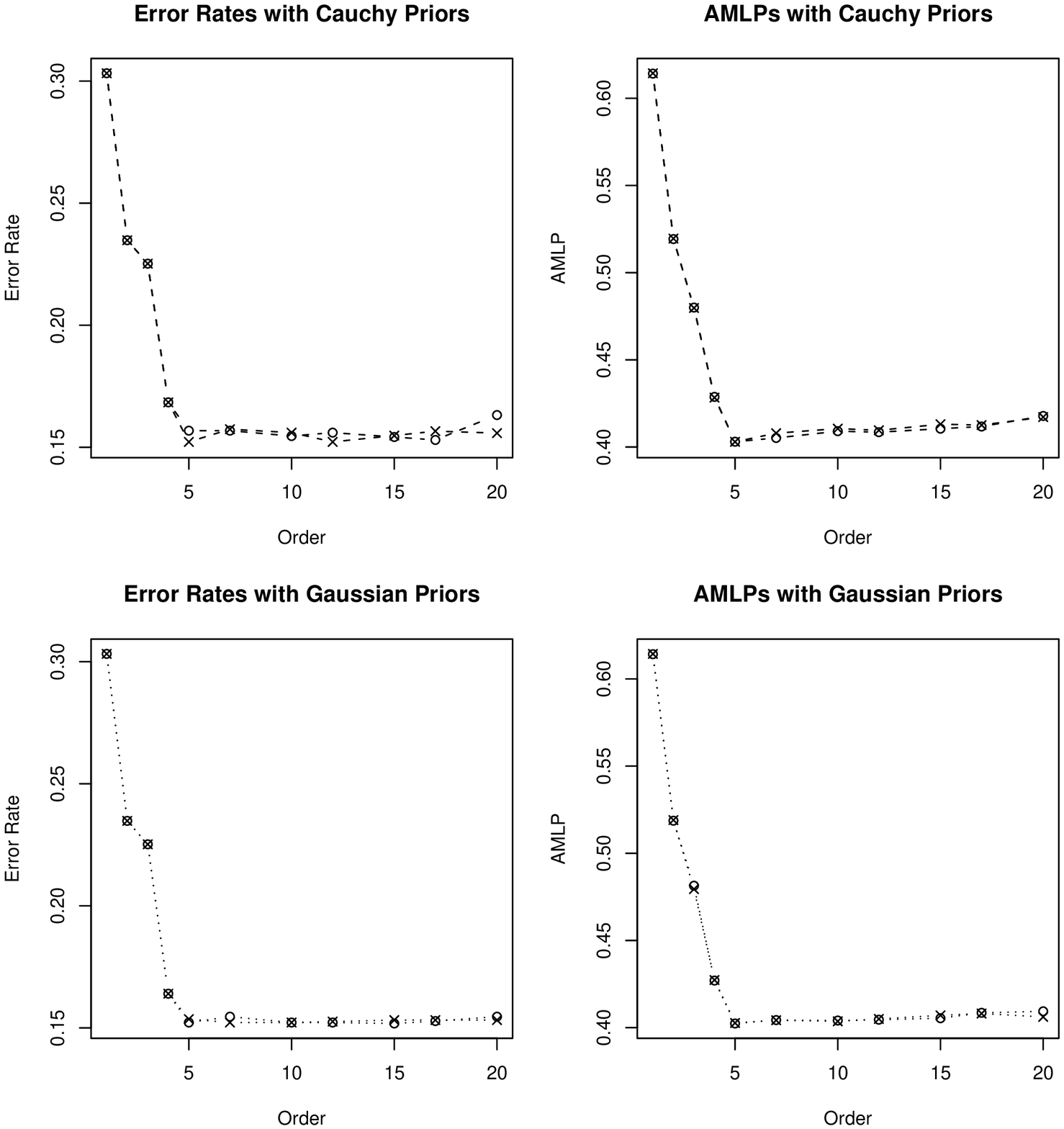}

\end{center}

\caption[Plots showing the predictive performance using the experiments on a
data set generated by a HMM]{Plots showing the predictive performance using the
experiments on a data set generated by a HMM. The left plots show the error
rates and the right plots show the average minus log probabilities of the true
responses in the test cases. The upper plots show the results when using the
Cauchy priors and the lower plots shows the results when using the Gaussian
priors. In all plots, the lines with $\circ$ are for the methods with parameters
compressed, the lines with $\times$ are for the methods without parameters
compressed. The numbers of the training and test cases are respectively $500$
and $5000$. The number of classes of the response is $2$.}

\label{fig-hmm500-error}

\end{figure}

Compressing parameters also improves the quality of Markov chain sampling. Figure~\ref{fig-hmm500-acf} shows the autocorrelation plots of the hyperparameters $\sigma_o$, for $o=10,12,15,17,20$, when the length of the preceding sequence, $O$, is $20$. It is clear that the autocorrelation decreases more rapidly with lag when we compress the parameters. This results from the compressed parameters capturing the important directions of the likelihood function (i.e. the directions where a small change can result in large a change of the likelihood). We did not take the time reduction from compressing parameters into consideration in this comparison. If we rescaled the lags in the autocorrelation plots according to the computation time, the reduction of autocorrelation of Markov chains with the compressed parameters would be much more pronounced. 

Finally, we evaluated the predictive performance in terms of error rate (the fraction of wrong predictions in test cases), and the average minus log probability (AMLP) of observing the true response in a test case based on the predictive probability for different classes. The performance of with and without compressing parameters are the same, as should be the case in theory, and will be in practice when the Markov chains for the two methods converge to the same modes. Performance of methods with Cauchy and Gaussian priors is also similar for this example. The predictive performance is improved when $O$ goes from $1$ to $5$. When $O > 5$ the predictions are slightly worse than with $O=5$ in terms of AMLP. The error rates for $O>5$ are almost the same as for $O=5$. This shows that the Bayesian models can perform reasonably well even when we consider a very high order, as they avoid the overfitting problem in using complex models. We therefore do not need to restrict the order of the Bayesian sequence prediction models to a very small number, especially after applying our method for compressing parameters.


\subsection{Experiments with English Text} \label{sec-text-blsm}

We also tested our method using a data set created from an online article from the website of the Department of Statistics, University of Toronto. In creating the data set, we encoded each character as $1$ for vowel letters (a,e,i,o,u), $2$ for consonant letters, and $3$ for all other characters, such as space, numbers, special symbols, and we then collapsed multiple occurrences of ``$3$'' into only $1$ occurrence. The length of the whole sequence is 3930. Using it we created a data set with $3910$ overlaped sequences of length $21$, and used the first $1000$ as training data. 

\begin{figure}[p]

\begin{center}

\includegraphics[scale=0.8]{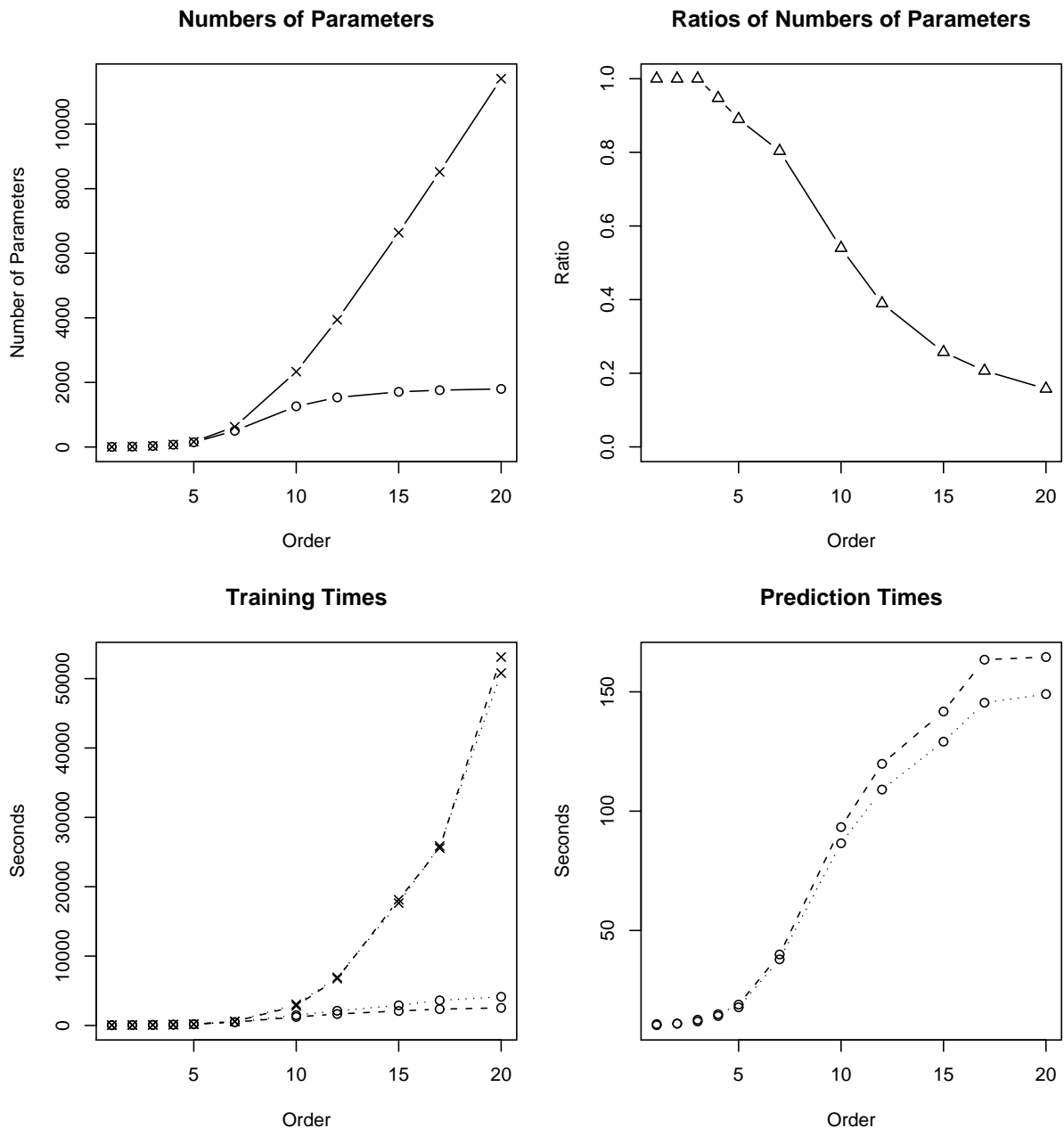}

\end{center}

\vspace*{-0.3in}\caption[Plots showing the reductions of the number of parameters and the training time with our compression method using the experiments on English text]{Plots showing the reductions of the number of parameters and the training and prediction time with our compression method using the experiments on English text. The upper-left plot shows the number of the compressed and the original parameters based on $500$ training sequences for $O=1,2,3,4,5,7,10,12,15,17,20$, their ratios are shown in the upper-right plot. In the above plots, the lines with $\circ$ are for the methods with parameters compressed, the lines with $\times$ are for the methods without parameters compressed, the dashed lines are for the methods with Gaussian priors, and the dotted lines are for the methods with Cauchy priors. The lower-left plot shows the training times for the methods with and without parameters compressed. The lower-right plot shows the prediction time only for the methods with parameters compressed. }

\label{fig-text-comp}

\end{figure}

\begin{figure}[p]

\begin{center}

\includegraphics[scale=0.65]{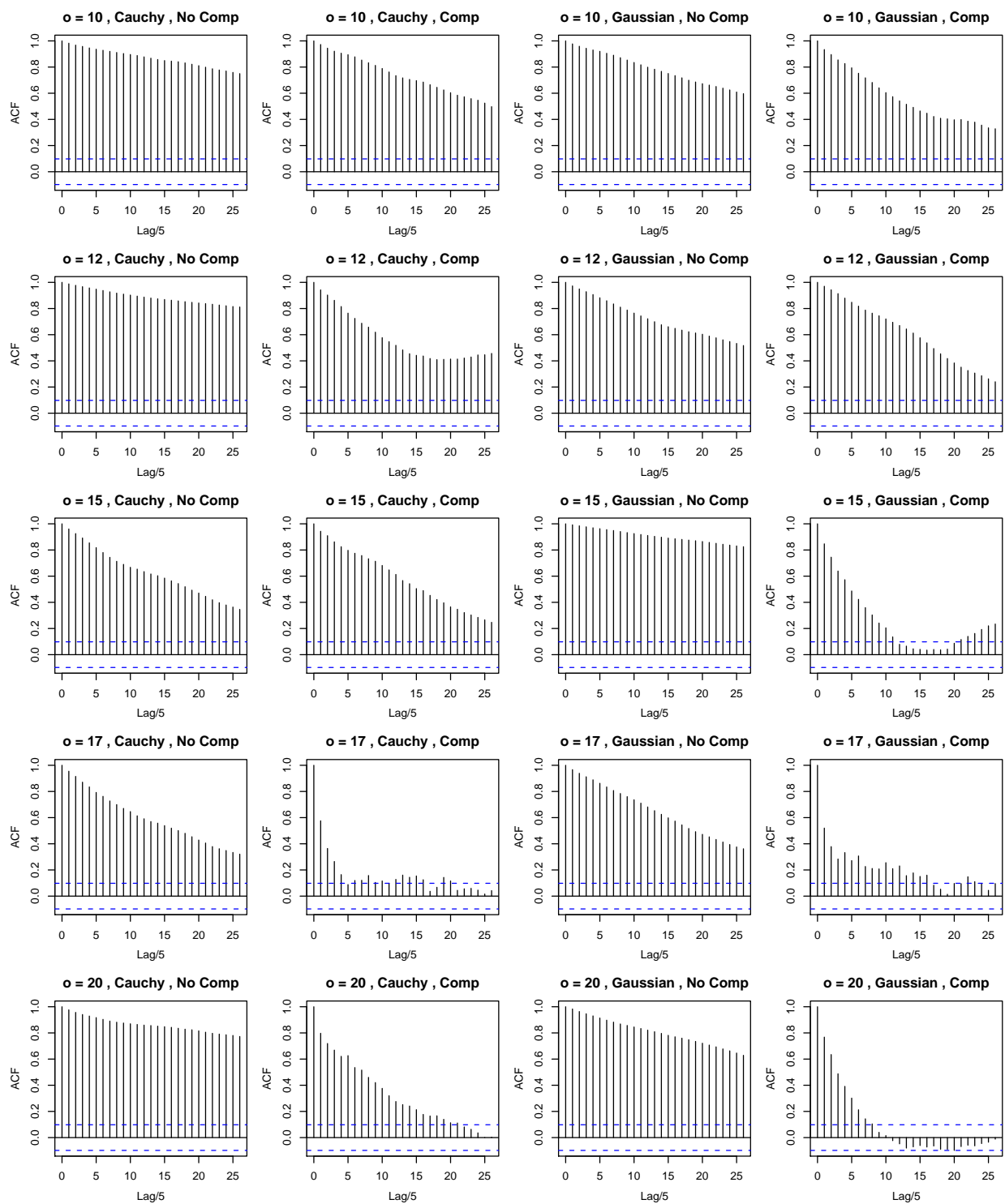}

\end{center}

\vspace*{-0.3in}\caption[The autocorrelation plots of the $\sigma_o$'s for the experiments on English text data]{The autocorrelation plots of the $\sigma_o$'s for the experiments on English text data, when the length of the preceding sequence $O=20$. We show the autocorrelation plot of $\sigma_o$, for $o=10,12,15,17,20$. In the above plots, ``Gaussian'' in the titles indicates the methods with Gaussian priors, ``Cauchy'' indicates with Cauchy priors, ``comp'' indicates with parameters compressed, ``no comp'' indicates without parameters compressed.}

\label{fig-text-acf}

\end{figure}

\begin{figure}[p]

\begin{center}

\includegraphics[height=6in,width=6.4in]{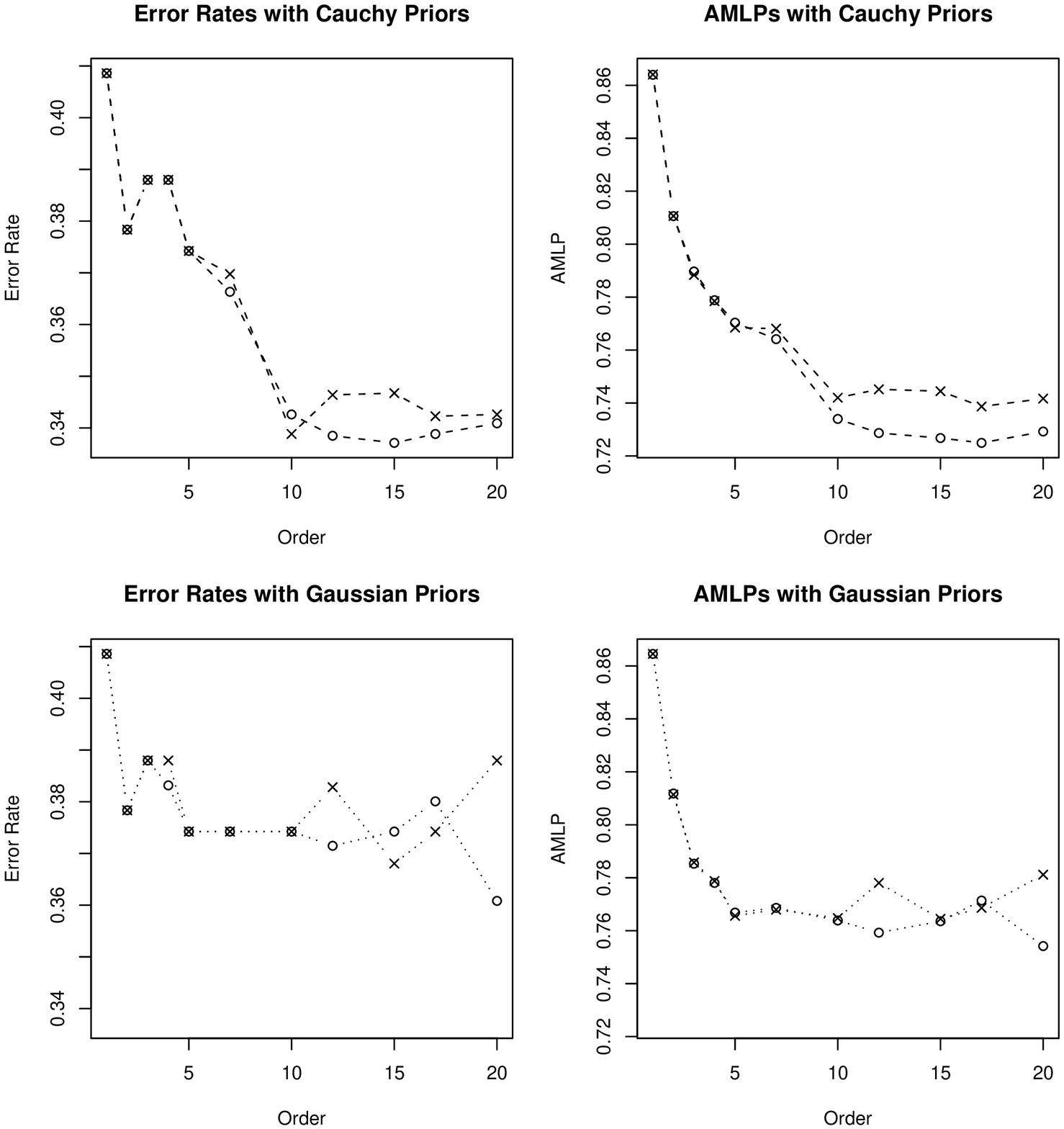}

\end{center}

\caption[Plots showing the predictive performance using the experiments on English text data]{Plots showing the predictive performance using the experiments on English text data. The left plots show the error rate and the right plots show the average minus log probability of the true response in a test case. The upper plots show the results when using the Cauchy priors and the lower plots shows the results when using the Gaussian priors. In all plots, the lines with $\circ$ are for the methods with parameters compressed, the lines with $\times$ are for the methods without parameters compressed. The numbers of the training and test cases are respectively $1000$ and $2910$. The number of classes of the response is $3$.}

\label{fig-text-error}

\end{figure}

\begin{figure}[t]

\begin{center}

\includegraphics[scale=0.65]{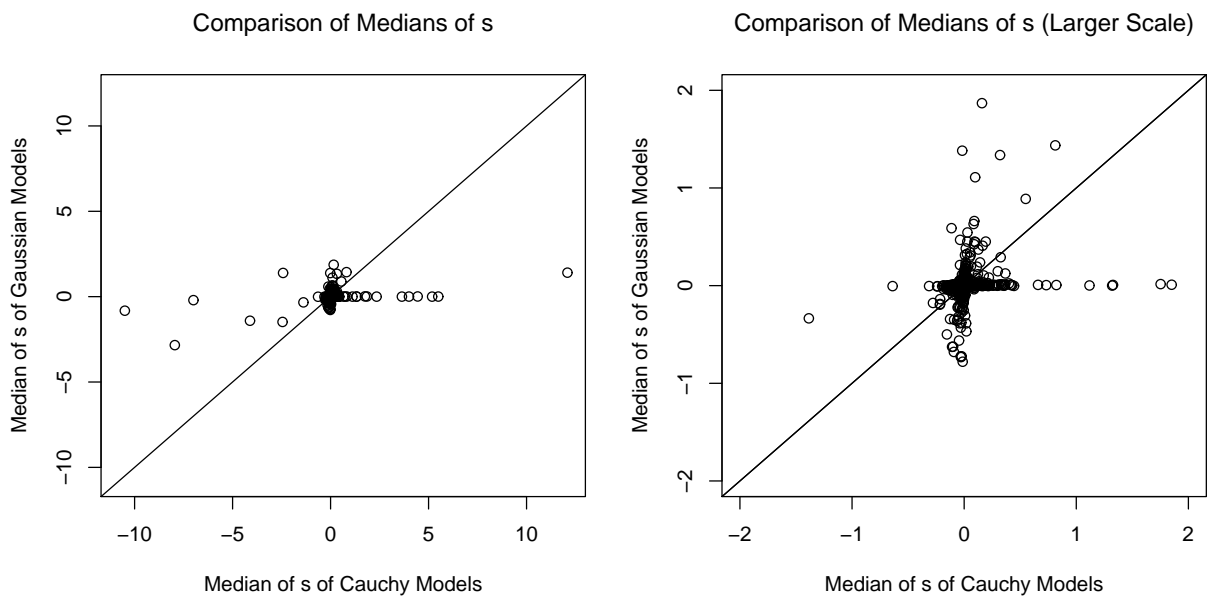}

\end{center}

\caption[Scatterplots of the medians of all the compressed parameters,for the models with Cauchy and Gaussian priors, fitted with English text data]{Scatterplots of medians of all compressed parameters, $s$, of Markov chain samples in the last $1250$ iterations, for the models with Cauchy and Gaussian priors, fitted with English text data, with the length of preceding sequence $O=10$, and with the parameters compressed. The right plot shows in a larger scale the rectangle $(-2,2)\times (-2,2)$. }

\label{fig-text-medians}

\end{figure}

The experiments were similar to those in Section~\ref{sec-sim-blsm}, with the same priors and the same computational specifications for Markov chain sampling. Figures~\ref{fig-text-comp},  \ref{fig-text-acf}, \ref{fig-text-error}, and \ref{fig-text-medians}  show the results. All the conclusions drawn from the experiments in Section~\ref{sec-sim-blsm} are confirmed in this example, with some differences in details. In summary, our compression method reduces greatly the number of parameters, and therefore shortens the training process greatly. The quality of Markov chain sampling is improved by compressing parameters. Prediction is very fast using our splitting methods. The predictions on the test cases are improved by considering higher order interactions. From Figure~\ref{fig-text-error}, at least some order $10$ interactions are useful in predicting the next character.

In this example we also see that when Cauchy priors are used Markov chain sampling with the original parameters may have been trapped in a local mode, resulting in slightly worse predictions on test cases than with the compressed parameters, even though the models used are identical.

\begin{figure}[p]

\begin{center}

\includegraphics[width=6.5in]{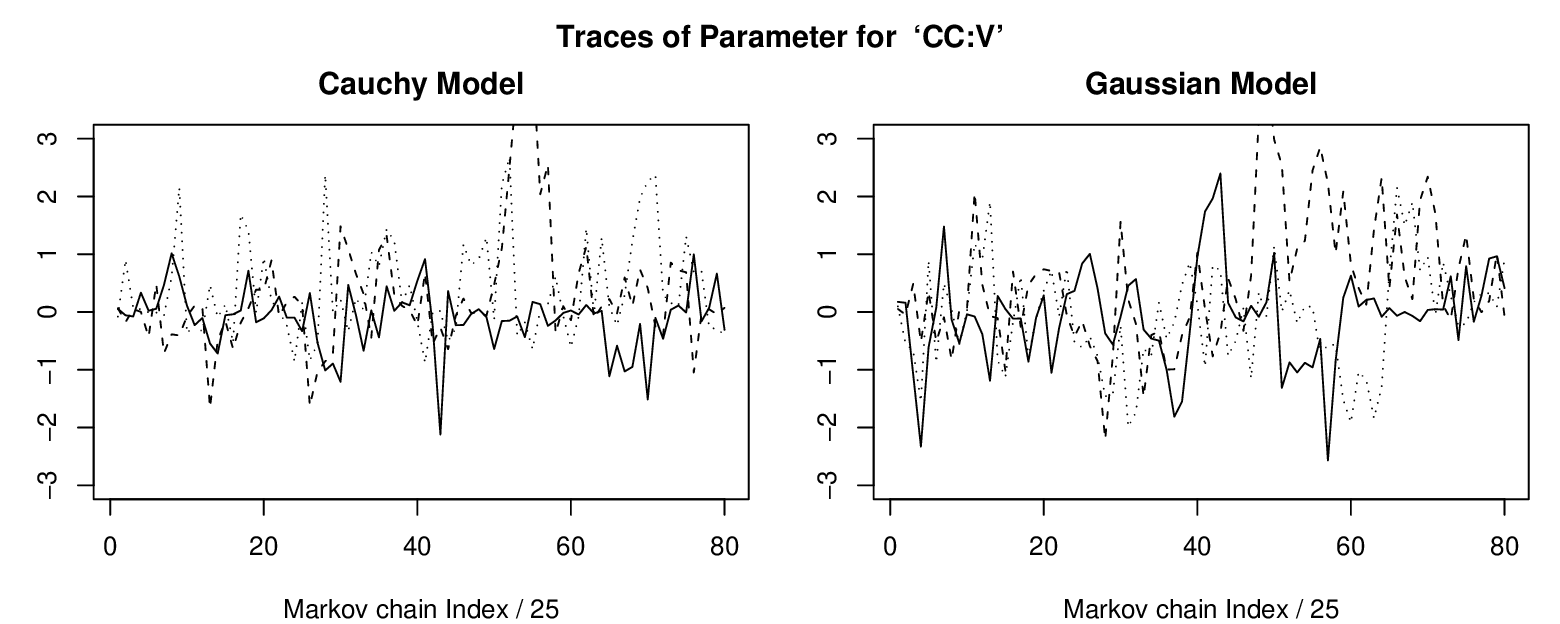}

\vspace*{0.1in}

\hrule


\includegraphics[width=6.5in]{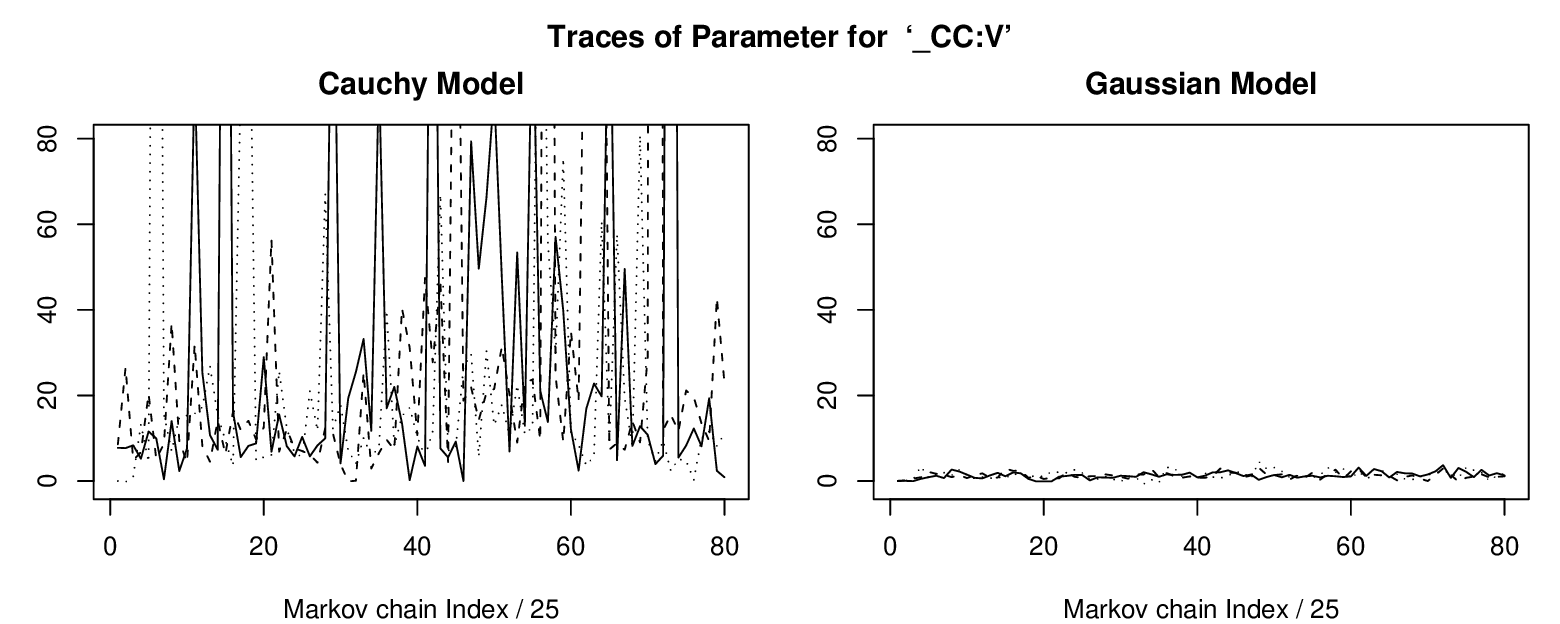}

\vspace*{0.1in}

\hrule


\includegraphics[width=6.5in]{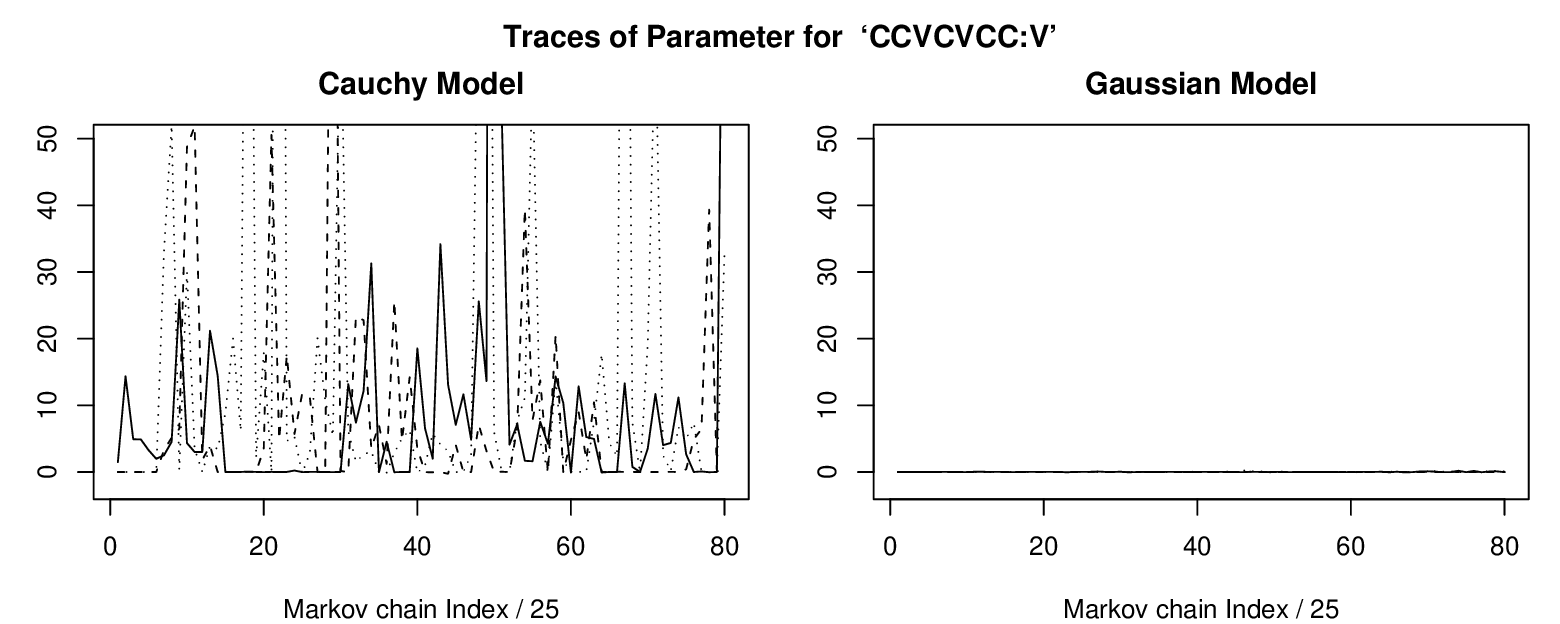}

\end{center}

\vspace*{-0.3in}\caption[Plots of Markov chain traces of three compressed parameters from Experiments on English text]{Plots of Markov chain traces of three compressed parameters (each contains only one $\beta$) from experiments on English text with $10$ preceding states, with Cauchy or Gaussian priors. In each plot, three different lines show three indepedent runs. The parameters are annotated by their original meanings in English sequence. For example, `\_\_CC:V' stands for the parameter for predicting that the next character is a ``vowel'' given preceding three characters are ``space, consonant, consonant''.}

\label{fig-text-betas}

\end{figure}

We also see that the models with Cauchy priors result in better predictions than those with Gaussian priors for this data set, as seen from the plots of error rates and AMLPs. To investigate the difference of using Gaussian and Cauchy priors, we first plotted the medians of Markov chains samples (in the last $1250$ iterations) of all compressed parameters, $s$, for the model with $O=10$, shown in Figure~\ref{fig-text-medians}, where the right plot shows in a larger scale the rectangle $(-2,2)\times(-2,2)$.  This figure shows that a few $\beta$ with large medians in the Cauchy model have very small corresponding medians in the Gaussian model.

We also looked at the traces of some compressed parameters, as shown in Figure~\ref{fig-text-betas}. The three compressed parameters shown all contain only a single $\beta$. The plots on the top are for the $\beta$ for ``CC:V'', used for predicting whether the next character is a vowel given the preceding two characters are consonants; the plots in the middle are for ``\_\_CC:V", where ``\_\_'' denotes a space or special symbol; the plots on the bottom are for ``CCVCVCC:V'', which had the largest median among all compressed parameters in the Cauchy model, as shown by Figure~\ref{fig-text-medians}. The regression coefficient $\beta$ for ``CC:V'' should be close to $0$ by our common sense, since two consonants can be followed by any of three types of characters. We can very commonly see ``CCV'', such as ``the'',  and  ``CC\_\_'', such as ``with\_\_'', and not uncommonly see ``CCC'', such as ``technique'',``world'', etc. The Markov chain trace of this $\beta$ with a Cauchy prior moves in a smaller region around $0$ than with a Gaussian prior. But if we look back one more character, things are different. The regression coefficient $\beta$ for ``\_\_CC:V" is fairly large, which is not surprising. The two consonants in ``\_\_CC:V" stand for two letters in the beginning of a word. We rarely see a word starting with three consonants or a word consisting of only two consonants. The posterior distribution of this $\theta$ for both Cauchy and Gaussian models favor positive values, but the Markov chain trace for the Cauchy model can move to much larger values than for the Gaussian model. As for the high-order pattern ``CCVCVCC'', it matches words like  ``statistics'' or ``statistical'', which repeatedly appear in an article introducing a statistics department. Again, the Markov chain trace of this $\beta$ for the Cauchy model can move to much larger values than for Gaussian model, but sometimes it is close to $0$, indicating that there might be two modes for its posterior distribution.

The above investigation reveals that a Cauchy model allows some useful $\beta$ to be much larger in absolute value than others while keeping the useless $\beta$ in a smaller region around $0$ than a Gaussian model. In other words, Cauchy models are more powerful in finding the information from the many possible high-order interactions than Gaussian models, due to the heavy two-sided tails of Cauchy distributions.

\section{Application to Logistic Classification Models}
\label{sec-blcm}

\subsection{Grouping Parameters of Classification Models}
\label{sec-comp-blcm}

As we have seen in sequence prediction models, the regression coefficients for the patterns that are expressed by the same training cases can be compressed into a single parameter. We present an algorithm to find such groups of patterns. Our algorithm may not the only one possible and may not be the best.

Our algorithm uses a ``superpattern'', $SP$, to represent a set of patterns with some common property, written as $\mychoose{A_1\ldots A_p}{I_1\ldots I_p}_f^o$\,, where $A_t$ is the pattern value for position $t$ (an integer from $0$ to $K_t$), $I_t$ is binary ($0/1$) with $1$ indicating $A_t$ is fixed for this superpattern, $f$ is the number of fixed positions (ie $f=\sum_{t=1}^p I_t$), and $o$ indicates the smallest order of all patterns in this superpattern, equal to the sum of nonzero values of those $A_t$ with $I_t=1$ (i.e. $o=\sum_{t=1}I(I_t=1,A_t\not=0)$). Such a superpattern represents the union of all patterns with order not greater than $O$, with values at the fixed positions (with $I_t=1$) being $A_t$, but the pattern values at unfixed positions (with $I_t=0$) being either $0$ or $A_t$. For example, if $O=3$, the superpattern $\mychoose{12304}{10010}_2^1$\, is composed of $\mychoose{3}{0}$, $\mychoose{3}{1}$, and $\mychoose{3}{2}$ patterns respectively of order $1,2$ and $3$, listed as follows: 

\noindent\beq
\begin{array}{cccc}
\mbox{order}\ 1: & [10000] &        &       \\
\mbox{order}\ 2: & [12000],& [10300],& [10004] \\
\mbox{order}\ 3: & [10304],& [12004],& [12300] 
\end{array}
\eeq

\begin{figure}[p]

\begin{center}

\includegraphics[scale=0.85]{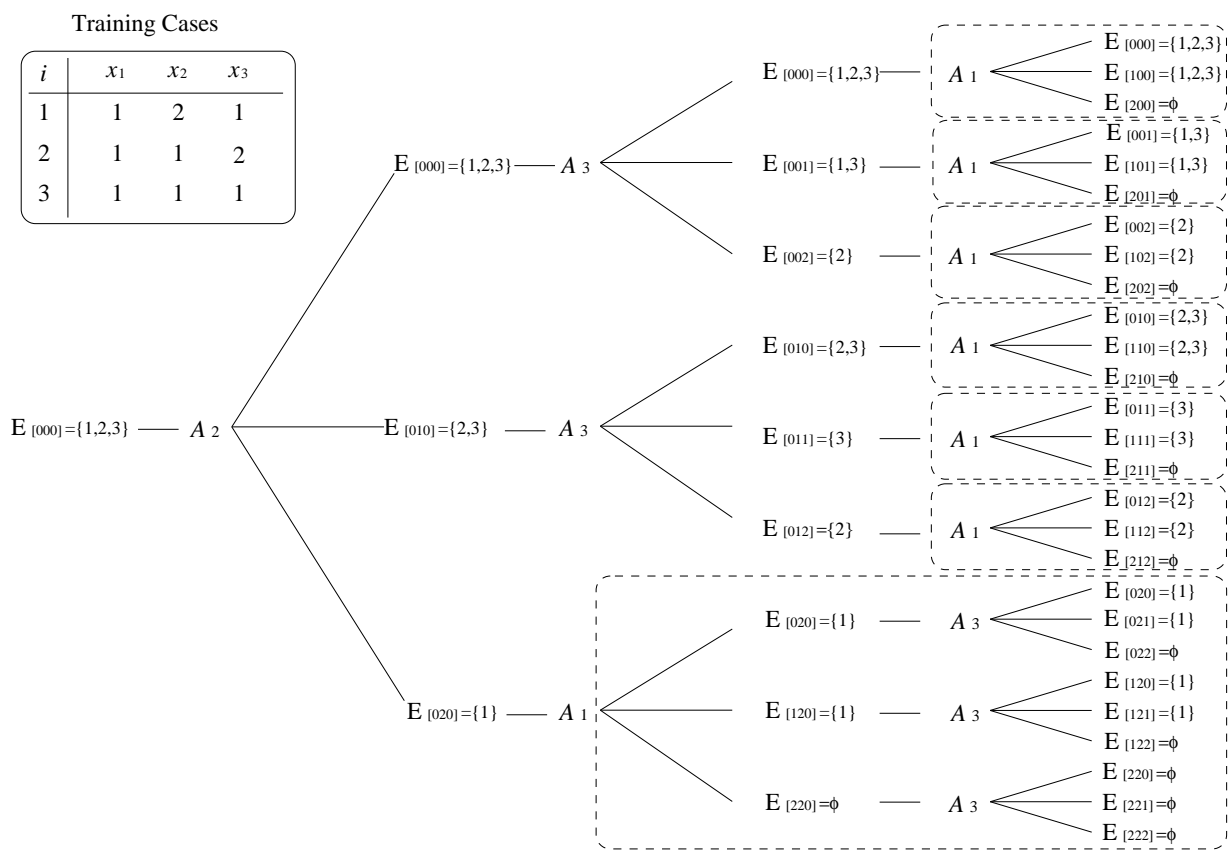}

\end{center}

\caption[The picture illustrates the algorithm for grouping the patterns of classification models]{This picture illustrates the algorithm for grouping the patterns of classification models using a training data of $3$ binary features and $3$ cases shown on the left-top corner. Starting from the expression for the intercept and all features in the unconsidered features set, we recursively split the current expression by the values of the feature with the biggest entropy (diversity) among the remaining unconsidered features. When the values of all remaining unconsidered features are the same for all training cases, for example, when the expression contains only one training case, all the following patterns can be grouped together. After grouping, all the expressions in dashed box are removed and grouped into their parent expressions, with the group of patterns being represented using a superpattern.}

\label{fig-group-cls}

\end{figure}

\begin{figure}[p]

\begin{center}

\includegraphics[scale=0.95]{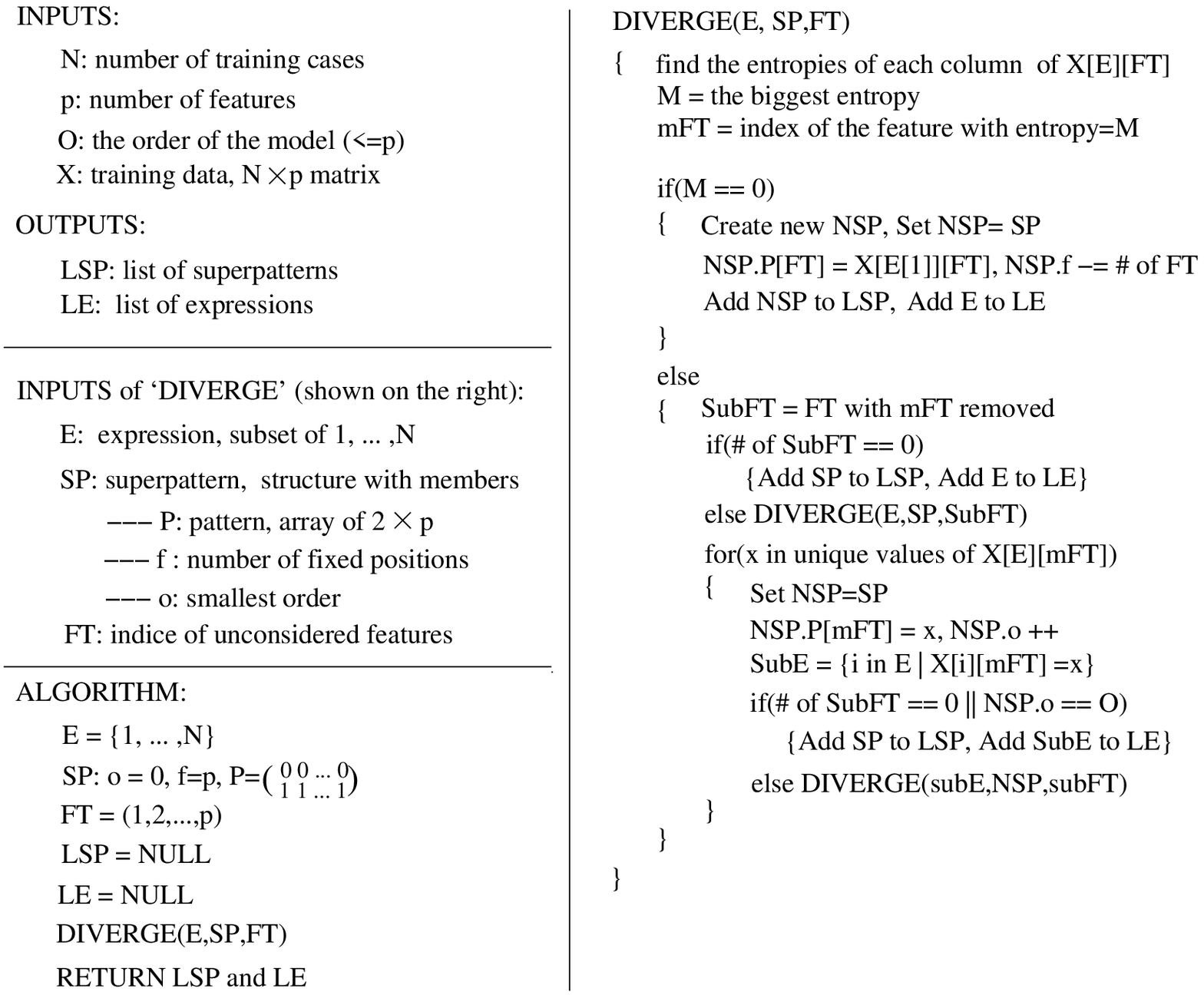}

\end{center}

\caption[The algorithm for grouping the patterns of Bayesian logistic classification models]{The algorithm for grouping the patterns of Bayesian logistic classification models. To do grouping, we call the function ``DIVERGE'' with the initial values of expression $E$, superpattern $SP$ and the unconsidered features $FT$ shown as above, resulting in two lists of the same length, LE and LSP, respectively storing the expressions and the corresponding superpatterns. Note that the first index of an array are assumed to be $1$, and that$X[E][FT]$ means sub matrix of $X$ by restricting the rows in $E$ and columns in $FT$. 3) The resulting expressions are not unique for each superpattern in $LSP$. We still need to merge those superpatterns with the same expression by directly comparing the expressions.}

\label{fig-alg-cls}

\end{figure}

The algorithm is inspired by the display of all interaction patterns in Figure~\ref{fig-ptn-cls}. It starts with the expression $\{1,\ldots,N\}$ for the superpattern $\mychoose{00\ldots 0}{11\ldots 1}_p^0$\,, and the unconsidered features $(1,\ldots,p)$. After choosing a feature $x_t$ from the unconsidered features by the way described below, the current expression is split for each value of the $x_t$, as done by the algorithm for sequence prediction models, additionally the whole expression is also passed down for the pattern with $A_t=0$. When we see that we can not split the expression by any of the unconsidered features, all the following patterns can be grouped together and represented with a superpattern. In Figure~\ref{fig-group-cls}, we use training data with $3$ binary features and only $3$ cases to illustrate the algorithm. We give the algorithm in a C-like language in Figure~\ref{fig-alg-cls}, which uses a recursive function. 

In choosing a feature for splitting a expression, we look at the diversity of the values of the unconsidered features restricted on the current expression. By this way the expression is split into more expressions but each may be smaller. We therefore more rapidly reach the expression that can not be split further. The diversity of a feature $x_t$ restricted on the current expression is measured with the entropy of the relative frequency of the values of $x_t$, i.e., $-\sum_i p_i \log(p_i)$, where $p_i$'s are the relative frequency of the possible values of the feature restricted on the expression. When two features have the same entropy value, we choose the one with smaller index $t$. Note that the entropy is always positive unless all the values of the feature $x_t$ restricted on the expression are the same. The resulting expressions found by this algorithm are not unique for each superpattern.  

In training the model, we need to compute the width of a parameter associated with a superpattern given the values of the hyperparameters $\sigma_o$'s. For Cauchy models, the width is equal to the sum of the hyperparameters of all patterns in the superpattern. For Gaussian models, the width is the square roots of the sum of the squares of the hyperparameters of all patterns in the superpattern.  We therefore need only know the number of the patterns in the superpattern belonging to each order from $0$ to $O$. For a superpattern $\mychoose{A_1\ldots A_p}{I_1\ldots I_p}_f^o$\,, they are given as follows:

\beq
\#(\mbox{patterns of order }o + d) = \left\{
        \begin{array}{ll}
          \mychoose{p-f}{d}, & \mbox{for } d=0,\ldots,\min(O-o,p-f) \\
          0,              & \mbox{Otherwise}
        \end{array} \right.
\label{nopatterns}
\eeq

In predicting for a test case $\x^*$, we need to identify the patterns in a superpattern $\mychoose{A_1\ldots A_p}{I_1\ldots I_p}_f^o$\, that is also expressed by $\x^*$. If $A_t\not=x^*_t$ for any $t$ with $I_t=1$ and $A_t\not=0$, none of the patterns in the superpatterns are expressed by $\x^*$. Otherwise, if $x^*_t\not=A_t$ for any unfixed positions (with $I_t=0$) then the $A_t$ is set to $0$. This results in a smaller superpattern, from which we can count the number of patterns belonging to each order using the formula in~(\ref{nopatterns}).

For a fixed very large number of features, $p$, and a fixed number of cases, $N$, the number of the compressed parameters may have converged before considering the highest possible order, $p$. This can be verified by regarding the interaction patterns of a classification model as the union (non-exclusive) of the interaction patterns of $p!$ sequence models from permutating the indice of features. As shown for sequence models in Section~\ref{compress-seq}, there is a certain order $O_j$, for $j=1,\ldots,p!$, for each of the $p!$ sequence models, the number of superpatterns with unique expressions will not grow after considering order higher than $O_j$. The number of the superpatterns with unique expressions for all $p!$ sequence models will not grow after we consider the order higher than the maximum value of $O_j$, for $j=1,\ldots,p!$. If this maximum value is smaller than $p$, the number of the {\it compressed parameters} converges before considering the highest possible order, $p$. On the contrary, the number of the {\it original parameters} (the regression coefficients for those interaction patterns expressed by some training case) will keep growing until considering the highest order, $p$. 

\subsection{Experiments Demonstrating Parameter Reduction}
\label{sec-sim-blcm1}

In this section, we use simulated data sets to illustrate our compression method. We apply our compression method to many training data sets with different properties to demonstrate how the rate of parameters reduction and the number of compressed parameters depend on the properties of the data, but without running MCMC to train the models and assessing the predictive performance with test cases.

\begin{figure}[p]

\begin{center}

\includegraphics[scale=1]{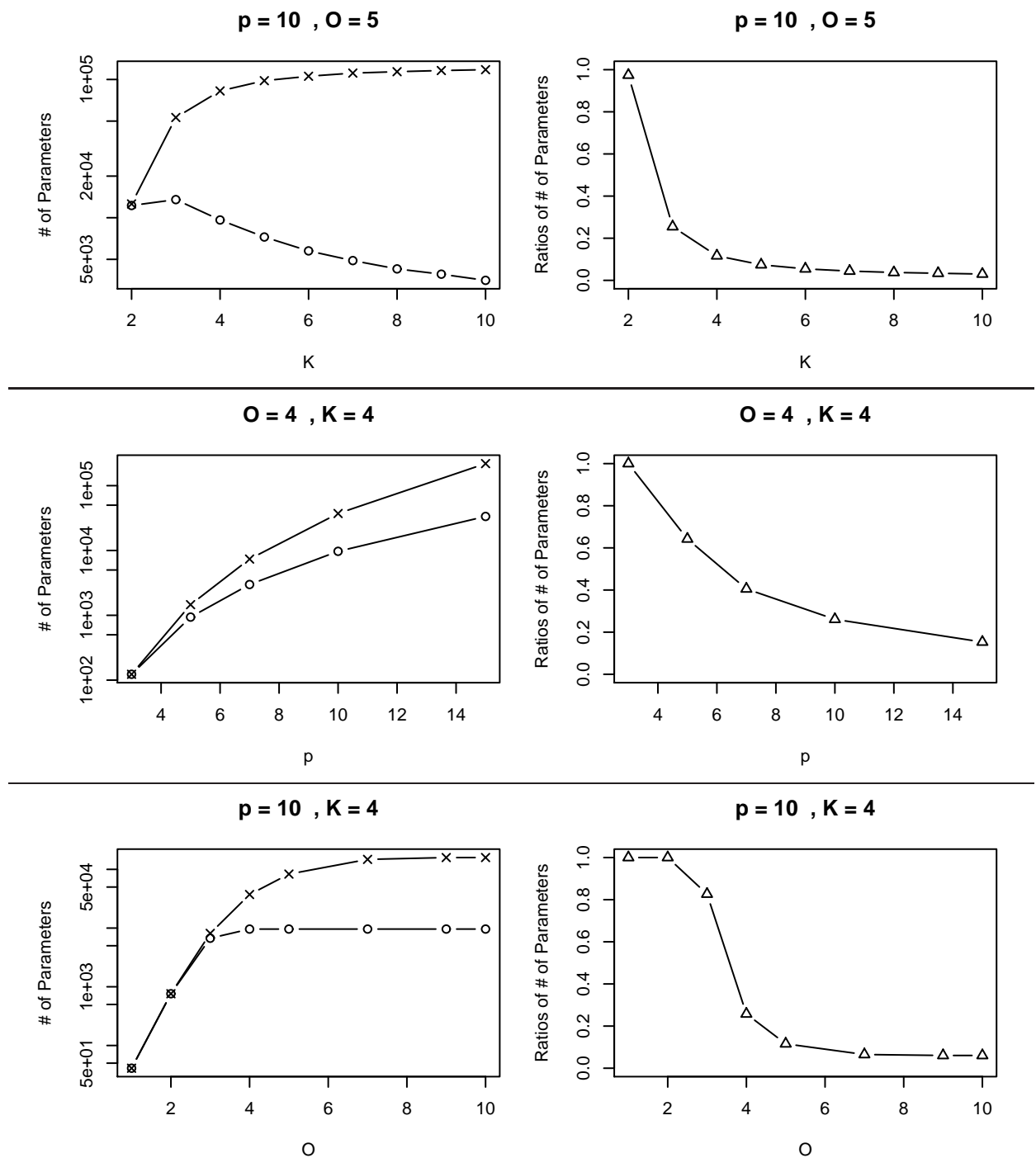}

\vspace*{-4.9in}\hrule \vspace*{4.9in}

\vspace*{-2.4in}\hrule \vspace*{2.4in}

\end{center}

\vspace*{-0.2in}\caption[Plots of the number of compressed parameters and the original parameters]{Plots of the number of the compressed parameters (the lines with $\circ$) and the original parameters (the lines with $\times$), in log scale, their ratios (the lines with $\triangle$) for Bayesian logistic classification models. The number of training cases is $200$ for all data sets. The titles and the horizontal axis labels indicates the values of $p,K$ and $O$ in compressing parameters, where $p$ is the number of features, $K$ is the number of possibilities of each feature, and $O$ is the order of interactions considered .}

\label{fig-comp-cls-notest}

\end{figure}

We generated data sets with varying dimension $p$, and number of possibilities of each feature, $K$, the same for all $p$ features. We consider varying order, $O$. In all datasets, the values of features are drawn uniformly from the set $\{1,\ldots,K\}$, with the number of training cases $N=200$. We did three experiments, in each of which two of the three values $p,K$ and $O$ are fixed, with the remaining one varied to see how the performance of the compression method changes. The values of $p,K$ and $O$ and the results are shown in Figure~\ref{fig-comp-cls-notest}.

From Figure~\ref{fig-comp-cls-notest}, we can informally assess the performance of our compression method in different situations. First, when $p$ and $O$ are fixed, as shown by the top plots, the number of compressed parameters decreases with increasing $K$, but the number of the original parameters does not, showing that our compression method is more useful when $K$ is large, in other words, when $K$ is larger, more patterns that do not need to be represented explicitly will exist. Note, however, that this does not mean the predictive performance for large $K$ is better. On the contrary, when $K$ is larger, each pattern will be expressed by fewer training cases, possibly leading to worse predictive performance. Second, as shown by the middle plots where $O$ and $K$ are fixed, the numbers of both the original parameters and the compressed parameters increase very quickly with $p$, but their ratio decreases with $p$. In the bottom plots with fixed $p$ and $K$, the number of the original parameters increases with $O$ but at a much slower rate than with $p$. The number of compressed parameters have converged when $O=4$, earlier than $O=7$.

\subsection{Experiments with Data from Cauchy Models}

We tested the Bayesian logistic classification models on a data set generated using the true model defined in Section~\ref{sec-def-blcm}. The number of features, $p$, is $7$, and each feature was drawn uniformly from $\{1,2,3\}$. For generating the responses, we consider only the interactions from order $0$ to order $4$, we let $\sigma_o = 1/o$, for $o=1,\ldots,4$, then generated regression coefficients $\beta$'s from Cauchy distributions, as shown by~(\ref{eqn-cc-prior}), except fixing the intercept at $0$. We generated $5500$ cases, of which $500$  were used as training cases and the remainder as test cases. We did experiments with and without the parameters compressed, for order $O=1,\ldots,7$. 

\begin{figure}[p]

\begin{center}

\includegraphics[width=6.5in]{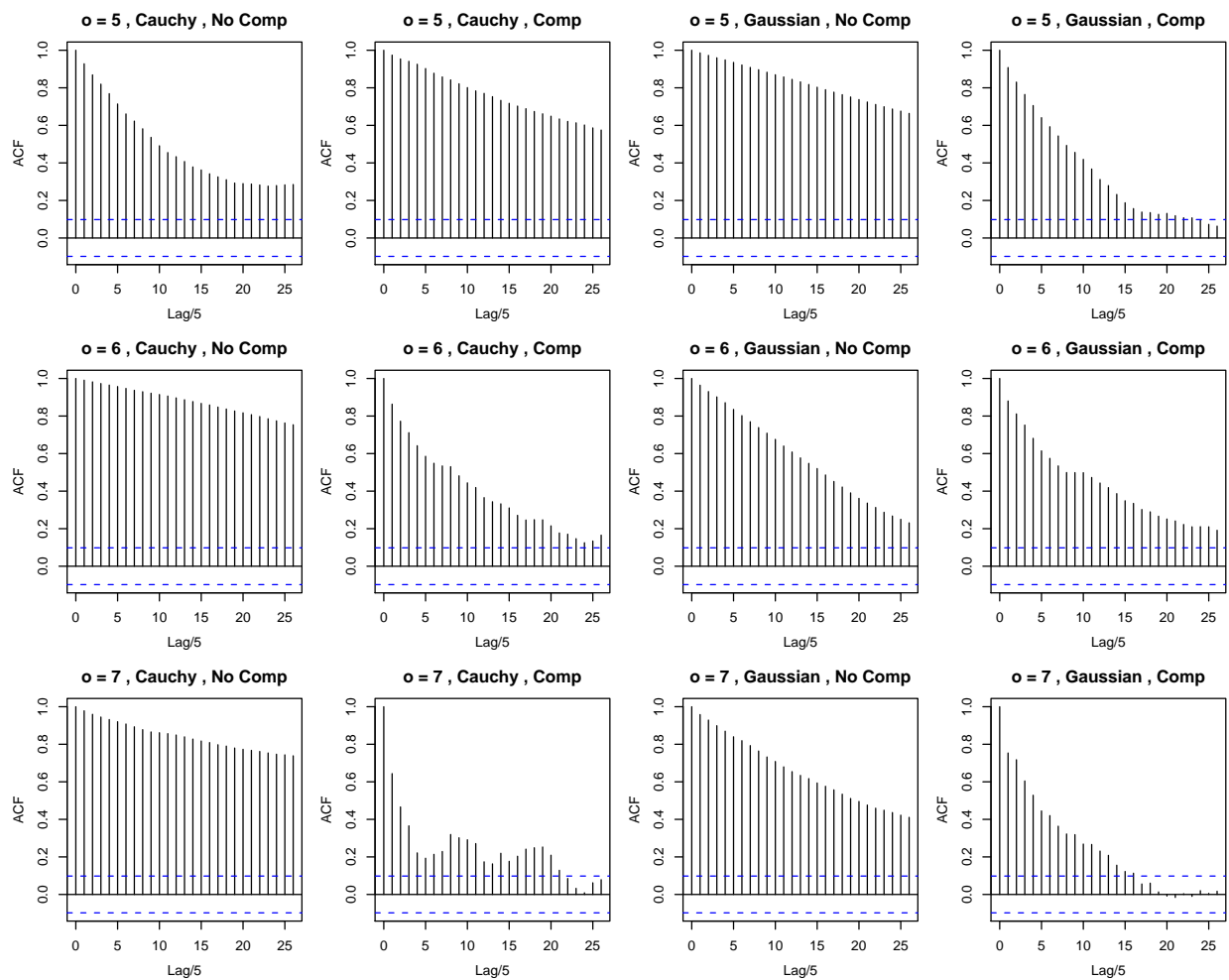}

\end{center}

\vspace*{-0.2in}\caption[The autocorrelation plots of the Markov chains of $\sigma_o$'s for the experiments on a data from a Cauchy model]{The autocorrelation plots of $\sigma_o$'s for the experiments on a data from a Cauchy model, when the order $O=7$. We show the autocorrelations of $\sigma_o$, for $o=5,6,7$. In the above plots, ``Gaussian'' in the titles indicates the methods with Gaussian priors, ``Cauchy'' indicates with Cauchy priors, ``comp'' indicates with parameters compressed, ``no comp'' indicates without compressing parameters.}

\label{fig-clscc-acf}

\end{figure}

\begin{figure}[t]

\begin{center}

\includegraphics[width=6.5in]{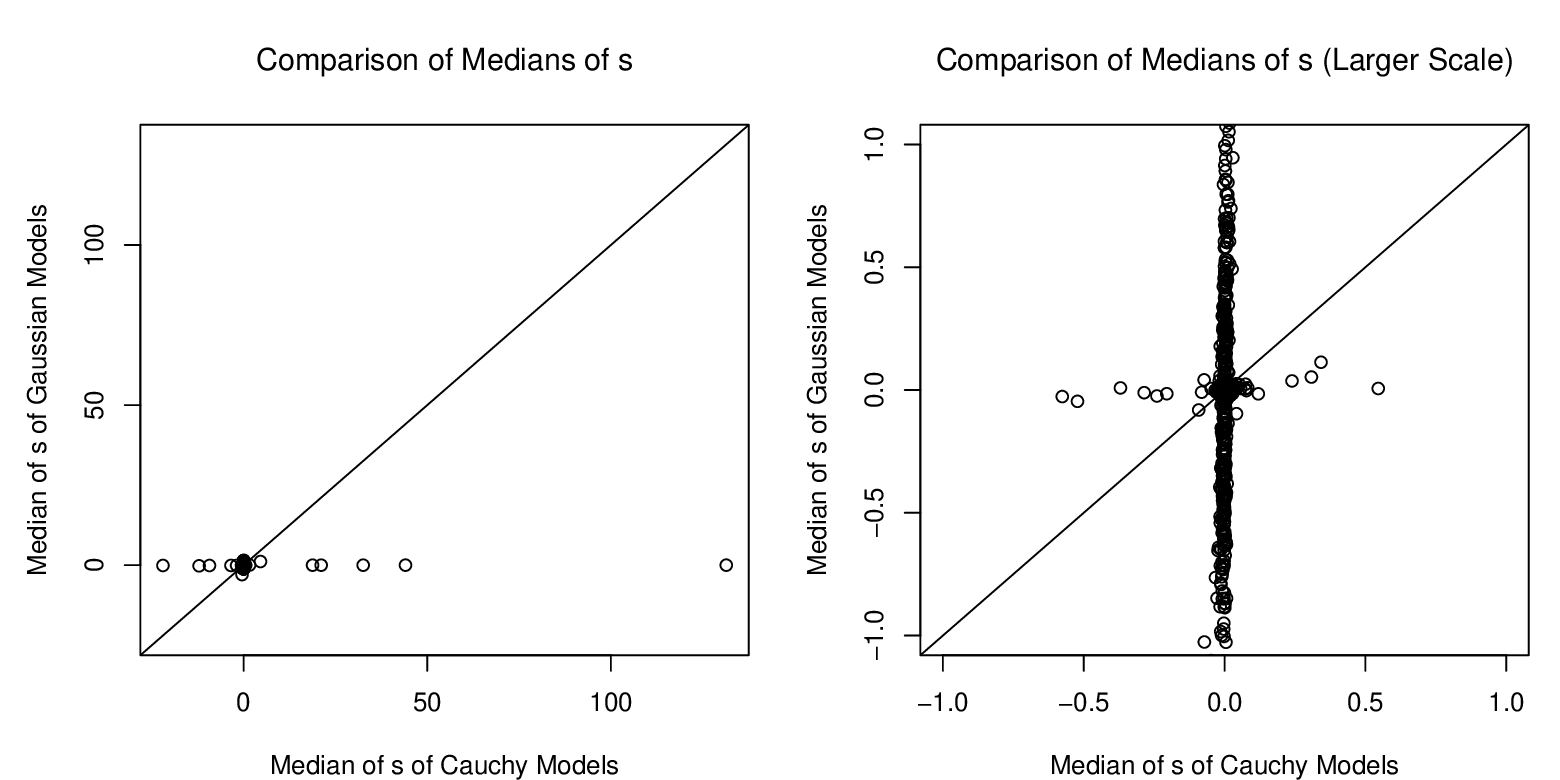}

\end{center}

\caption[Scatterplots of medians of all $\beta$ for the models with Cauchy and Gaussian priors, from the experiment with data from a Cauchy model ]{Scatterplots of medians of all $\beta$ of the last $1250$ iterations of Markov chain samples, for the models with Cauchy and Gaussian priors, from the experiment with data from a Cauchy model, with the order $O=4$, and with the parameters compressed. The right plot shows in a larger scale the rectangle $(-1,1)\times (-1,1)$.}

\label{fig-medians-clscc}

\end{figure}

From these experiments, we see that our compression method can reduce greatly the number of parameters and therefore saves a huge amount of time for training the models with MCMC. The number of the compressed parameters does not grow any more after considering $O=4$, earlier than $O=7$. After compressing the parameters, the quality of Markov chain sampling is improved, seen from Figure~\ref{fig-clscc-acf}, where $5$ out of $6$ experiments show that the autocorrelation of the $\sigma_o$ decreases more rapidly with lag after compressing parameters. The predictive performance with and without the parameters compressed are similar, with the optimal performance obtained from the Cauchy model with order $O=4$, as is expected since this is the true model generating the data set. From the left plot of Figure~\ref{fig-medians-clscc}, we see that a Cauchy model allows some $\beta$ to be much larger than others, whereas a Gaussian model keeps all of $\beta$ in a  small region. For those truly small $\beta$, Cauchy priors can keep them smaller than  Gaussian priors can, as shown by the right plot of Figure~\ref{fig-medians-clscc}.

\begin{figure}[hp]

\begin{center}

\includegraphics[width=6.5in]{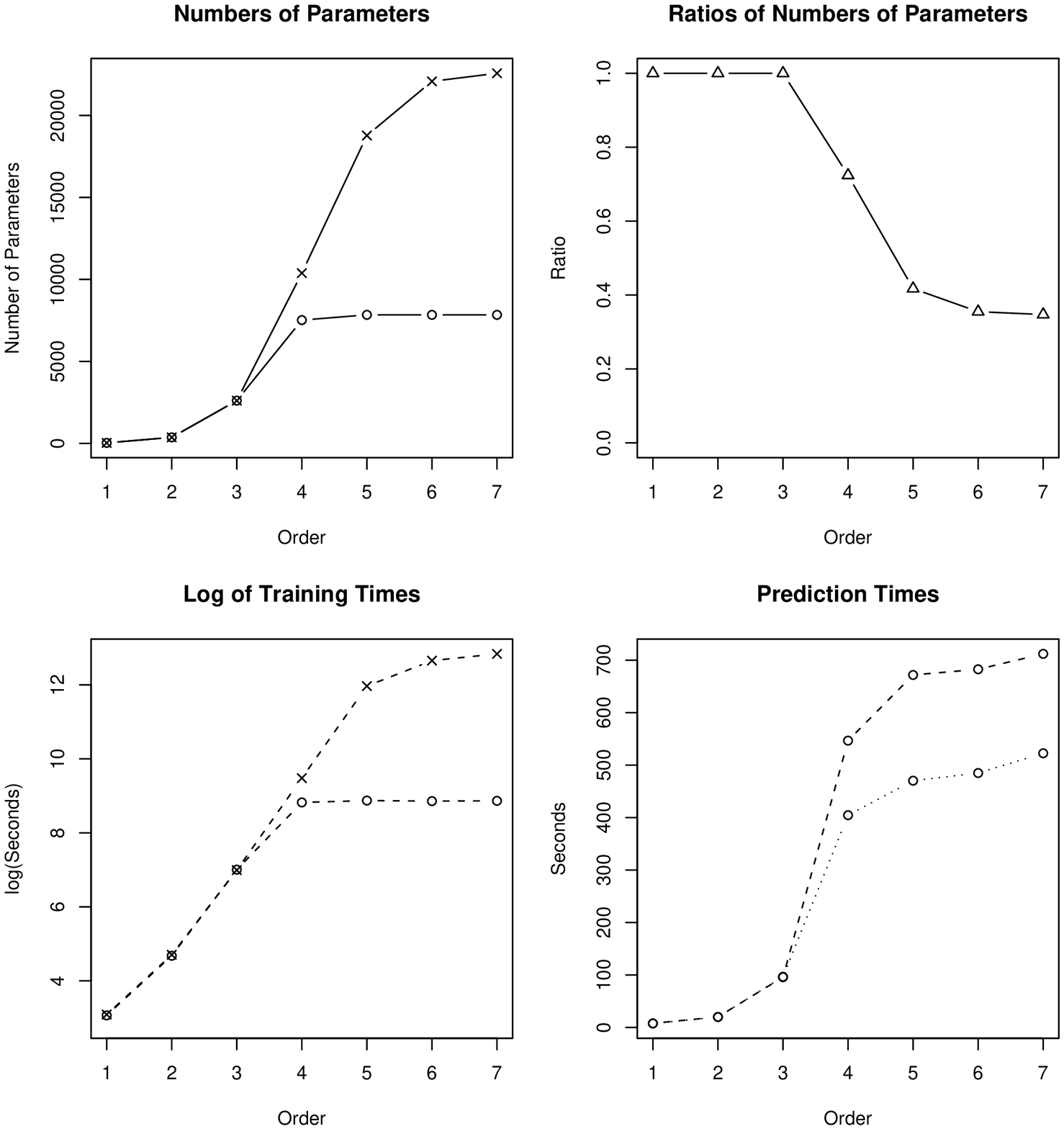}

\end{center}

\vspace*{-0.2in}\caption[Plots showing the reductions of the number of parameters and the training time with our compression method using the experiments on a data from a Cauchy model]{Plots showing the reductions of the number of parameters and the training time with our compression method using the experiments on a data from a Cauchy model. The upper-left plot shows the numbers of the compressed and the original parameters based on $500$ training sequences for $O=1,2,\ldots,7$, their ratios are shown in the upper-right plot. In the above plots, the lines with $\circ$ are for the methods with parameters compressed, the lines with $\times$ are for the methods without parameters compressed, the dashed lines are for the methods with Gaussian priors, and the dotted lines are for the methods with Cauchy priors. The lower-left plot shows the training times for the methods with and without parameters compressed. The lower-right plot shows the prediction time only for  the methods with parameters compressed. }

\label{fig-clscc-comp}

\end{figure}

\begin{figure}[p]

\begin{center}

\includegraphics[width=6.5in]{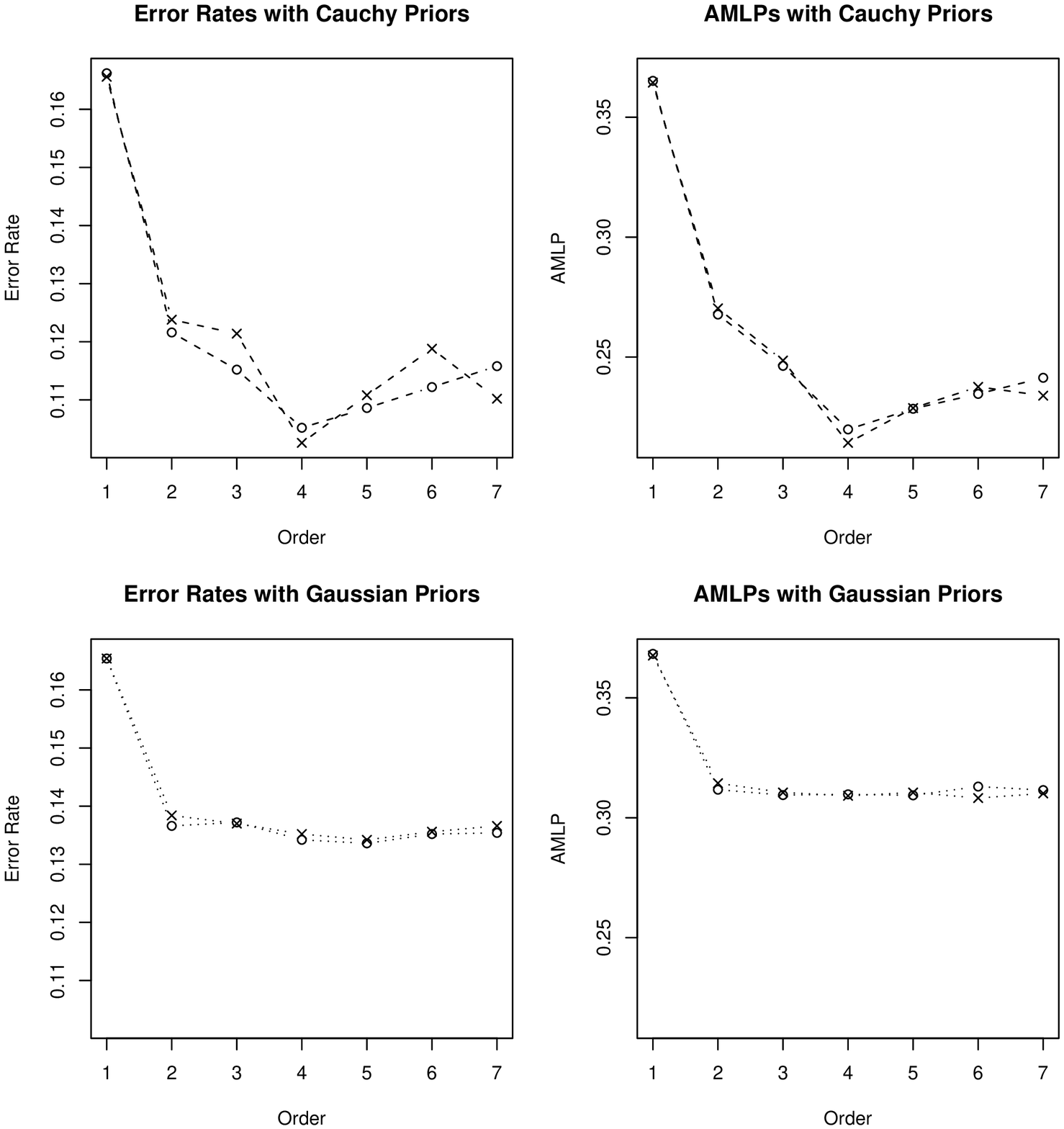}

\end{center}

\caption[Plots showing the predictive performance using the experiments on  data from a Cauchy model]{Plots showing the predictive performance using the experiments on  data from a Cauchy model. The left plots show the error rates and the right plots show the average minus log probabilities of the true responses in the test cases. The upper plots show the results when using the Cauchy priors and the lower plots shows the results when using the Gaussian priors. In all plots, the lines with $\circ$ are for the methods that compress parameters, the lines with $\times$ are for the methods do not  compress parameters. The number of the training and test cases are respectively $500$ and $5000$. The number of classes of the response is $2$.}

\label{fig-clscc-error}

\end{figure}

\section{Conclusion and Discussion}\label{sec-comp-conclude}

In this chapter, we have proposed a method to effectively reduce the number of parameters of Bayesian regression and classification models with high-order interactions, using a compressed parameter to represent the sum of all the regression coefficients for the predictor variables that have the same values for all the training cases. Working with these compressed parameters, we greatly shorten the training time with MCMC. These compressed parameters can later be split into the original parameters efficiently. We have demonstrated, theoretically and empirically, that given a data set with fixed number of cases, the number of compressed parameters will have converged before considering the highest possible order. Applying Bayesian methods to regression and classification models with high-order interactions therefore become much easier after compressing the parameters, as shown by our experiments with simulated and real data. The predictive performance will be improved by considering high-order interactions if some useful high-order interactions do exist in the data.

We have devised schemes for compressing parameters of Bayesian logistic sequence prediction models and Bayesian logistic classification models, as described in Section~\ref{sec-blsm} and~\ref{sec-blcm}. The algorithm  for sequence prediction models is efficient. The resulting groups of interaction patterns have unique expressions. In contrast, the algorithm for classification models works well for problems of moderate size, but will be slow for problems with a large number of cases and with very high order. The resulting groups of interaction patterns may not have unique expressions, requiring extra work to merge the groups with the same expression afterward. A better algorithm that can compress the parameters in shorter time and have a more concise representation of the group of parameters may be found for classification models, though the improvement will not shorten the training time.

We have also empirically demonstrated that Cauchy distributions with location parameter $0$, which have heavy two-sided tails, are more appropriate than Gaussian distributions in capturing the prior belief that most of the parameters in a large group are very close to $0$ but a few of them may be much larger in absolute value, as we may often think appropriate for the regression coefficients of a high order.

We have implemented the compression method only for classification models in which the response and the features are both discrete. Without any difficulty, the compression method can be used in regression models in which the response is continuous but the features are discrete, for which we need only use another distribution to model the continuous response variable, for example, a Gaussian distribution. Unless one converts the continuous features into discrete values, it is not clear how to apply the method described in this thesis to continuous features. However it seems possible to apply the more general idea that we need to work only with those parameters that matter in likelihood function when training models with MCMC, probably by transforming the original parameters.

%% file: ref.tex
\chapter*{Bibliography}

\chaptermark{}

\renewcommand{\thechapter}{}
\renewcommand{\chaptername}{Bibliography}

\addcontentsline{toc}{section}{Bibliography}

\begin{description}

\itemsep 2pt

\item[]
  Alon, U., Barkai, N., Notterman, D.~A., Gish, K., Ybarra, S., Mack, D.,
  and Levine, A.~J.\ (1999)
  ``Broad patterns of gene expression revealed by clustering analysis
  of tumor and normal colon tissues probed by oligonucleotide arrays",
  \textit{Proceedings of the National Academy of Sciences (USA)}, vol.~96,
  pp. 6745-6750.

\item[]
Ambroise, C. and McLachlan, G. J. (2002) ``Selection Bias in Gene Extraction on the Basis of Microarray Gene-expression Data'', \textit{PNAS}, volumn 99, number 10, pages 6562-6566. Available from http://www.pnas.org/cgi/content/abstract/99/10/6562,

\item[]

Bishop, C. M. (2006) \textit{Pattern Recognition and Machine Learning}, Springer.

\item[] Baker, J.~K.\ (1975). ``The Dragon system - an overview'', \textit{IEEE
Transactions on. Acoustic Speech Signal Processing} ASSP-23(1): 24-29.


\item []

  Bell, T.~C., Cleary, J.~G., and Witten, I.~H. (1990) \textit{Text Compression},
  Prentice-Hall

\item[]

Cowles, M. K. and Carlin, B. P. (1996) ``Markov Chain Monte Carlo Convergence Diagnostics: A Comparative Review'', {\em Journal of the American Statistical Association}, Vol. 91, Pages 883--904

\item[]

  Dawid, A.~P.\  (1982) ``The well-calibrated Bayesian'', \textit{Journal
  of the American Statistical Association}, vol.~77, no.~379, pp.~605-610.

\item[]

  Everitt, B.~S.\ and Hand, D.~J.\ (1981) \textit{Finite Mixture Distributions},
  London: Chapman and Hall.

\item[]

Eyheramendy, S., Lewis, D.  and Madigan, D.  (2003)  On the Naive Bayes Model for Text Categorization. {\it Artificial Intelligence and Statistics}.

\item[]

Feller, W.\ (1966) ``An Introduction to Probability Theory and its  Applications'', Volume II, New York: John Wiley

\item[]

  Friedman,  J.H. (1998) ``Regularized Discriminant Analysis'', \textit{Journal of the American Statistical Association}, volume 84, number 405, pp. 165--175

\item[]

Gelfand, A.E. and  Smith, A.F.M. (1990) ``Sampling-Based Approaches to Calculating Marginal Densities". {\it  Journal of American Statistical Association}, 85:398-409.

\item[]

Gelman, A., Bois, F. Y.,  and Jiang, J. (1996)  Physiological pharmacokinetic analysis using population modeling and informative prior distributions. {\em Journal of the American Statistical Association} {91}, 1400--1412. 

\item[]

Geman, S.  and Geman, D. (1984) ``Stochastic Relaxation, Gibbs Distributions, and the Bayesian Restoration of Images". {\it IEEE Transactions on Pattern Analysis and Machine Intelligence}, 6:721-741.

\item[]

 Guyon, I., Gunn, S., Nikravesh, M., and Zadeh, L.~A.\ (2006) \textit{Feature Extraction: Foundations and Applications} (edited volume), Studies in Fuzziness and Soft Computing, Volume 207, Springer.

\item[]

Hastie, T., Tibshirani, R., and Friedman, J.H. (2001) {\em The Elements of Statistical Learning}, Springer

\item[]

Hastings, W.K. (1970) ``Monte Carlo Sampling Methods Using Markov Chains and Their Applications'',	{\it Biometrika}, Vol. 57, No. 1., pp. 97-109.

\item[]

Jacques, I.\ and Judd, C.\ (1987) \textit{Numerical Analysis}, Chapman and Hall.

\item[]

Jelinek, F.  (1998) {\it Statistical Methods for Speech Recognition}, The MIT Press. Available from http://mitpress.mit.edu/catalog/item/default.asp?ttype=2\&tid=7447

\item[]

 Khan, J., Wei, J.S., Ringn{\'e}r, M., Saal, L.H., Ladanyi, M., Westermann, F., Berthold, F., Schwab, M., Antonescu, C.R., Peterson, C., (2001) ``Classification and diagnostic prediction of cancers using gene expression profiling and artificial neural networks'', \textit{Nature Medicine},  vol 7,  pp. 673--679

\item[]

Lee, C., Landgrebe, D.A., and  National Aeronautics and Space Administration and United States (1993), ``Feature Extraction and Classification Algorithms for High Dimensional Data'', \textit{School of Electrical Engineering, Purdue University; National Aeronautics and Space Administration; National Technical Information Service, distributor}

\item[] 

Lecocke, M. L. and Hess K. (2004) ``An Empirical Study of Optimism and Selection Bias in Binary Classification with Microarray Data". UT MD Anderson Cancer Center Department of Biostatistics Working Paper Series. Working Paper 3. Available from http://www.bepress.com/mdandersonbiostat/paper3

\item[]

Li, L., Zhang, J., and Neal, R. M. (2007) ``A Method for Avoiding Bias from Feature Selection with Application to Naive Bayes Classification Models'', {\em Technical Report No. 0705, Department of Statistics, University of Toronto}. \\ Available from http://www.cs.toronto.edu/$\sim$radford/selbias.abstract.html.

\item[]

Li, Y. H. and  Jain, A. K. (1998) ``Classification of Text Documents'', {\it The Computer Journal}  41(8): 537-546.

\item[]

 Liu,  J.~S.\ (2001) \textit{Monte Carlo Strategies in Scientific Computing}, Springer-Verlag.

\item[]

Metropolis, N., Rosenbluth, A. W., Rosenbluth, M. N., Teller, A. H., Teller, E. (1953) ``Equation of State Calculations by Fast Computing Machines'', {\it 	The Journal of Chemical Physics}, Vol. 21, No. 6, pp. 1087-1092.

\item[]
  
McLachlan, G.~J.\ and Basford, K.~E.\ (1988) \textit{Mixture Models: Inference and Applications to Clustering}, New York: Springer-Verlag.

\item[]

Neal, R.~M.\ (1992) ``Bayesian mixture modeling'', in C.~R.~Smith, G.~J.~Erickson, and P.~O.~Neudorfer (editors) \textit{Maximum Entropy and Bayesian Methods: Proceedings of the 11th International Workshop on Maximum Entropy and Bayesian Methods of Statistical Analysis, Seattle, 1991}, p.~197-211, Dordrecht: Kluwer Academic Publishers.

\item[]

Neal, R.~M.\ (1993) \textit{Probabilistic Inference Using Markov Chain Monte Carlo Methods}, Technical Report CRG-TR-93-1, Dept.\ of Computer Science, University of Toronto, 140 pages.  Available from \texttt{http://www.cs.utoronto.ca/$\sim$radford/}.

\item[] 

Neal, R.~M.\ (1996) \textit{Bayesian Learning for Neural Networks}, Lecture Notes in Statistics No. 118, New York: Springer-Verlag.

\item[]

Neal, R.~M. \ (2003) ``Slice Sampling'', \textit{Annals of Statistics}, vol. 31, p.~705-767

\item[]

Raudys, S., Baumgartner, R. and Somorjai, R. (2005) ``On Understanding and Assessing Feature Selection Bias'', \textit{Artificial Intelligence in Medicine}, Page 468-472, Springer. Available from  http://www.springerlink.com/content/8e41e3wncj7yqhx3

\item[]

Ritchie, M.~D., Hahn, L.~W., Roodi, N., Bailey, L.~R., Dupont,W.~D.,  Parl,F.~F.,\ and Moore, J.H.\ (2001) ``Multifactor-Dimensionality Reduction Reveals High-Order Interactions among Estrogen-Metabolism Genes in Sporadic Breast Cancer'', \textit{The American Journal of Human Genetics}, volume 69, pages 138-147 

\item[] 

Roberts, G.O., and Rosenthal, J.S. (2004) ``General state space Markov chains and MCMC algorithms'', \textit{Probability Surveys} 1:20-71

\item[] 

Romberg, J., Choi, H. and Baraniuk, R. (2001) ``Bayesian tree-structured image modeling using wavelet-domain hidden Markov models'', \textit{IEEE Transactions on image processing} 10(7): 1056-1068.

\item[]
Rosenthal, J. S. (1995)   Minorization Conditions and Convergence Rates for Markov Chain Monte Carlo, {\em Journal of the American Statistical Association} 90:558-566

\item[]

Singhi, S. K. and Liu, H. (2006) ``Feature Subset Selection Bias for Classification Learning'', \textit{Proceedings of the 23rd International Conference on Machine Learning}. Available from http://imls.engr.oregonstate.edu/www/htdocs/proceedings/icml2006/\\ 107\_Feature\_Subset\_Selec.pdf

\item[]

Sun, S. (2006) ``Haplotype Inference Using a Hidden Markov Model with Efficient Markov Chain Sampling'', PhD Thesis, University of Toronto

\item[]

Tadjudin, S. and Landgrebe, D. A.  (1998) ``Classification of High Dimensional Data with Limited Training Samples'', \textit{TR-ECE 98-8, School of Electrical and Computer Engineering Purdue University}. \\ Available from http://cobweb.ecn.purdue.edu/$\sim$landgreb/Saldju\_TR.pdf

\item[]

Tadjudin, S. and Landgrebe, D. A.  (1999) ``Covariance estimation with limited training samples'', \textit{Geoscience and Remote Sensing, IEEE Transactions on}, volume 37,number 4, pp. 2113--2118

\item[]

Tierney, L. (1994). ``Markov chains for exploring posterior distributions."  {\it Annals of Statistics}, 22: 1701-1762.

\item[]

Titterington, D.~M., Smith, A.~F.~M., and Makov, U.~E.\ (1985) \textit{Statistical Analysis of Finite Mixture Distributions}, Chichester, New York: Wiley.

\item[]

Thisted, R.~A.\ (1988) \textit{Elements of Statistical Computing}, Chapman and Hall.

\item[]

Vaithyanathan, S., Mao, J. C., and Dom, B. (2000) ``Hierarchical Bayes for Text Classification'', {\it PRICAI Workshop on Text and Web Mining},	pages 36--43

\end{description}